\documentclass[10pt,twocolumn,letterpaper]{article}

\usepackage[pagenumbers]{iccv} %

\usepackage{tikz}

\usepackage{amssymb}
\usepackage{caption}

\usepackage{float}
\usepackage{pifont}
\usepackage{soul}
\usepackage{xfrac}
\newcommand{\cmark}{\ding{51}}
\newcommand{\xmark}{\ding{55}}

\newcommand{\scivid}{\textsc{SciVid}\xspace}

\newcommand{\flyvsfly}{FlyVsFly\xspace} %
\newcommand{\calms}{CalMS21\xspace}
\newcommand{\digitaltyphoon}{Digital Typhoon\xspace}
\newcommand{\weatherbench}{Weatherbench 2\xspace}
\newcommand{\stir}{STIR\xspace}

\newcommand{\withcond}{\emph{w. cond.}\xspace}
\newcommand{\withoutcond}{\emph{w.o. cond.}\xspace}

\usepackage[toc,page,header]{appendix}

\usepackage{tabularx} %
\newcolumntype{Y}{>{\centering\arraybackslash}X}
\usepackage{adjustbox}
\usepackage{multirow}
\usepackage{array}
\newcommand{\PreserveBackslash}[1]{\let\temp=\\#1\let\\=\temp}
\newcolumntype{C}[1]{>{\PreserveBackslash\centering}p{#1}}
\newcolumntype{R}[1]{>{\PreserveBackslash\raggedleft}p{#1}}
\newcolumntype{L}[1]{>{\PreserveBackslash\raggedright}p{#1}}
\newcommand{\frozen}{\raisebox{-0.2ex}{\includegraphics[width=0.3cm]{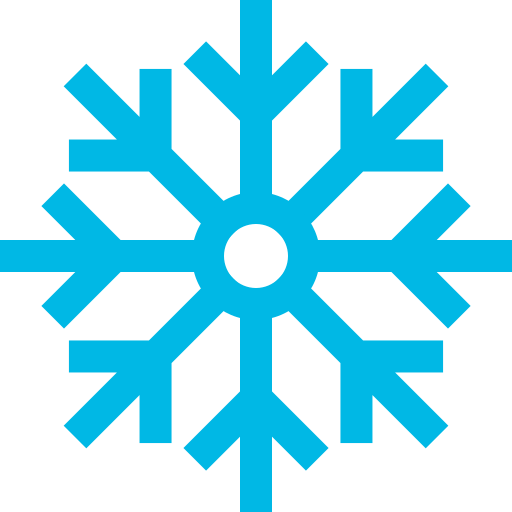}}}
\newcommand{\finetuned}{\raisebox{-0.2ex}{\includegraphics[width=0.3cm]{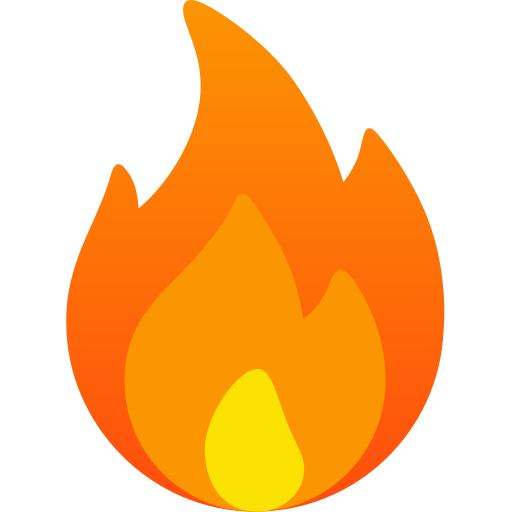}}}

\newcommand{\tightpara}[1]{\vspace{-2mm}\paragraph{#1}}

\usepackage{titletoc} %
\definecolor{iccvblue}{rgb}{0.21,0.49,0.74}
\usepackage[pagebackref,breaklinks,colorlinks,allcolors=iccvblue]{hyperref}

\definecolor{lightyellow}{rgb}{1, 1, 0.7}
\definecolor{lightblue}{rgb}{0.52, 0.78, 1}
\definecolor{lightpink}{rgb}{1, 0.8, 0.8}

\newcommand{\hly}[1]{\colorbox{lightyellow}{\textcolor{black}{#1}}} %
\newcommand{\hlb}[1]{\colorbox{lightblue}{\textcolor{black}{#1}}}   %
\newcommand{\hlp}[1]{\colorbox{lightpink}{\textcolor{black}{#1}}}   %

\usepackage{array}

\def\paperID{5334} %
\def\confName{ICCV}
\def\confYear{2025}

\title{\scivid: Cross-Domain Evaluation of Video Models in Scientific Applications}

\author{Yana Hasson \quad Pauline Luc \quad Liliane Momeni \quad Maks Ovsjanikov \quad Guillaume Le Moing  \\ Alina Kuznetsova \quad Ira Ktena \quad Jennifer J. Sun \quad Skanda Koppula \quad Dilara Gokay \\ Joseph Heyward \quad Etienne Pot \quad Andrew Zisserman  \\
Google DeepMind \\
{\tt\small Corresponding authors: \{yhasson,paulineluc\}@google.com}
}

\begin{document}
\maketitle 

\begin{figure*}[htbp]
\centering
\includegraphics[width=\linewidth]{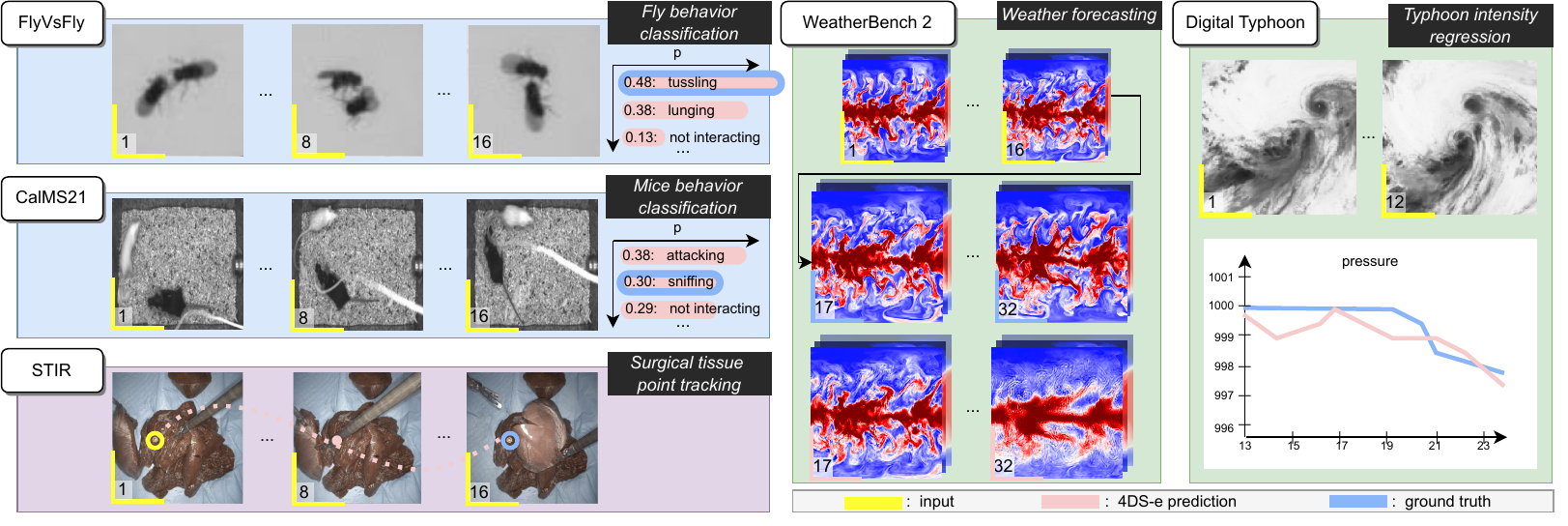}
\vspace{-5mm}
\caption{\scivid: A diverse collection of five scientific video tasks -- sourced from animal behavior, medical imaging and weather forecasting domains -- that evaluates a range of spatio-temporal reasoning capabilities, including classification, point tracking and forecasting. We show for each task, \hly{inputs} (yellow), \hlb{ground truth} (blue) and \hlp{predictions} (pink) from our framework with the 4DS-e~\cite{carreira2024scaling4drepresentations} backbone.\vspace{-2mm}}
\label{fig:teaser}
\end{figure*}

\begin{abstract}
In recent years, there has been a proliferation of spatiotemporal foundation models in different scientific disciplines. While promising, these models are often domain-specific and are only assessed within the particular applications for which they are designed. Given that many tasks can be represented as video modeling problems, video foundation models (ViFMs) hold considerable promise as \textit{general-purpose} domain-agnostic approaches. However, it is not known whether the knowledge acquired on large-scale but potentially out-of-domain data can be effectively transferred across diverse scientific disciplines, and if a single, pretrained ViFM can be competitive with domain-specific baselines.  To address this, we introduce \scivid, a comprehensive benchmark comprising five \underline{\textsc{Sci}}entific \underline{\textsc{Vid}}eo tasks, across medical computer vision, animal behavior, and weather forecasting. We adapt six leading video models to \scivid using simple trainable readout modules, establishing strong baselines and demonstrating the potential for effective transfer learning. Specifically, we show that state-of-the-art results can be obtained in several applications by leveraging the general-purpose representations from ViFM backbones. Furthermore, our results reveal the limitations of existing ViFMs, and highlight opportunities for the development of generalizable models for high-impact scientific applications.
We release our code at \url{https://github.com/google-deepmind/scivid} to facilitate further research in the development of cross-domain ViFMs.

\end{abstract}

\section{Introduction}
\label{sec:intro}

The past decade has seen substantial progress in scientific applications through the use of computer vision and machine learning techniques~\cite{Wang2023ScientificDI}. This has led to novel insights in disciplines ranging from medical image analysis~\cite{caron_endoscopy}, animal behavior understanding~\cite{Bain_primates}, and climate change modelling~\cite{rolnick_2023_tackling_climate_change_ai}, among others. 
Despite these successes, this progress has been relatively slow, typically requiring machine learning scientists and domain specialists to work hand in hand, collect training data and design specialized models.%

In parallel, the emergence of foundation models\footnote{By foundation models, we refer to ``models that are trained on broad data and that can be adapted (e.g., finetuned) to a wide range of downstream tasks''~\cite{Bommasani2021FoundationModels}.} has considerably democratized the use of machine learning models across a myriad of applications, 
either through prompting or by training lightweight modules.
While image encoders~\cite{radford2021clip,oquab2023dinov2} have proven valuable for processing medical imaging modalities such as X-rays and MRIs~\cite{zhao2023clip,baharoon2023evaluating}, the potential of recent video foundation models (ViFMs)~\cite{madan2024foundation} for diverse spatio-temporal modeling tasks remains largely unexplored. 
Currently, the applications of ViFMs in scientific disciplines typically rely on domain-specific training data~\cite{christensen2024vision,wang2023foundation,schmidgall2024general} and, even when trained more broadly, are only evaluated within a particular  discipline~\cite{sun2024video}.  As a result, it is unclear, first, to what extent there might be information sharing \textit{across} disciplines, and, second, whether foundation models pretrained on large, but possibly out-of-domain datasets, can provide a useful signal for vastly different spatio-temporal applications.

To answer these questions, we build a common test bed and
study whether pretrained foundation models can be effectively repurposed for diverse scientific applications. We introduce \scivid, an open-source benchmark suite featuring a broad range of domains, tasks and data regimes (see Fig.~\ref{fig:teaser} and Tab.~\ref{tab:stats}). For each application in \scivid, we perform a careful comparison between state-of-the-art task-specific baselines and discipline-agnostic ViFMs. We study the effect of using different ViFMs, pretraining strategies and adaptation protocols following current best practices. Remarkably, our study shows that ViFMs can attain state-of-the-art results in science-related tasks, even when pretrained on out-of-domain data and employing simple trainable readout modules. In addition to these positive findings, we also observe that in several applications ViFMs have modest performance, paving the way for future research.

We stress that \scivid allows evaluation with limited domain expertise. We hope that this will facilitate the adoption and prioritization of scientific applications by the computer vision community, and increase the utility of video foundation models in science-related tasks. %

To summarize, our key contributions are: (1) A dedicated benchmark for ViFMs across multiple scientific disciplines. (2) A thorough comparison of existing ViFMs. Our results reveal that ViFMs attain state-of-the-art performance in multiple domains, while also highlighting their limitations in others.
(3) A comprehensive analysis to identify important factors of performance in ViFM adaptation.

\begin{table*}[!htbp]
\small
\centering
\setlength{\tabcolsep}{5pt}
\begin{tabularx}{\textwidth}{l|c|c|c|c|c}
\toprule
Dataset         & FlyVsFly~\cite{eyjolfsdottir2014detecting} & CalMS21~\cite{sun2021multi} & \stir~\cite{stir_Schmidt_2024} & WeatherBench 2~\cite{rasp2024wb2} & Digital Typhoon~\cite{digital_typhoon_NEURIPS2023} \\
\midrule
Domain & Animal behavior &  Animal behavior & Medical & Weather & Weather \\
Task & Classification & Classification & Point Tracking & Spatiotemporal Forecasting & Temporal Forecasting            \\
Input type           & Grayscale & Grayscale & RGB & Z500, T850, Q700 & Infrared satellite   \\
Outputs            & Behavior labels & Behavior labels & Final positions & 16 future frames &  Pressure (12-step)        \\
Label source            & Expert & Expert & Expert \& Non-expert & Reanalysis & Reanalysis           \\
Grid resolution            & $144 \times 144$     &  $285 \times 512$ & $1024 \times 1280$   & $181 \times 360$    & $128 \times 128$  \\
FPS            & 30 & 30 & 25 & 12-hourly &  hourly \\
\# Input frames            & 16 & 16 & 7 to 19419 & 16 (8 days) & 12            \\
\# Samples train           & 1M & 27K & N/A & 57K & 696           \\
\# Samples val           & 162K & 77K & 487 & 730 & 174           \\
\# Samples test          & 322K & 262K & 60 & 732 & 219          \\
\bottomrule
\end{tabularx}
\vspace{-3mm}
\caption{\textbf{\scivid characteristics and statistics.} We provide an overview of each benchmark. Z500, T850 and Q700 respectively stand for geopotential, temperature and specific humidity at pressure levels 500, 850 and 700. No training data is provided for STIR.\vspace{-3mm}} 

\label{tab:stats}
\end{table*}

\section{Related work}
\label{sec:related_work}

We summarize relevant works on video modeling and the application of foundation models in scientific domains.

\subsection{Video models}
Inspired by the success of foundation models in language and image domains, numerous approaches have emerged to train general-purpose models on \textit{video} data~\cite{madan2024foundation,goodge2025spatio,awais2025foundation}. These ViFMs broadly fall into two categories: (1) Multimodal ViFMs, which use supervision from another modality  (often text), and are typically evaluated on language-centric tasks~\cite{maaz2023video,alayrac2022flamingo,lin2023video,dubey2024llama,team2024gemini,wang2022internvideo,zhaovideoprism}. Despite their generality, such multimodal ViFMs often struggle with spatio-temporal reasoning~\cite{yuan2023videoglue}. (2) Unimodal ViFMs,
which focus on self-supervised learning from videos alone through Masked Auto-Encoding in either the pixel~\cite{tong2022videomae,wang2023videomae} or latent spaces~\cite{bardes2023vjepa}. In this work, we focus mainly on the latter.

\subsection{Foundation models in scientific domains}
We briefly review applications in medicine, animal behavior and atmospheric science. We refer to recent surveys~\cite{awais2025foundation,jiang2024foundation,chen2023foundation} for more in-depth discussions.

While initial efforts in developing and using foundation models for medical applications primarily focused on images and text~\cite{zhao2024biomedparse,yan2024general,zhang2024generalist}, ViFMs are more recently gaining traction. Examples include Endo-FM for endoscopy~\cite{wang2023foundation}, EchoCLIP for echocardiograms~\cite{christensen2024vision}, and a ViFM for forecasting across diverse surgical procedures~\cite{schmidgall2024general}.

ViFMs are also relevant in animal behavior studies, where data can be particularly scarce~\cite{sun2025toward}. Pretrained task-specific models have been used in low-level tasks such as pose estimation from images or videos~\cite{ye2024superanimal,yang2024x}, and ViFMs have been adapted to tasks such as classification, retrieval, and localization~\cite{sun2024video}. While \scivid includes animal behavior tasks similarly to~\cite{zhaovideoprism}, our benchmark addresses a broader range of scientific applications, across many backbones, and evaluating a range of adaptation strategies.

Another domain in which foundation models have been developed and used is weather and climate modeling~\cite{chen2023foundation,shi2025deep,mukkavilli2023ai}. 
Because weather and climate are governed by universal physical laws,
one might expect to see improved performance and generalization from training on a range of different weather and climate modeling tasks. %
For example, ClimaX~\cite{nguyen2023climax} is trained using datasets spanning different variables, spatio-temporal coverage, and physical groundings %
and Bodnar et al.~\cite{bodnar2024aurora} introduced a large-scale Earth system model for diverse forecasts.

While there has been a surge of foundation models in nearly every scientific domain~\cite{neurips2024fm4science}, relatively little effort has been made to evaluate \emph{different ViFMs} %
across \emph{multiple scientific disciplines}. 
ViFMs have been benchmarked mostly within specific scenarios, such as animal behavior understanding~\cite{jing2024animal}, geospatial modeling~\cite{fibaek2024phileo} and computational pathology~\cite{neidlinger2024benchmarking}.
Perhaps most closely related to ours is a \textit{concurrent} effort VideoEval \cite{li2024videoeval}, which evaluates a range of ViFMs on a new benchmark of challenging tasks. Our work differs in that \scivid covers a broader scope of domains, and in the careful comparison between ViFM backbones and the best performing task-specific methods.

\section{\scivid suite}

In this section, we present our \scivid benchmark suite. We first describe how we selected tasks in Sec.~\ref{subsec:criteria}. We then describe each task: animal behavior classification (Sec.~\ref{subsec:animal-behavior}), tissue tracking (Sec.~\ref{subsec:tracking}), weather forecasting (Sec.~\ref{subsec:weather}), pressure forecasting (Sec.~\ref{subsec:pressure-forecasting}), along with the respective datasets and evaluation metrics. In the supp.~mat, we provide descriptions of domain-specific baselines (Sec.~\ref{sec:supp_baselines}) and further specification of evaluation metrics (Sec.~\ref{sec:supp_impl_details}).

\subsection{Selection of benchmark tasks}\label{subsec:criteria}

The \scivid suite is constructed to provide a rigorous evaluation of video models across important scientific applications. Our selection of benchmark tasks was driven by the following principles: (1)~ensuring a broad coverage of challenges inherent in scientific applications, (2)~incorporating diverse domains and distribution shifts to assess generalization capabilities beyond standard RGB video data, (3)~emphasizing tasks that require temporal understanding, (4)~integrating both under-explored and well-established tasks, to drive innovation while providing a solid foundation for performance comparison.

\subsection{Animal behavior classification} \label{subsec:animal-behavior}

\noindent \textbf{Task definition:} Given a video clip of an interacting pair of animals, predict the annotated behavior of the center frame.

\noindent \textbf{Datasets:} We evaluate models on \flyvsfly and \calms:

 \textbf{(i) \flyvsfly~\cite{eyjolfsdottir2014detecting}.} To study the effect of genetic manipulation on fly behavior, biologists have recorded and annotated 20 hours of videos of interacting pairs of flies. \flyvsfly contains these videos with frame-level, expert annotations for 7 social behaviors (lunge, wing threat, tussle,
wing extension, circle, copulation, not interacting). We use the Aggression and Courtship sets, where the flies are recorded in a circular 16mm diameter chamber with uniform food surface at 30Hz. Following previous work~\cite{Sun_2021_CVPR_task_programming}, we use 16-frame video clips.
The test and train splits are the same as those used in~\cite{zhaovideoprism}, allowing for direct comparison.%

\textbf{(ii) \calms~\cite{sun2021multi}.} Quantifying animal behavior is crucial for studying the relationship between brain and behavior, and understanding neurological disorders. \calms contains videos of mice social behavior recorded and annotated frame-by-frame by neuroscientists. We focus on \emph{Task 1} from~\cite{sun2021multi}, annotated with 4 social behaviors (close investigation, attack, mount, background). The videos are recorded from the top-view in a standard resident-intruder assay containing a pair of mice, with genetic or other background differences between the animals. 
The original dataset is extracted with a temporal stride of 1, resulting in input redundancies.
In order to reduce the memory footprint of the dataset, we subsample the \emph{train} set by using a temporal stride of 16, and spatially downscale all frames by a factor of 2.
Despite using a subset of the training data, we outperform the previous state of the art (see Tab.~\ref{tab:videoprism_sota}).

\noindent \textbf{Evaluation metrics:} Following VideoPrism~\cite{zhaovideoprism}, we use mean average precision (mAP), discarding the background class, and averaging over classes to address class imbalance.

\subsection{Tissue tracking}\label{subsec:tracking}

\noindent \textbf{Task definition:} Given a set of query points in the first frame of a video clip, track all the points till the final frame.

\noindent \textbf{Dataset: \stir~\cite{stir_Schmidt_2024}.} Accurate tracking of surface motion is crucial in surgical robotics, and necessary for understanding tissue dynamics and deformation. \stir includes animal tissue motion tracks during surgical manipulations for 487 RGB videos ranging from 0.28 seconds to 13 minutes.
Locations are tattooed using infrared ink invisible in RGB images.
Tracked locations are automatically extracted by segmenting infrared tattoos in the initial and final video frames.

\noindent \textbf{Evaluation metrics:}  
Following the STIR challenge~\cite{STIRChallenge, doersch2022tapvid}, we report $\delta_{avg}^{x}$ which corresponds to the 2D position accuracy averaged over thresholds of \{4, 8, 16, 32, 64\} pixels.

\subsection{Weather forecasting}\label{subsec:weather}

\noindent \textbf{Task definition:} Given the data for the past 8 days sampled at 12 hour intervals, the goal is to generate an 8-day forecast with the same time step, for three  of the most commonly studied weather variables: Geopotential at 500hPa (Z500), Temperature at 850hPa (T850) and Specific humidity 700hPa (Q700).\footnote{In weather modeling, it is common to use pressure, measured in hectopascals (hPa), as a vertical coordinate instead of altitude.}

\noindent \textbf{Dataset: WeatherBench 2 ERA5.~\cite{rasp2024wb2}} Accurately forecasting the weather is crucial for agriculture, energy demand prediction and planning timely response to extreme events~\cite{keisler2022gnn, bi2023pangu, lam2023graphcast, chen2023cfuxi, esteves2023sphericalcnn, kurth2023fourcastnet, kochkov2023neuralgcm, price2025gencast}. Weatherbench 2 ERA5~\cite{rasp2024wb2} consists of reanalysis data, that is, a ``best guess'' of the state of variables in medium-range weather forecasting, at any point in space and time, estimated by ECWMF's HRES simulator, given a wide range of direct (but potentially sparse) and remote sensing observations. We split the data following~\cite{rasp2024wb2,lam2023graphcast}: the train, validation and test splits consists of all trajectories for the years 1979-2017, 2018 and 2020, respectively.
While generally a spatial resolution of 0.25° is used, we choose a coarser 1° resolution: this corresponds to a spatial resolution of $181 \times 360$ pixels, close to the pretraining resolution of existing ViFMs.

\noindent \textbf{Evaluation metrics:} Following common practice~\cite{rasp2024wb2}, we measure performance using the area-weighted root mean squared error (wRMSE). We further average performance for each variable across frames.

\subsection{Pressure forecasting}\label{subsec:pressure-forecasting}

\noindent \textbf{Task definition:} Given a sequence of infrared cyclone images, predict a 12-hour forecast of the cyclone's central pressure (a single scalar per hour).

\noindent \textbf{Dataset: Digital Typhoon~\cite{digital_typhoon_NEURIPS2023}.} Tropical cyclones rank among the most destructive natural disasters~\cite{hurricane}. Accurate forecasting of cyclone intensity can help minimize human casualties and mitigate economic losses~\cite{econometrics8020018, chavas2017physical}. Digital Typhoon presents typhoon-centered infrared images for 1099 events with meteorological annotations. %
Following~\cite{digital_typhoon_NEURIPS2023}, for evaluation, we use the hourly data for the first 24 hours of each sequence. Specifically, given 12 input infrared frames, we regress the hourly pressure for the subsequent 12 hours. While the results in the original paper~\cite{digital_typhoon_NEURIPS2023} are computed for 20\% train/test sequence splits for 5 random seeds, we generate a fixed 80\%/20\%/20\% train/validation/test split. %

\noindent \textbf{Evaluation metrics:}
The original implementation~\cite{digital_typhoon_NEURIPS2023} reported RMSE errors for forecast time steps \{1, 2, 3, 6, 12\}. In order to provide a single metric value for comparison, we report the average RMSE across these five time steps.

\section{Evaluation methodology}
\label{sec:eval}

\begin{figure}[htbp]
\centering
\includegraphics[width=\linewidth]{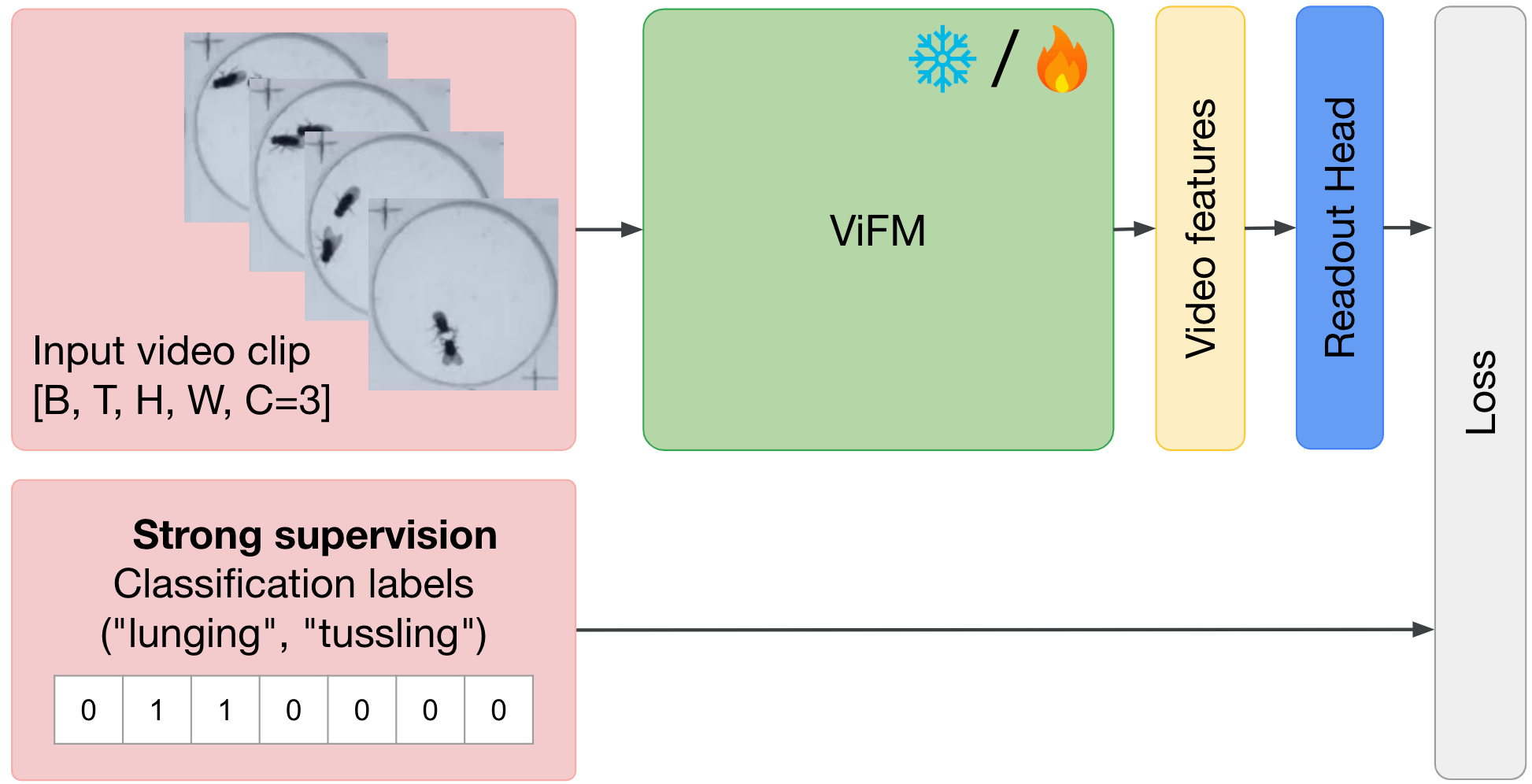}
\caption{\textbf{Evaluation overview.} We extract video features using pretrained ViFMs, providing 3-channel spatio-temporal clips as input.
For each target task, we train lightweight readouts which take the video features from the frozen backbone (\frozen) as input. We also explore finetuning the backbone (\finetuned) for certain tasks. The readouts are trained from scratch using strong supervision.
\vspace{-3mm}}
\label{fig:evaluation_overview}
\end{figure}

We evaluate a large set of ViFMs on \scivid. We follow~\cite{carreira2024scaling4drepresentations} and use a consistent representation learning paradigm where pretrained models are appended with task-specific readouts and subsequently finetuned on each downstream dataset, with or without freezing the backbone (see Fig.~\ref{fig:evaluation_overview}). In this section, we provide the details on the vision foundation backbones (Sec.~\ref{subsec:backbones}) and readout architectures (Sec.~\ref{subsec:task_readouts}). 
Our experimental setup with frozen backbones requires less than a day to complete on a single H100 GPU, for all ViFMs and all five \scivid tasks.
Further implementation details can be found in Sec.~\ref{sec:supp_impl_details} of the supp.~mat. 

\subsection{Backbones}
\label{subsec:backbones}

We evaluate the encoding capabilities of multiple ViFMs, prioritizing works with self-supervised objectives. %

\noindent \textbf{Image models.} We assess the performance of DINOv2~\cite{oquab2023dinov2}, a prominent image model, utilizing both the ViT-L and ViT-g variants. DINOv2 is trained exclusively on images, without any language supervision, using a self-distillation loss. To ensure fair evaluation with video models in our frozen backbone setting, learnable temporal positional embeddings are added to the image model's features before task readout training, following~\cite{carreira2024scaling4drepresentations}.

\noindent \textbf{Video models.} We evaluate several state-of-the-art video models: VideoPrism~\cite{zhaovideoprism}, VideoMAE~\cite{tong2022videomae, wang2023videomae}, V-JEPA~\cite{bardes2023vjepa} and the recent 4DS~\cite{carreira2024scaling4drepresentations} model family.
VideoPrism, based on ViViT~\cite{vivit}, employs a two-stage training process: first, contrastive learning between video-text pairs, followed by masked autoencoding on video-only clips. Specifically, in the second stage, given masked input video patches, the model is required to predict global video embeddings and token-level embeddings of unmasked frames from the initial stage. We examined both VideoPrism-B and VideoPrism-g.
VideoMAE, V-JEPA and 4DS are all video-only self-supervised models, based on a space-time ViT backbone, trained by predicting masked spatio-temporal regions. While VideoMAE and 4DS are tasked with predicting in the pixel space, V-JEPA predicts in the learned latent space of a teacher network. We assessed a range of model sizes, from ViT-B to ViT-g, for VideoMAEv1, VideoMAEv2, and V-JEPA. For 4DS, we evaluate 4DS-e and 4DS-L, which correspond to 4B and 300M model parameters configurations.

\noindent \textbf{Resize baseline.} As a naive baseline, we also use resized versions of the input video as a simplified parameter-free backbone. In practice, we use the average value across 14x14 spatial pixel patches as inputs to the readout.
Comparing against this baseline allows us to verify that the backbones extract non-trivial information from the input videos.

\subsection{Task readouts}
\label{subsec:task_readouts}

We reuse or adapt the readouts implemented in~\cite{carreira2024scaling4drepresentations} for similar tasks. Further training details can be found in Sec.~\ref{subsec:supp_readouts} of the supp.~mat.

\noindent \textbf{Classification.} Similarly to the classification readout for SSv2 in~\cite{carreira2024scaling4drepresentations}, we use a cross-attention readout module. A single learned query is used to predict logits for 4 classes (including the background class) for CalMS21 and 7 classes for FlyVsFly. We train by minimizing the sigmoid cross-entropy loss between the predicted and ground truth classes.

\noindent \textbf{Tracking.} Following the design of the Perception Test tracking readout  in~\cite{carreira2024scaling4drepresentations}, we use a cross-attention readout, where queries are given by embeddings of the query positions, and keys and values are provided by the backbone features, to predict all target positions, visibility and uncertainty estimates, simultaneously, for a fixed length window. 
We minimise the Huber loss for the positions, and binary cross-entropy for the other variables.

\noindent \textbf{Weather Forecasting.} We use the DPT readout~\cite{Ranftl2021}, which consists of a series of trainable convolution and reassemble layers, upsampling the input features into pixel-level predictions for all output frames. We also experimented with a pure attention-based readout, but it gave significantly worse accuracy (see Sec.~\ref{subsec:ablations}). We slightly adapt the capacity of the readout to accommodate for dense 3-channel future frame prediction, as it was initially designed for single channel depth estimation  (see  Tab.~\ref{table:readout_modules} and Sec.~\ref{subsec:supp_readouts} of the supp.~mat.\ for details). The readout is trained by minimizing a channel- and area-weighted L1 loss following~\cite{lam2023graphcast}.

\noindent \textbf{Pressure Forecasting.} We use the same cross-attention readout module as for classification tasks. In this case, from the single learned query, we predict a 12-dimensional vector of pressure values for 12 consecutive time steps. We train by minimising the L2 loss on offsets to the average train pressure, of 983.9 hPa, computed across the train set.

\section{Results}
\label{sec:results}

In this section, we first provide quantitative analysis of the desirable properties of \scivid (Sec.~\ref{subsec:quant_desirable_properties}). We then compare the best-performing ViFMs to state-of-the-art methods for each task (Sec.~\ref{subsec:sota_comparison}). We provide an extensive comparison of a range of ViFMs under the frozen evaluation protocol with trainable readouts (Sec.~\ref{subsec:backbone_comparison}).  We investigate different design choices in our adaptation framework (Sec.~\ref{subsec:ablations}). Finally, we describe qualitative results (Sec.~\ref{subsec:qualitative}). 

\subsection{Quantitative analysis of \scivid properties}
\label{subsec:quant_desirable_properties}

\begin{figure}[H]
\centering
\includegraphics[trim = 0 1.1cm 0 0, clip=true,width=\linewidth]{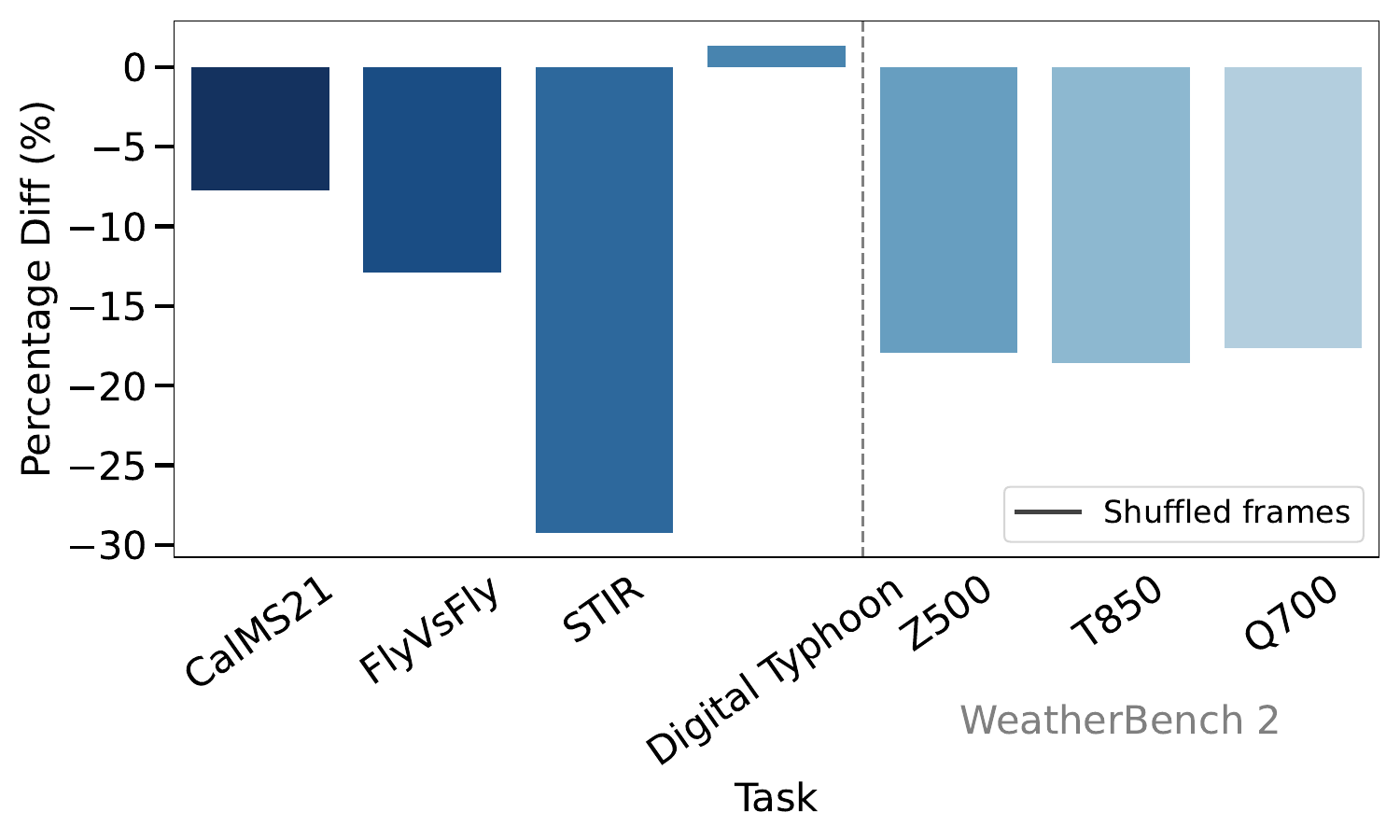}
\vspace{-7mm}
\caption{\textbf{Temporal modeling.} We show the (percentage) performance degradation with respect to our readout training with the frozen 4DS-e backbone~(\frozen) and using ordered video frames, compared to random shuffling. For STIR we ensure  that the tracked points correctly correspond to each frame after shuffling.\vspace{-3mm}}
\label{fig:temporal_dynamics}
\vspace{-3mm}
\end{figure}

\tightpara{Temporal dynamics.} We investigate to what extent our selected tasks require temporal modeling.
To shed light on this we compare our readout with 4DS-e (in Fig.~\ref{fig:temporal_dynamics}) to a counterpart trained on videos with shuffled frames. 
In this scenario we observe performance degradation across tasks against the standard setting, highlighting the importance of temporal dynamics.
The only exception is for \digitaltyphoon, but this discrepancy is within the noise range (see Fig.~\ref{fig:eval_noise} of the supp.~mat.). Overall we observe a stronger effect of maintaining temporal dynamics on the tracking task, followed by the weather forecasting task, and observe a lower impact on the classification tasks. %

\tightpara{Distribution shift.} We further provide a study of the shift between the video distributions of typical ViFM pretraining and \scivid datasets in Sec.~\ref{subsec:supp_input_distribution} of the supp. mat.

\begin{table}[t!]
\centering
\small
\setlength{\tabcolsep}{5pt}
\begin{tabular}{l|c|c|c|c}
\toprule
\multirow{2}{*}{Backbone} & Readout & Training & \flyvsfly & \calms \\
         &  Training & Steps    & mAP      & mAP    \\
\midrule
VideoPrism-g & ~\cite{zhaovideoprism} & $\approx$ 100k & \underline{92.0} & 91.5 \\
VideoPrism-g & Ours     & 40k         & 87.2     & \underline{92.0} \\
VideoPrism-g & Ours     & 400k        & \textbf{92.5} & 90.9    \\ %
\midrule
4DS-e~\cite{carreira2024scaling4drepresentations} & Ours     & 40k         & 84.6     & \underline{92.0} \\  %
DINOv2-L~\cite{oquab2023dinov2} & Ours     & 40k & 83.9       & 91.9       \\ %
VJEPA-H~\cite{bardes2023vjepa} & Ours & 40k & 84.5 & \textbf{92.4} \\ %
\bottomrule
\end{tabular}
\caption{\textbf{SOTA Comparison \flyvsfly \& \calms.} We outperform the VideoPrism-g~\cite{zhaovideoprism} state-of-the-art on the \calms test set with different frozen backbones~(\frozen): VJEPA, VideoPrism-g, 4DS-e and DINOv2-L. On \flyvsfly test, we observe that longer training is required to outperform the previous state of the art~\cite{zhaovideoprism}.\vspace{-1mm}}\label{tab:videoprism_sota} %
\end{table}
\subsection{Comparison to SOTA}
\label{subsec:sota_comparison}
We compare our framework with the top-2 best performing backbones, referred to as \emph{Ours}, against state-of-the-art methods. We report results for frozen backbones~(\frozen) on \scivid, reaching state-of-the-art results on three out five tasks. For the two remaining tasks (\stir and \weatherbench), we additionally report results with finetuned backbones~(\finetuned). We report the test set results at the final time step, and the validation set results at the best performing time step. For all tables, we display the best results in \textbf{bold} and \underline{underline} second best.

\tightpara{\calms and \flyvsfly.} As shown in Tab.~\ref{tab:videoprism_sota}, when training with \emph{our readout} using frozen features from VideoPrism-g~\cite{zhaovideoprism}, we outperform the VideoPrism-g~\cite{zhaovideoprism} previous state-of-the-art results on \flyvsfly (with longer training at 400k steps) and \calms (at 40k steps), validating our framework set up. On \calms, we reach state-of-the-art performance by using the frozen VJEPA-H backbone, achieving 92.4 mAP. 

\tightpara{\stir.} In Tab.~\ref{tab:stir_sota}, we report results on the \stir validation and test sets. All approaches demonstrate significant improvement over the control baseline which statically propagates the query positions in the first frame throughout the sequence. The Multi-Flow dense Tracker (MFT)~\cite{neoral2024mft}, which established a strong baseline by outperforming all submissions on \stir during the Endoscopic Vision Challenge~\cite{STIRChallenge}, achieves 68.5\% (resp.\ 77.6\%) accuracy on the validation set (resp.\ test set). %
Our readout with 4DS-e backbone achieves 51.3\% (resp.\ 57.8\%) with frozen features, and improves to 61.2\% (resp.\ 69.2\%) with finetuning. A per-threshold performance breakdown, detailed in Fig.~\ref{fig:threshold} of the supp.~mat., reveals that 4DS-e matches MFT's performance on the largest (64 pixels) threshold distance.  Importantly, our framework achieves this performance using a simple transformer architecture with a lightweight cross-attention readout, without any sophisticated tracking components (\eg, feature pyramids, correlation volumes, iterative refinements) employed by competing methods.
\tightpara{\digitaltyphoon.} In Tab.~\ref{tab:typhoon_sota}, we compare our readout training with the 4DS-L and V-JEPA-H backbones to the method from~\cite{digital_typhoon_NEURIPS2023}, of which numbers are obtained from their paper, on both the test and validation splits. We observe that our results are significantly better.
For this reason, we also include two other simple, stronger references: (i) \emph{mean train pressures}: we use average train-time pressure values for each time step as the prediction for all samples, i.e.~dataset statistics are used to produce input-agnostic predictions, (ii) \emph{copy last pressure}: we use the pressure from the last input step and copy it for the 12 output steps as the prediction.
Our method outperforms the mean train pressure baseline on both the validation and test sets, achieving state-of-the-art performance.
We note that while we report the \emph{copy last pressure} oracle to provide additional insights, this reference uses extra information (the last input central pressure) which is not accessible to other models at inference time. As shown in Fig.~\ref{fig:typhoon_sota} of the supp.~mat., we see that the \emph{copy last pressure} oracle achieves a lower RMSE at smaller time steps, benefitting from temporal continuity. However, at later time steps, our readout with the 4DS backbone consistently outperforms other approaches.

\begin{table}[t!]
\centering
\small
\begin{tabular}{lC{2.2cm}C{2.2cm}}
\toprule
\multirow{2}{*}{Method} & \multicolumn{2}{c}{$\delta_{avg}^{x}$ Accuracy (\%)} \\
\cline{2-3}\\[-0.9em]
                        & Val          & Test         \\
\midrule
Baseline (Control)      & 28.6         &  41.1 \\
CSRT~\cite{Lukezic_IJCV2018}      & 52.6         & 58.6 \\
RAFT~\cite{teed2020raft}      & 47.0         &  42.4\\
MFT~\cite{neoral2024mft}       & \textbf{68.5}         &  \textbf{77.6} \\
\midrule
VideoMAE-B \frozen ~(Ours)       & 47.8         & 52.1 \\
4DS-e \frozen ~(Ours)            & 51.3         &  57.8\\
VideoMAE-B \finetuned ~(Ours)       & 57.2         & 63.4 \\
4DS-e \finetuned ~(Ours)            & \underline{61.2}         &  \underline{69.2}\\
\bottomrule
\end{tabular}
\caption{\textbf{SOTA Comparison \stir.} We report $\delta_{avg}^{x}$, the average accuracy across pixel thresholds, on the validation and test sets. While MFT achieves SOTA, our readout using finetuned features from 4DS-e (\finetuned) achieves good performance given its simplicity.\vspace{-2mm}}
\label{tab:stir_sota}
\end{table}
\begin{table}[t]
\centering
\small
\begin{tabular}{lC{2.2cm}C{2.2cm}}
\toprule
\multirow{2}{*}{Method} & \multicolumn{2}{c}{Avg. RMSE} \\
\cline{2-3}\\[-0.9em]
                        & Val          & Test         \\
\midrule
Copy last pressure & \textit{3.71} & \textit{3.84} \\
\midrule
Mean train pressures & 10.04 & 9.59 \\
Kitamoto et al. 2023 & N/A & 11.71 \\
\midrule
4DS-L \frozen ~(Ours) & \textbf{3.88} & \textbf{5.23} \\
V-JEPA-H \frozen ~(Ours) & \underline{4.16} & \underline{5.53} \\
\bottomrule
\end{tabular}
\caption{\textbf{SOTA Comparison \digitaltyphoon}. We report the RMSE averaged across time steps 1, 2, 3, 6, 12. Our readout training with frozen features from 4DS-e backbone (\frozen) achieves state-of-the-art performance. \textit{Copy last pressure} corresponds to an oracle, benefitting from additional information.
\vspace{-3mm}}
\label{tab:typhoon_sota}
\end{table}

\begin{table*}[!htbp]
\setlength{\tabcolsep}{5.5pt}
\centering
\begin{tabularx}{\textwidth}{l|cc|ccccc}
\toprule
 & \# Params. & Language & CalMS21 & FlyVsFly & STIR & Digital Typhoon & WB2 Z500/T850/Q700 \\
 &  & Pretraining & mAP $\uparrow$ & mAP $\uparrow$ & Acc $\uparrow$ & RMSE $\downarrow$ & wRMSE  $\downarrow$ \\
\hline
4DS-e & 3811 & \xmark & 0.817 & \underline{0.894} & \textbf{0.513} & 4.23 & 601/2.87/\textbf{1.56e-03} \\
4DS-L & 310 & \xmark & 0.778 & 0.855 & 0.477 & \textbf{3.84} & 597/2.85/1.58e-03 \\
\hline
DinoV2-g & 1135 & \xmark & \underline{0.866} & 0.866 & 0.215 & 6.33 & 627/2.97/1.61e-03 \\
DinoV2-L & 303 & \xmark & \textbf{0.867} & 0.787 & 0.191 & 6.31 & 658/3.14/1.69e-03 \\
\hline
VideoMAE-H & 633 & \xmark & 0.782 & 0.863 & 0.430 & 4.82 & 604/2.89/1.58e-03 \\
VideoMAE-L & 305 & \xmark & 0.779 & 0.859 & 0.456 & 4.57 & \underline{594}/\underline{2.84}/1.57e-03 \\
VideoMAE-B & 87 & \xmark & 0.665 & 0.807 & \underline{0.478} & 4.93 & 606/2.90/1.62e-03 \\
\hline
VideoPrism-g & 1113 & \cmark & 0.855 & 0.839 & 0.351 & 5.01 & 635/3.05/1.71e-03 \\
VideoPrism-B & 114 & \cmark & 0.832 & 0.824 & 0.328 & 4.58 & 646/3.09/1.74e-03 \\
\hline
VideoMAEv2-g & 1,013 & \xmark & 0.862 & 0.887 & 0.344 & 4.53 & \textbf{594}/\textbf{2.83}/\underline{1.56e-03} \\  %
\hline
V-JEPA-H & 635 & \xmark & 0.828 & \textbf{0.901} & 0.443 & \underline{4.16} & 619/2.96/1.61e-03 \\
V-JEPA-L & 307 & \xmark & 0.829 & 0.890 & 0.408 & 4.93 & 614/2.93/1.60e-03 \\
\hline
Resize & 0 & N/A & 0.122 & 0.0951 & 0.280 & 10.0 & 642/3.17/1.89e-03 \\
\bottomrule
\end{tabularx}
\vspace{-3mm}
\caption{\textbf{Comparisons across frozen backbones~(\frozen)} using the same per-task evaluation protocol, data, readouts and hyperparameters.\vspace{-3mm}}
\label{tab:backbone_comparison}
\end{table*}

\begin{figure}[t]
\centering
\includegraphics[width=0.97\linewidth]{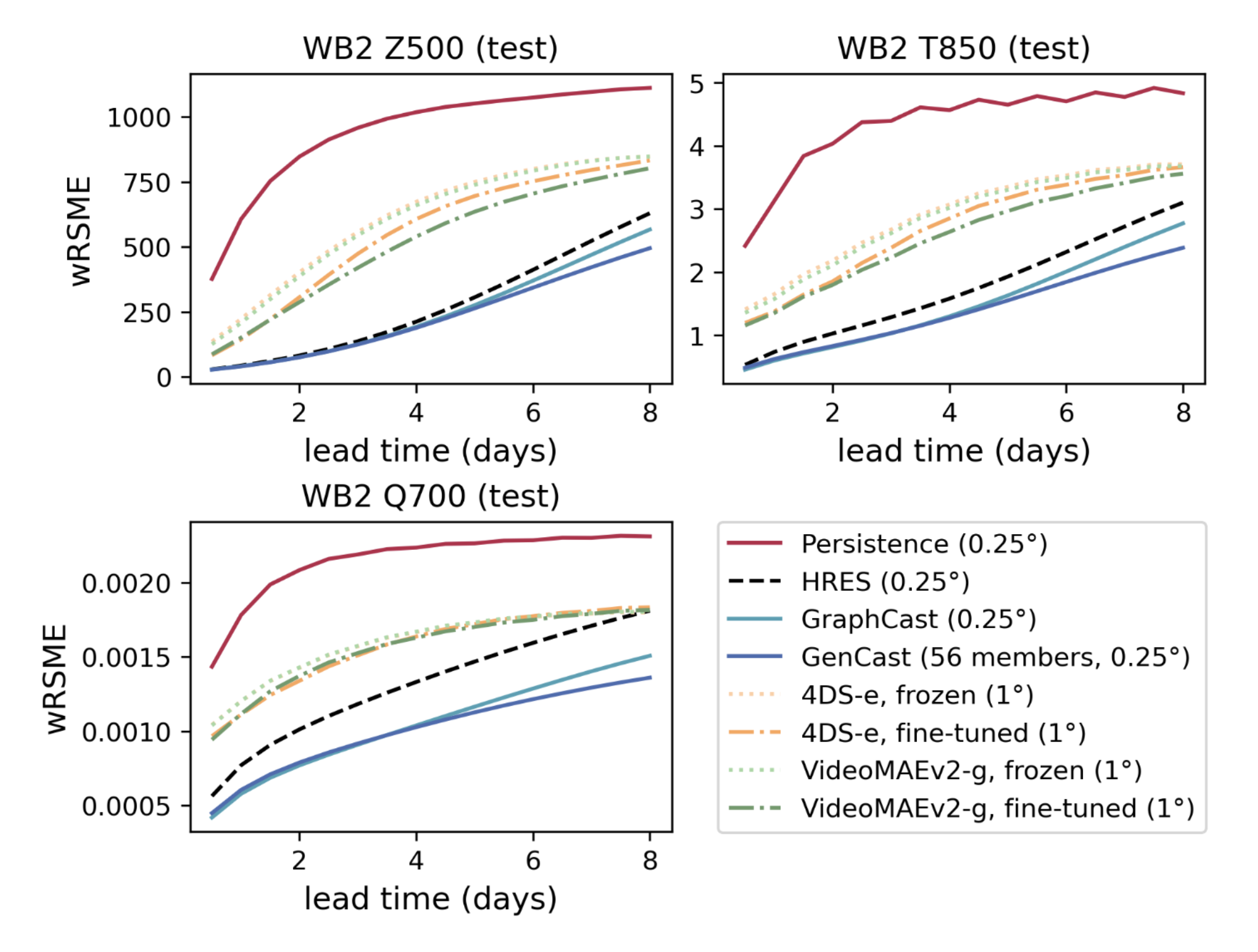}
\vspace{-3mm}
\caption{\textbf{SOTA Comparison Weatherbench 2.} Our readout on top of frozen~(\frozen) 4DS-e and VideoMAEv2-g backbones obtains non-trivial, yet modest performance compared with state-of-the-art approaches (HRES, GraphCast and GenCast). Finetuning the backbones~(\finetuned) helps significantly but an important gap remains.}
\vspace{-2mm}
\label{fig:wb2_sota}
\end{figure}

\tightpara{\weatherbench.} In Fig.~\ref{fig:wb2_sota}, we observe that even the best performing pretrained video models 4DS-e and VideoMAEv2-g obtain relatively modest performance compared with the numerical weather prediction reference HRES and the data-driven baselines GraphCast and GenCast. They do obtain non-trivial performance, as evidenced by the improvement upon the persistence baseline, which simply copies the last input frame. However, this may be mainly due to the readout learning, as suggested by the analysis of optimal readout depth on this benchmark (see Sec.~\ref{subsec:backbone_feature_depth} of the supp.~mat.). 
Finetuning the backbone helps significantly but an important gap remains. 
While we establish initial baselines, further investigation is needed to determine the effectiveness of video backbones for this task -- whether through better pretraining or adaptation techniques.

\begin{figure*}[t!]
\centering
\includegraphics[width=\linewidth]{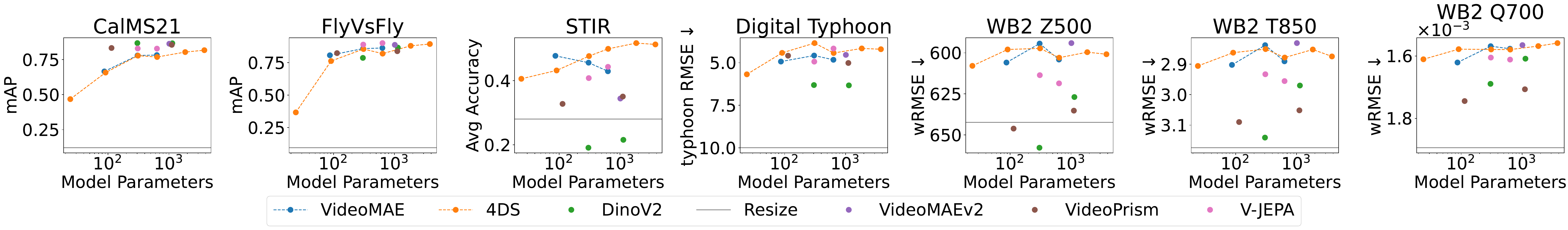}
\vspace{-5mm}
\caption{\textbf{Backbone scaling.} We compare performance for smaller and larger variants of models from different model families, when training readouts on top of frozen backbones~(\frozen). %
While larger variants from a same family tend to perform better on \flyvsfly and \calms, we observe task-dependant changes in performance when increasing the model size on the other tasks.
}
\label{fig:scaling}
\vspace{-3mm}
\end{figure*}

\subsection{Pretrained backbone comparison}
\label{subsec:backbone_comparison}

We show quantitative results for the six selected ViFMs on \scivid in Tab.~\ref{tab:backbone_comparison}.
Video models are generally much stronger than the image baseline, with the exception of Calms21 where DINOv2 performs similarly to VideoPrism and VideoMAEv2-g. Investigating the per-class performance sheds more light on this result: while DinoV2-L outperforms VideoPrism-g on the ``mounting class'' (0.916 vs. 0.823 AP), its performance is notably lower on the ``attacking class'' (0.849 vs 0.885 AP). This disparity likely stems from the nature of the behaviors: attacking is high-intensity, and temporal information is crucial, whereas mounting can often be accurately annotated from single images~\cite{sun2021multi}.
Next, we observe that V-JEPA shines on FlyVsFly where we found that temporal modeling is important (this has also been illustrated on SSV2 in~\cite{carreira2024scaling4drepresentations}).
Also consistent with findings in~\cite{carreira2024scaling4drepresentations}, where the 4DS models outperform other approaches on motion understanding tasks, we find that on STIR, the 4DS-e backbone outperforms all other models by a large margin.
On Digital Typhoon, interestingly, it is 4DS-L which leads, ahead of the larger 4DS-e, with V-JEPA-H following closely behind.
For WeatherBench 2, we find that pure MAE approaches in pixel space (VideoMAE, VideoMAEv2 and 4DS) tend to perform better than other models. DINOv2 lags far behind video-based models, highlighting the importance of temporal modeling.
Finally, we observe that compared with the best performing ViFMs, such as 4DS-e,
the naive feature resizing approach 
drops catastrophically across all tasks, demonstrating the benefits of using pretrained ViFMs in all applications. %
Surprisingly, this baseline outperforms DINOv2 models on \stir tracking, while also forecasting Z500 from \weatherbench more accurately than both DINOv2-L and VideoPrism-B, further illustrating the challenge posed by these transfer settings and some shortcomings of these models on low-level tasks.

\subsection{Ablations}
\label{subsec:ablations}

For our ablations, we select the 4DS-e backbone as it has recently achieved SOTA performance on various \emph{classical} video understanding tasks~\cite{carreira2024scaling4drepresentations}, and the 4DS model family yields strong overall performance on \scivid (see Tab.~\ref{tab:backbone_comparison}).

\tightpara{Backbone scaling.} In Fig.~\ref{fig:scaling}, we explore how scaling the backbone model sizes affects the performance on each of the tasks. %
We observe task- and model-dependent irregularities in performance scaling with backbone size.
While larger models generally yield improvements -- for example, 24M-parameter 4DS-S~\cite{carreira2024scaling4drepresentations} results in a 40\% absolute decrease in mAP on \flyvsfly compared to the its 4B 4DS-e counterpart -- specific tasks, such as \stir with VideoMAE~\cite{tong2022videomae}, reveal instances where smaller variants surpass their larger counterparts. Furthermore, on the \weatherbench Z500 forecasting performance, variations of 4DS model size have limited effect. Rather than relying on presumed scaling laws, empirical evaluation across a range of model sizes allows to identify the optimal configuration for a given task.

\tightpara{Readout architecture.} In Tab.~\ref{tab:readout_comparison}, we ablate our readout design. For \flyvsfly, \calms and \digitaltyphoon, we observe that the cross-attention readout performs significantly better than a linear readout (consisting of a simple average operation followed by a projection). Please refer to Sec.~\ref{subsec:supp_readout_wb_comparison} of the supp.~mat.~for the comparison between DPT and attention-based readouts on \weatherbench.

\begin{table}[H]
\centering
\small
\begin{tabular}{lC{2.2cm}C{2.2cm}}
\toprule
\multirow{2}{*}{Benchmark} & \multicolumn{2}{c}{Readout} \\
\cline{2-3}\\[-0.9em]
   & Linear          & Cross-attention         \\
\midrule
FlyVsFly $\uparrow$ & 0.568 & \textbf{0.894}  \\
\calms $\uparrow$& 0.525 & \textbf{0.817}  \\
\digitaltyphoon $\downarrow$  & 7.45 & \textbf{4.23}  \\
\bottomrule
\end{tabular}
\vspace{-1mm}
\caption{\textbf{Readout architecture}. We compare the performance of different readouts on top of the frozen 4DS-e backbone~(\frozen). For \flyvsfly, \calms and \digitaltyphoon, we observe the \emph{cross-attention} variant performs significantly better than \emph{linear}.}
\vspace{-3mm}
\label{tab:readout_comparison}
\end{table}

\subsection{Qualitative results} 
\label{subsec:qualitative}
We show qualitative results in Fig.~\ref{fig:teaser}. In the bottom left, the \stir example highlights successful tracking despite significant occlusions. In \weatherbench, we visually confirm the quantitative trend of prediction accuracy diminishing with increasing lead times, with this degradation manifesting as visible artifacts. For \digitaltyphoon, we show an illustration of accurate future pressure prediction.  On \flyvsfly and \calms, we show both success and failure cases of handling ambiguous behavioral transitions. More qualitative results can be found in Sec.~\ref{sec:supp_qual_results} of the supp.\ mat.

\section{Conclusion, limitations and future work}
\label{sec:conclusion}
We introduced a new benchmark suite \scivid aimed at evaluating video models in the context of scientific applications. Our suite comprises  highly diverse tasks spanning medical computer vision, animal behavior and weather forecasting. 
We systematically evaluate and compare the performance of leading ViFMs across a wide range of scientific domains within a single, unified framework.
Remarkably, our analysis shows that when adapted with simple trainable readouts, 
ViFMs provide a strong signal in highly diverse spatiotemporal applications and can outperform domain-specific state-of-the-art models in some cases.
We also observe that no single backbone performs best and that there is still a noticeable performance gap on certain generalization scenarios, motivating investigation into truly versatile backbones and better adaptation strategies.

Although our work fills an important gap in evaluating ViFMs for scientific applications,
it has several limitations. 
First, SciVid does not emphasize the development of data-efficient adaptation methods, although we provide a preliminary investigation of performance in low-data regimes in Sec.~\ref{subsec:data_efficiency}.
Collecting datasets with tens of thousands of training samples requires considerable effort, limiting the generalizability of our findings to more data-constrained scenarios. %
Second, while SciVid contains a range of diverse tasks, our benchmark has far-from-comprehensive coverage of domains and it could be extended to, e.g., satellite image times series, underwater video, microscopy imaging etc. We release our code\footnote{\url{https://github.com/google-deepmind/scivid}} and hope that this will empower computer vision and domain specialists to evaluate ViFMs on additional tasks in the future.
Finally, SciVid focuses on short clip evaluation settings (with the exception of STIR, where videos last up to several minutes), whereas certain applications require longer-term modelling. 
One step in this direction would be to develop better adaptation approaches for STIR, beyond naive frame subsampling.

\clearpage
\section*{Acknowledgements}
\label{sec:acknowledgements}

We warmly thank Alvaro Sanchez and Stephan Rasp for providing results for weather forecasting baselines and for sharing valuable insights, as well as Ting Liu and Long Zhao for their help in evaluating VideoPrism.
We also thank Phoebe Kirk for her help on open-sourcing and Ramona Merhej, 
Mehdi Bennani, Jean Bastien Grill, Chuhan Zhang, Rahma Chaabouni and Nikhil Narayanan, Quham Adefila for their contributions at the start of the project.
We further thank Andrew El-Kadi, Ilan Price, Remi Lam, Guillaume Couairon, Armand Joulin, Tengda Han, Cordelia Schmid, Evan Shelhamer, Chris Brown, Drew Purves and Claire Monteleoni for additional discussions and inputs.
We additionally thank Laurel Prince for helping us with presentation material.
Finally, we thank Viorica Patraucean, Joao Carreira, Dima Damen, Joelle Barral and Raia Hadsell for giving valuable feedback on earlier versions of this paper and throughout the project.

{
    \small
    \bibliographystyle{ieeenat_fullname}
    \bibliography{main}

\begin{thebibliography}{72}
\providecommand{\natexlab}[1]{#1}
\providecommand{\url}[1]{\texttt{#1}}
\expandafter\ifx\csname urlstyle\endcsname\relax
  \providecommand{\doi}[1]{doi: #1}\else
  \providecommand{\doi}{doi: \begingroup \urlstyle{rm}\Url}\fi

\bibitem[STI(2024)]{STIRChallenge}
Stir challenge 2024.
\newblock \url{https://www.synapse.org/Synapse:syn54126082/wiki/626617}, 2024.
\newblock Accessed: November 7, 2024.

\bibitem[Alayrac et~al.(2022)Alayrac, Donahue, Luc, Miech, Barr, Hasson, Lenc,
  Mensch, Millican, Reynolds, et~al.]{alayrac2022flamingo}
Jean-Baptiste Alayrac, Jeff Donahue, Pauline Luc, Antoine Miech, Iain Barr,
  Yana Hasson, Karel Lenc, Arthur Mensch, Katherine Millican, Malcolm Reynolds,
  et~al.
\newblock Flamingo: a visual language model for few-shot learning.
\newblock \emph{NeurIPS}, 2022.

\bibitem[Arnab et~al.(2021)Arnab, Dehghani, Heigold, Sun, Lučić, and
  Schmid]{vivit}
Anurag Arnab, Mostafa Dehghani, Georg Heigold, Chen Sun, Mario Lučić, and
  Cordelia Schmid.
\newblock Vivit: A video vision transformer.
\newblock In \emph{ICCV}, 2021.

\bibitem[Awais et~al.(2025)Awais, Naseer, Khan, Anwer, Cholakkal, Shah, Yang,
  and Khan]{awais2025foundation}
Muhammad Awais, Muzammal Naseer, Salman Khan, Rao~Muhammad Anwer, Hisham
  Cholakkal, Mubarak Shah, Ming-Hsuan Yang, and Fahad~Shahbaz Khan.
\newblock Foundation models defining a new era in vision: a survey and outlook.
\newblock \emph{PAMI}, 2025.

\bibitem[Baharoon et~al.(2023)Baharoon, Qureshi, Ouyang, Xu, Aljouie, and
  Peng]{baharoon2023evaluating}
Mohammed Baharoon, Waseem Qureshi, Jiahong Ouyang, Yanwu Xu, Abdulrhman
  Aljouie, and Wei Peng.
\newblock Evaluating general purpose vision foundation models for medical image
  analysis: An experimental study of dinov2 on radiology benchmarks.
\newblock \emph{arXiv preprint arXiv:2312.02366}, 2023.

\bibitem[Bain et~al.(2021)Bain, Nagrani, Schofield, Berdugo, Bessa, Owen,
  Hockings, Matsuzawa, Hayashi, Biro, Carvalho, and Zisserman]{Bain_primates}
Max Bain, Arsha Nagrani, Daniel Schofield, Sophie Berdugo, Joana Bessa, Jake
  Owen, Kimberley Hockings, Tetsuro Matsuzawa, Misato Hayashi, Dora Biro,
  Susana Carvalho, and Andrew Zisserman.
\newblock Automated audiovisual behavior recognition in wild primates.
\newblock \emph{Science Advances}, 2021.

\bibitem[Bardes et~al.(2024)Bardes, Garrido, Ponce, Chen, Rabbat, LeCun,
  Assran, and Ballas]{bardes2023vjepa}
Adrien Bardes, Quentin Garrido, Jean Ponce, Xinlei Chen, Michael Rabbat, Yann
  LeCun, Mido Assran, and Nicolas Ballas.
\newblock Revisiting feature prediction for learning visual representations
  from video.
\newblock \emph{TMLR}, 2024.

\bibitem[Bi et~al.(2023)Bi, Xie, Zhang, Chen, Gu, and Tian]{bi2023pangu}
Kaifeng Bi, Lingxi Xie, Hengheng Zhang, Xin Chen, Xiaotao Gu, and Qi Tian.
\newblock Accurate medium-range global weather forecasting with 3d neural
  networks.
\newblock \emph{Nature}, 2023.

\bibitem[Bodnar et~al.(2024)Bodnar, Bruinsma, Lucic, Stanley, Vaughan,
  Brandstetter, Garvan, Riechert, Weyn, Dong, Gupta, Thambiratnam, Archibald,
  Wu, Heider, Welling, Turner, and Perdikaris]{bodnar2024aurora}
Cristian Bodnar, Wessel~P. Bruinsma, Ana Lucic, Megan Stanley, Anna Vaughan,
  Johannes Brandstetter, Patrick Garvan, Maik Riechert, Jonathan~A. Weyn, Haiyu
  Dong, Jayesh~K. Gupta, Kit Thambiratnam, Alexander~T. Archibald, Chun-Chieh
  Wu, Elizabeth Heider, Max Welling, Richard~E. Turner, and Paris Perdikaris.
\newblock Aurora: A foundation model for the earth systemaurora: A foundation
  model for the earth system.
\newblock \emph{arXiv preprint arXiv:2405.13063}, 2024.

\bibitem[Bommasani et~al.(2021)Bommasani, Hudson, Adeli, Altman, Arora, von
  Arx, Bernstein, Bohg, Bosselut, Brunskill, Brynjolfsson, Buch, Card,
  Castellon, Chatterji, Chen, Creel, Davis, Demszky, Donahue, Doumbouya,
  Durmus, Ermon, Etchemendy, Ethayarajh, Fei-Fei, Finn, Gale, Gillespie, Goel,
  Goodman, Grossman, Guha, Hashimoto, Henderson, Hewitt, Ho, Hong, Hsu, Huang,
  Icard, Jain, Jurafsky, Kalluri, Karamcheti, Keeling, Khani, Khattab, Koh,
  Krass, Krishna, Kuditipudi, Kumar, Ladhak, Lee, Lee, Leskovec, Levent, Li,
  Li, Ma, Malik, Manning, Mirchandani, Mitchell, Munyikwa, Nair, Narayan,
  Narayanan, Newman, Nie, Niebles, Nilforoshan, Nyarko, Ogut, Orr,
  Papadimitriou, Park, Piech, Portelance, Potts, Raghunathan, Reich, Ren, Rong,
  Roohani, Ruiz, Ryan, R'e, Sadigh, Sagawa, Santhanam, Shih, Srinivasan,
  Tamkin, Taori, Thomas, Tram{\`e}r, Wang, Wang, Wu, Wu, Wu, Xie, Yasunaga,
  You, Zaharia, Zhang, Zhang, Zhang, Zhang, Zheng, Zhou, and
  Liang]{Bommasani2021FoundationModels}
Rishi Bommasani, Drew~A. Hudson, Ehsan Adeli, Russ Altman, Simran Arora, Sydney
  von Arx, Michael~S. Bernstein, Jeannette Bohg, Antoine Bosselut, Emma
  Brunskill, Erik Brynjolfsson, S. Buch, Dallas Card, Rodrigo Castellon,
  Niladri~S. Chatterji, Annie~S. Chen, Kathleen~A. Creel, Jared Davis, Dora
  Demszky, Chris Donahue, Moussa Doumbouya, Esin Durmus, Stefano Ermon, John
  Etchemendy, Kawin Ethayarajh, Li Fei-Fei, Chelsea Finn, Trevor Gale,
  Lauren~E. Gillespie, Karan Goel, Noah~D. Goodman, Shelby Grossman, Neel Guha,
  Tatsunori Hashimoto, Peter Henderson, John Hewitt, Daniel~E. Ho, Jenny Hong,
  Kyle Hsu, Jing Huang, Thomas~F. Icard, Saahil Jain, Dan Jurafsky, Pratyusha
  Kalluri, Siddharth Karamcheti, Geoff Keeling, Fereshte Khani, O. Khattab,
  Pang~Wei Koh, Mark~S. Krass, Ranjay Krishna, Rohith Kuditipudi, Ananya Kumar,
  Faisal Ladhak, Mina Lee, Tony Lee, Jure Leskovec, Isabelle Levent, Xiang~Lisa
  Li, Xuechen Li, Tengyu Ma, Ali Malik, Christopher~D. Manning, Suvir~P.
  Mirchandani, Eric Mitchell, Zanele Munyikwa, Suraj Nair, Avanika Narayan,
  Deepak Narayanan, Benjamin Newman, Allen Nie, Juan~Carlos Niebles, Hamed
  Nilforoshan, J.~F. Nyarko, Giray Ogut, Laurel Orr, Isabel Papadimitriou,
  Joon~Sung Park, Chris Piech, Eva Portelance, Christopher Potts, Aditi
  Raghunathan, Robert Reich, Hongyu Ren, Frieda Rong, Yusuf~H. Roohani, Camilo
  Ruiz, Jack Ryan, Christopher R'e, Dorsa Sadigh, Shiori Sagawa, Keshav
  Santhanam, Andy Shih, Krishna~Parasuram Srinivasan, Alex Tamkin, Rohan Taori,
  Armin~W. Thomas, Florian Tram{\`e}r, Rose~E. Wang, William Wang, Bohan Wu,
  Jiajun Wu, Yuhuai Wu, Sang~Michael Xie, Michihiro Yasunaga, Jiaxuan You,
  Matei~A. Zaharia, Michael Zhang, Tianyi Zhang, Xikun Zhang, Yuhui Zhang,
  Lucia Zheng, Kaitlyn Zhou, and Percy Liang.
\newblock On the opportunities and risks of foundation models.
\newblock \emph{arXiv preprint arXiv:2108.07258}, 2021.

\bibitem[Carreira et~al.(2024)Carreira, Gokay, King, Zhang, Rocco, Mahendran,
  Keck, Heyward, Koppula, Pot, Erdogan, Hasson, Yang, Greff, Moing, van
  Steenkiste, Zoran, Hudson, Vélez, Polanía, Friedman, Duvarney, Goroshin,
  Allen, Walker, Kabra, Aboussouan, Sun, Kipf, Doersch, Pătrăucean, Damen,
  Luc, Sajjadi, and Zisserman]{carreira2024scaling4drepresentations}
João Carreira, Dilara Gokay, Michael King, Chuhan Zhang, Ignacio Rocco,
  Aravindh Mahendran, Thomas~Albert Keck, Joseph Heyward, Skanda Koppula,
  Etienne Pot, Goker Erdogan, Yana Hasson, Yi Yang, Klaus Greff, Guillaume~Le
  Moing, Sjoerd van Steenkiste, Daniel Zoran, Drew~A. Hudson, Pedro Vélez,
  Luisa Polanía, Luke Friedman, Chris Duvarney, Ross Goroshin, Kelsey Allen,
  Jacob Walker, Rishabh Kabra, Eric Aboussouan, Jennifer Sun, Thomas Kipf, Carl
  Doersch, Viorica Pătrăucean, Dima Damen, Pauline Luc, Mehdi S.~M. Sajjadi,
  and Andrew Zisserman.
\newblock Scaling 4d representations.
\newblock \emph{arXiv preprint arXiv:2412.15212}, 2024.

\bibitem[Chavas et~al.(2017)Chavas, Reed, and Knaff]{chavas2017physical}
Daniel~R Chavas, Kevin~A Reed, and John~A Knaff.
\newblock Physical understanding of the tropical cyclone wind-pressure
  relationship.
\newblock \emph{Nature Communications}, 2017.

\bibitem[Chen et~al.(2023{\natexlab{a}})Chen, Zhong, Zhang, Cheng, Xu, Qi, and
  Li]{chen2023cfuxi}
Lei Chen, Xiaohui Zhong, Feng Zhang, Yuan Cheng, Yinghui Xu, Yuan Qi, and Hao
  Li.
\newblock Fuxi: A cascade machine learning forecasting system for 15-day global
  weather forecast.
\newblock \emph{npj Climate and Atmospheric Science}, 2023{\natexlab{a}}.

\bibitem[Chen et~al.(2023{\natexlab{b}})Chen, Long, Jiang, Liu, and
  Zhang]{chen2023foundation}
Shengchao Chen, Guodong Long, Jing Jiang, Dikai Liu, and Chengqi Zhang.
\newblock Foundation models for weather and climate data understanding: A
  comprehensive survey.
\newblock \emph{arXiv preprint arXiv:2312.03014}, 2023{\natexlab{b}}.

\bibitem[Christensen et~al.(2024)Christensen, Vukadinovic, Yuan, and
  Ouyang]{christensen2024vision}
Matthew Christensen, Milos Vukadinovic, Neal Yuan, and David Ouyang.
\newblock Vision--language foundation model for echocardiogram interpretation.
\newblock \emph{Nature Medicine}, 2024.

\bibitem[Doersch et~al.(2022)Doersch, Gupta, Markeeva, Continente, Smaira,
  Aytar, Carreira, Zisserman, and Yang]{doersch2022tapvid}
Carl Doersch, Ankush Gupta, Larisa Markeeva, Adria~Recasens Continente, Lucas
  Smaira, Yusuf Aytar, Joao Carreira, Andrew Zisserman, and Yi Yang.
\newblock {TAP}-vid: A benchmark for tracking any point in a video.
\newblock In \emph{NeurIPS}, 2022.

\bibitem[Dubey et~al.(2024)Dubey, Jauhri, Pandey, Kadian, Al-Dahle, Letman,
  Mathur, Schelten, Yang, Fan, et~al.]{dubey2024llama}
Abhimanyu Dubey, Abhinav Jauhri, Abhinav Pandey, Abhishek Kadian, Ahmad
  Al-Dahle, Aiesha Letman, Akhil Mathur, Alan Schelten, Amy Yang, Angela Fan,
  et~al.
\newblock The llama 3 herd of models.
\newblock \emph{arXiv preprint arXiv:2407.21783}, 2024.

\bibitem[Esteves et~al.(2023)Esteves, Slotine, and
  Makadia]{esteves2023sphericalcnn}
Carlos Esteves, Jean-Jacques Slotine, and Ameesh Makadia.
\newblock Scaling spherical cnns.
\newblock \emph{arXiv preprint arXiv:2306.05420}, 2023.

\bibitem[Eyjolfsdottir et~al.(2014)Eyjolfsdottir, Branson, Burgos-Artizzu,
  Hoopfer, Schor, Anderson, and Perona]{eyjolfsdottir2014detecting}
Eyrun Eyjolfsdottir, Steve Branson, Xavier~P Burgos-Artizzu, Eric~D Hoopfer,
  Jonathan Schor, David~J Anderson, and Pietro Perona.
\newblock Detecting social actions of fruit flies.
\newblock In \emph{ECCV}, 2014.

\bibitem[Fibaek et~al.(2024)Fibaek, Camilleri, Luyts, Dionelis, and
  Le~Saux]{fibaek2024phileo}
Casper Fibaek, Luke Camilleri, Andreas Luyts, Nikolaos Dionelis, and Bertrand
  Le~Saux.
\newblock {PhilEO bench: Evaluating geo-spatial foundation models}.
\newblock In \emph{International Geoscience and Remote Sensing Symposium},
  2024.

\bibitem[Goodge et~al.(2025)Goodge, Ng, Hooi, and Ng]{goodge2025spatio}
Adam Goodge, Wee~Siong Ng, Bryan Hooi, and See~Kiong Ng.
\newblock Spatio-temporal foundation models: Vision, challenges, and
  opportunities.
\newblock \emph{arXiv preprint arXiv:2501.09045}, 2025.

\bibitem[Greff et~al.(2022)Greff, Belletti, Beyer, Doersch, Du, Duckworth,
  Fleet, Gnanapragasam, Golemo, Herrmann, Kipf, Kundu, Lagun, Laradji, Liu,
  Meyer, Miao, Nowrouzezahrai, Oztireli, Pot, Radwan, Rebain, Sabour, Sajjadi,
  Sela, Sitzmann, Stone, Sun, Vora, Wang, Wu, Yi, Zhong, and
  Tagliasacchi]{Greff_2022_CVPR}
Klaus Greff, Francois Belletti, Lucas Beyer, Carl Doersch, Yilun Du, Daniel
  Duckworth, David~J. Fleet, Dan Gnanapragasam, Florian Golemo, Charles
  Herrmann, Thomas Kipf, Abhijit Kundu, Dmitry Lagun, Issam Laradji,
  Hsueh-Ti~(Derek) Liu, Henning Meyer, Yishu Miao, Derek Nowrouzezahrai, Cengiz
  Oztireli, Etienne Pot, Noha Radwan, Daniel Rebain, Sara Sabour, Mehdi S.~M.
  Sajjadi, Matan Sela, Vincent Sitzmann, Austin Stone, Deqing Sun, Suhani Vora,
  Ziyu Wang, Tianhao Wu, Kwang~Moo Yi, Fangcheng Zhong, and Andrea
  Tagliasacchi.
\newblock Kubric: A scalable dataset generator.
\newblock In \emph{CVPR}, 2022.

\bibitem[Hirsch et~al.(2023)Hirsch, Caron, Cohen, Livne, Shapiro, Golany,
  Goldenberg, Freedman, and Rivlin]{caron_endoscopy}
Roy Hirsch, Mathilde Caron, Regev Cohen, Amir Livne, Ron Shapiro, Tomer Golany,
  Roman Goldenberg, Daniel Freedman, and Ehud Rivlin.
\newblock Self-supervised learning for endoscopic video analysis.
\newblock In \emph{MICCAI}, 2023.

\bibitem[Hu et~al.(2022)]{hu2022lora_rebuttal}
Edward~J Hu et~al.
\newblock Lo{RA}: Low-rank adaptation of large language models.
\newblock In \emph{ICLR}, 2022.

\bibitem[Jiang et~al.(2024)Jiang, Wang, and Xu]{jiang2024foundation}
Zhe Jiang, Yu Wang, and Zelin Xu.
\newblock Foundation models for spatiotemporal tasks in the physical world.
\newblock In \emph{SIAM International Conference on Data Mining}, 2024.

\bibitem[Jing et~al.(2024)Jing, Zhang, Liang, Li, He, Ma, and
  Guo]{jing2024animal}
Yinuo Jing, Ruxu Zhang, Kongming Liang, Yongxiang Li, Zhongjiang He, Zhanyu Ma,
  and Jun Guo.
\newblock Animal-bench: Benchmarking multimodal video models for animal-centric
  video understanding.
\newblock \emph{NeuRIPS}, 2024.

\bibitem[Kay et~al.(2017)Kay, Carreira, Simonyan, Zhang, Hillier,
  Vijayanarasimhan, Viola, Green, Back, Natsev, Suleyman, and
  Zisserman]{kay2017kineticshumanactionvideo}
Will Kay, Joao Carreira, Karen Simonyan, Brian Zhang, Chloe Hillier, Sudheendra
  Vijayanarasimhan, Fabio Viola, Tim Green, Trevor Back, Paul Natsev, Mustafa
  Suleyman, and Andrew Zisserman.
\newblock The kinetics human action video dataset.
\newblock \emph{arXiv preprint arXiv:1705.06950}, 2017.

\bibitem[Keisler(2022)]{keisler2022gnn}
Ryan Keisler.
\newblock Forecasting global weather with graph neural networks.
\newblock \emph{arXiv preprint arXiv:2202.07575}, 2022.

\bibitem[Kitamoto et~al.(2023)Kitamoto, Hwang, Vuillod, Gautier, Tian, and
  Clanuwat]{digital_typhoon_NEURIPS2023}
Asanobu Kitamoto, Jared Hwang, Bastien Vuillod, Lucas Gautier, Yingtao Tian,
  and Tarin Clanuwat.
\newblock Digital typhoon: Long-term satellite image dataset for the
  spatio-temporal modeling of tropical cyclones.
\newblock In \emph{NeurIPS}, 2023.

\bibitem[Kochkov et~al.(2024)Kochkov, Yuval, Langmore, Norgaard, Smith, Mooers,
  Kl{\"o}wer, Lottes, Rasp, D{\"u}ben, et~al.]{kochkov2023neuralgcm}
Dmitrii Kochkov, Janni Yuval, Ian Langmore, Peter Norgaard, Jamie Smith,
  Griffin Mooers, Milan Kl{\"o}wer, James Lottes, Stephan Rasp, Peter
  D{\"u}ben, et~al.
\newblock Neural general circulation models for weather and climate.
\newblock \emph{Nature}, 2024.

\bibitem[Kurth et~al.(2023)Kurth, Subramanian, Harrington, Pathak, Mardani,
  Hall, Miele, Kashinath, and Anandkumar]{kurth2023fourcastnet}
Thorsten Kurth, Shashank Subramanian, Peter Harrington, Jaideep Pathak, Morteza
  Mardani, David Hall, Andrea Miele, Karthik Kashinath, and Anima Anandkumar.
\newblock Fourcastnet: Accelerating global high-resolution weather forecasting
  using adaptive fourier neural operators.
\newblock In \emph{Proceedings of the platform for advanced scientific
  computing conference}, 2023.

\bibitem[Lam et~al.(2023)Lam, Sanchez-Gonzalez, Willson, Wirnsberger,
  Fortunato, Alet, Ravuri, Ewalds, Eaton-Rosen, Hu, Merose, Hoyer, Holland,
  Vinyals, Stott, Pritzel, Mohamed, and Battaglia]{lam2023graphcast}
Remi Lam, Alvaro Sanchez-Gonzalez, Matthew Willson, Peter Wirnsberger, Meire
  Fortunato, Ferran Alet, Suman Ravuri, Timo Ewalds, Zach Eaton-Rosen, Weihua
  Hu, Alexander Merose, Stephan Hoyer, George Holland, Orial Vinyals, Jacklynn
  Stott, Alexander Pritzel, Shakir Mohamed, and Peter Battaglia.
\newblock Learning skillful medium-range global weather forecasting.
\newblock \emph{Science}, 2023.

\bibitem[Li et~al.(2024)Li, Huang, Wang, Li, and Wang]{li2024videoeval}
Xinhao Li, Zhenpeng Huang, Jing Wang, Kunchang Li, and Limin Wang.
\newblock Video{E}val: Comprehensive benchmark suite for low-cost evaluation of
  video foundation model.
\newblock \emph{arXiv preprint arXiv:2407.06491}, 2024.

\bibitem[Lin et~al.(2023)Lin, Ye, Zhu, Cui, Ning, Jin, and Yuan]{lin2023video}
Bin Lin, Yang Ye, Bin Zhu, Jiaxi Cui, Munan Ning, Peng Jin, and Li Yuan.
\newblock Video-llava: Learning united visual representation by alignment
  before projection.
\newblock \emph{arXiv preprint arXiv:2311.10122}, 2023.

\bibitem[Luke{\v{z}}i{\v{c}} et~al.(2018)Luke{\v{z}}i{\v{c}}, Voj{'i}{\v{r}},
  {\v{C}}ehovin~Zajc, Matas, and Kristan]{Lukezic_IJCV2018}
Alan Luke{\v{z}}i{\v{c}}, Tom{'a}{\v{s}} Voj{'i}{\v{r}}, Luka
  {\v{C}}ehovin~Zajc, Ji{\v{r}}{'i} Matas, and Matej Kristan.
\newblock Discriminative correlation filter tracker with channel and spatial
  reliability.
\newblock \emph{IJCV}, 2018.

\bibitem[Maaz et~al.(2023)Maaz, Rasheed, Khan, and Khan]{maaz2023video}
Muhammad Maaz, Hanoona Rasheed, Salman Khan, and Fahad~Shahbaz Khan.
\newblock Video-chatgpt: Towards detailed video understanding via large vision
  and language models.
\newblock \emph{arXiv preprint arXiv:2306.05424}, 2023.

\bibitem[Madan et~al.(2024)Madan, M{\o}gelmose, Modi, Rawat, and
  Moeslund]{madan2024foundation}
Neelu Madan, Andreas M{\o}gelmose, Rajat Modi, Yogesh~S Rawat, and Thomas~B
  Moeslund.
\newblock Foundation models for video understanding: A survey.
\newblock \emph{arXiv preprint arXiv:2405.03770}, 2024.

\bibitem[Martinez(2020)]{econometrics8020018}
Andrew~B. Martinez.
\newblock Forecast accuracy matters for hurricane damage.
\newblock \emph{Econometrics}, 2020.

\bibitem[Mukkavilli et~al.(2023)Mukkavilli, Civitarese, Schmude, Jakubik,
  Jones, Nguyen, Phillips, Roy, Singh, Watson, et~al.]{mukkavilli2023ai}
S~Karthik Mukkavilli, Daniel~Salles Civitarese, Johannes Schmude, Johannes
  Jakubik, Anne Jones, Nam Nguyen, Christopher Phillips, Sujit Roy, Shraddha
  Singh, Campbell Watson, et~al.
\newblock Ai foundation models for weather and climate: Applications, design,
  and implementation.
\newblock \emph{arXiv preprint arXiv:2309.10808}, 2023.

\bibitem[Neidlinger et~al.(2024)Neidlinger, El~Nahhas, Muti, Lenz, Hoffmeister,
  Brenner, van Treeck, Langer, Dislich, Behrens,
  et~al.]{neidlinger2024benchmarking}
Peter Neidlinger, Omar~SM El~Nahhas, Hannah~Sophie Muti, Tim Lenz, Michael
  Hoffmeister, Hermann Brenner, Marko van Treeck, Rupert Langer, Bastian
  Dislich, Hans~Michael Behrens, et~al.
\newblock Benchmarking foundation models as feature extractors for
  weakly-supervised computational pathology.
\newblock \emph{arXiv preprint arXiv:2408.15823}, 2024.

\bibitem[Neoral et~al.(2024)Neoral, {\v{S}}er{\`y}ch, and Matas]{neoral2024mft}
Michal Neoral, Jon{\'a}{\v{s}} {\v{S}}er{\`y}ch, and Ji{\v{r}}{\'\i} Matas.
\newblock {MFT}: Long-term tracking of every pixel.
\newblock In \emph{WACV}, 2024.

\bibitem[{NeurIPS 2024 FM4Science Workshop}(2024)]{neurips2024fm4science}
{NeurIPS 2024 FM4Science Workshop}.
\newblock Foundation models for science: Progress, opportunities, and
  challenges (fm4science).
\newblock NeurIPS 2024 Workshop, 2024.

\bibitem[Nguyen et~al.(2023)Nguyen, Brandstetter, Kapoor, Gupta, and
  Grover]{nguyen2023climax}
Tung Nguyen, Johannes Brandstetter, Ashish Kapoor, Jayesh~K Gupta, and Aditya
  Grover.
\newblock Climax: A foundation model for weather and climate.
\newblock \emph{arXiv preprint arXiv:2301.10343}, 2023.

\bibitem[Oquab et~al.(2024)Oquab, Darcet, Moutakanni, Vo, Szafraniec, Khalidov,
  Fernandez, HAZIZA, Massa, El-Nouby, Assran, Ballas, Galuba, Howes, Huang, Li,
  Misra, Rabbat, Sharma, Synnaeve, Xu, Jegou, Mairal, Labatut, Joulin, and
  Bojanowski]{oquab2023dinov2}
Maxime Oquab, Timoth{\'e}e Darcet, Th{\'e}o Moutakanni, Huy~V. Vo, Marc
  Szafraniec, Vasil Khalidov, Pierre Fernandez, Daniel HAZIZA, Francisco Massa,
  Alaaeldin El-Nouby, Mido Assran, Nicolas Ballas, Wojciech Galuba, Russell
  Howes, Po-Yao Huang, Shang-Wen Li, Ishan Misra, Michael Rabbat, Vasu Sharma,
  Gabriel Synnaeve, Hu Xu, Herve Jegou, Julien Mairal, Patrick Labatut, Armand
  Joulin, and Piotr Bojanowski.
\newblock {DINO}v2: Learning robust visual features without supervision.
\newblock \emph{TMLR}, 2024.

\bibitem[Price et~al.(2025)Price, Sanchez-Gonzalez, Alet, Andersson, El-Kadi,
  Masters, Ewalds, Stott, Mohamed, Battaglia, et~al.]{price2025gencast}
Ilan Price, Alvaro Sanchez-Gonzalez, Ferran Alet, Tom~R Andersson, Andrew
  El-Kadi, Dominic Masters, Timo Ewalds, Jacklynn Stott, Shakir Mohamed, Peter
  Battaglia, et~al.
\newblock Probabilistic weather forecasting with machine learning.
\newblock \emph{Nature}, 2025.

\bibitem[Radford et~al.(2021)Radford, Kim, Hallacy, Ramesh, Goh, Agarwal,
  Sastry, Askell, Mishkin, Clark, et~al.]{radford2021clip}
Alec Radford, Jong~Wook Kim, Chris Hallacy, Aditya Ramesh, Gabriel Goh,
  Sandhini Agarwal, Girish Sastry, Amanda Askell, Pamela Mishkin, Jack Clark,
  et~al.
\newblock Learning transferable visual models from natural language
  supervision.
\newblock In \emph{ICML}, 2021.

\bibitem[Ranftl et~al.(2021)Ranftl, Bochkovskiy, and Koltun]{Ranftl2021}
Ren{\'e} Ranftl, Alexey Bochkovskiy, and Vladlen Koltun.
\newblock Vision transformers for dense prediction.
\newblock In \emph{ICCV}, 2021.

\bibitem[Rasp et~al.(2024)Rasp, Hoyer, Merose, Langmore, Battaglia, Russell,
  Sanchez-Gonzalez, Yang, Carver, Agrawal, et~al.]{rasp2024wb2}
Stephan Rasp, Stephan Hoyer, Alexander Merose, Ian Langmore, Peter Battaglia,
  Tyler Russell, Alvaro Sanchez-Gonzalez, Vivian Yang, Rob Carver, Shreya
  Agrawal, et~al.
\newblock Weatherbench 2: A benchmark for the next generation of data-driven
  global weather models.
\newblock \emph{Journal of Advances in Modeling Earth Systems}, 2024.

\bibitem[Rolnick et~al.(2022)Rolnick, Donti, Kaack, Kochanski, Lacoste,
  Sankaran, Ross, Milojevic-Dupont, Jaques, Waldman-Brown, Luccioni, Maharaj,
  Sherwin, Mukkavilli, Kording, Gomes, Ng, Hassabis, Platt, Creutzig, Chayes,
  and Bengio]{rolnick_2023_tackling_climate_change_ai}
David Rolnick, Priya~L. Donti, Lynn~H. Kaack, Kelly Kochanski, Alexandre
  Lacoste, Kris Sankaran, Andrew~Slavin Ross, Nikola Milojevic-Dupont, Natasha
  Jaques, Anna Waldman-Brown, Alexandra~Sasha Luccioni, Tegan Maharaj, Evan~D.
  Sherwin, S.~Karthik Mukkavilli, Konrad~P. Kording, Carla~P. Gomes, Andrew~Y.
  Ng, Demis Hassabis, John~C. Platt, Felix Creutzig, Jennifer Chayes, and
  Yoshua Bengio.
\newblock Tackling climate change with machine learning.
\newblock \emph{ACM Comput. Surv.}, 2022.

\bibitem[Schmidgall et~al.(2024)Schmidgall, Kim, Jopling, and
  Krieger]{schmidgall2024general}
Samuel Schmidgall, Ji~Woong Kim, Jeffrey Jopling, and Axel Krieger.
\newblock General surgery vision transformer: A video pre-trained foundation
  model for general surgery.
\newblock \emph{arXiv preprint arXiv:2403.05949}, 2024.

\bibitem[Schmidt et~al.(2024)Schmidt, Mohareri, DiMaio, and
  Salcudean]{stir_Schmidt_2024}
Adam Schmidt, Omid Mohareri, Simon~P. DiMaio, and Septimiu~E. Salcudean.
\newblock Surgical tattoos in infrared: A dataset for quantifying tissue
  tracking and mapping.
\newblock \emph{IEEE Transactions on Medical Imaging}, 2024.

\bibitem[Shi et~al.(2025)Shi, Shirali, Jin, Zhou, Hu, Rangaraj, Wang, Han,
  Wang, Lall, et~al.]{shi2025deep}
Jimeng Shi, Azam Shirali, Bowen Jin, Sizhe Zhou, Wei Hu, Rahuul Rangaraj,
  Shaowen Wang, Jiawei Han, Zhaonan Wang, Upmanu Lall, et~al.
\newblock Deep learning and foundation models for weather prediction: A survey.
\newblock \emph{arXiv preprint arXiv:2501.06907}, 2025.

\bibitem[Smith and Katz(2013)]{hurricane}
Adam Smith and Richard Katz.
\newblock Us billion-dollar weather and climate disasters: Data sources,
  trends, accuracy and biases.
\newblock \emph{Natural Hazards}, 2013.

\bibitem[Sun(2025)]{sun2025toward}
Jennifer~J Sun.
\newblock Toward collaborative artificial intelligence development for animal
  well-being.
\newblock \emph{Journal of the American Veterinary Medical Association}, 2025.

\bibitem[Sun et~al.(2021{\natexlab{a}})Sun, Karigo, Chakraborty, Mohanty, Wild,
  Sun, Chen, Anderson, Perona, Yue, et~al.]{sun2021multi}
Jennifer~J Sun, Tomomi Karigo, Dipam Chakraborty, Sharada~P Mohanty, Benjamin
  Wild, Quan Sun, Chen Chen, David~J Anderson, Pietro Perona, Yisong Yue,
  et~al.
\newblock The multi-agent behavior dataset: Mouse dyadic social interactions.
\newblock In \emph{NeurIPS}, 2021{\natexlab{a}}.

\bibitem[Sun et~al.(2021{\natexlab{b}})Sun, Kennedy, Zhan, Anderson, Yue, and
  Perona]{Sun_2021_CVPR_task_programming}
Jennifer~J. Sun, Ann Kennedy, Eric Zhan, David~J. Anderson, Yisong Yue, and
  Pietro Perona.
\newblock Task programming: Learning data efficient behavior representations.
\newblock In \emph{CVPR}, 2021{\natexlab{b}}.

\bibitem[Sun et~al.(2024)Sun, Zhou, Zhao, Yuan, Seybold, Hendon, Schroff, Ross,
  Adam, Hu, et~al.]{sun2024video}
Jennifer~J Sun, Hao Zhou, Long Zhao, Liangzhe Yuan, Bryan Seybold, David
  Hendon, Florian Schroff, David~A Ross, Hartwig Adam, Bo Hu, et~al.
\newblock Video foundation models for animal behavior analysis.
\newblock \emph{bioRxiv}, 2024.

\bibitem[Team et~al.(2024)Team, Anil, Borgeaud, Alayrac, Yu, Soricut,
  Schalkwyk, Dai, Hauth, Millican, et~al.]{team2024gemini}
Gemini Team, Rohan Anil, Sebastian Borgeaud, Jean-Baptiste Alayrac, Jiahui Yu,
  Radu Soricut, Johan Schalkwyk, Andrew~M Dai, Anja Hauth, Katie Millican,
  et~al.
\newblock Gemini: a family of highly capable multimodal models.
\newblock \emph{arXiv preprint arXiv:2312.11805}, 2024.

\bibitem[Teed and Deng(2020)]{teed2020raft}
Zachary Teed and Jia Deng.
\newblock Raft: Recurrent all-pairs field transforms for optical flow.
\newblock In \emph{ECCV}, 2020.

\bibitem[Tong et~al.(2022)Tong, Song, Wang, and Wang]{tong2022videomae}
Zhan Tong, Yibing Song, Jue Wang, and Limin Wang.
\newblock Videomae: Masked autoencoders are data-efficient learners for
  self-supervised video pre-training.
\newblock \emph{NeurIPS}, 2022.

\bibitem[Wang et~al.(2023{\natexlab{a}})Wang, Fu, Du, Gao, Huang, Liu, Chandak,
  Liu, Katwyk, Deac, Anandkumar, Bergen, Gomes, Ho, Kohli, Lasenby, Leskovec,
  Liu, Manrai, Marks, Ramsundar, Song, Sun, Tang, Velickovic, Welling, Zhang,
  Coley, Bengio, and Zitnik]{Wang2023ScientificDI}
Hanchen Wang, Tianfan Fu, Yuanqi Du, Wenhao Gao, Kexin Huang, Ziming Liu, Payal
  Chandak, Shengchao Liu, Peter~Van Katwyk, Andreea Deac, Anima Anandkumar,
  Karianne~J. Bergen, Carla~P. Gomes, Shirley Ho, Pushmeet Kohli, Joan Lasenby,
  Jure Leskovec, Tie-Yan Liu, Arjun~K. Manrai, Debora~S. Marks, Bharath
  Ramsundar, Le Song, Jimeng Sun, Jian Tang, Petar Velickovic, Max Welling,
  Linfeng Zhang, Connor~W. Coley, Yoshua Bengio, and Marinka Zitnik.
\newblock Scientific discovery in the age of artificial intelligence.
\newblock \emph{Nature}, 2023{\natexlab{a}}.

\bibitem[Wang et~al.(2023{\natexlab{b}})Wang, Huang, Zhao, Tong, He, Wang,
  Wang, and Qiao]{wang2023videomae}
Limin Wang, Bingkun Huang, Zhiyu Zhao, Zhan Tong, Yinan He, Yi Wang, Yali Wang,
  and Yu Qiao.
\newblock Videomae v2: Scaling video masked autoencoders with dual masking.
\newblock In \emph{CVPR}, 2023{\natexlab{b}}.

\bibitem[Wang et~al.(2022)Wang, Li, Li, He, Huang, Zhao, Zhang, Xu, Liu, Wang,
  et~al.]{wang2022internvideo}
Yi Wang, Kunchang Li, Yizhuo Li, Yinan He, Bingkun Huang, Zhiyu Zhao, Hongjie
  Zhang, Jilan Xu, Yi Liu, Zun Wang, et~al.
\newblock Internvideo: General video foundation models via generative and
  discriminative learning.
\newblock \emph{arXiv preprint arXiv:2212.03191}, 2022.

\bibitem[Wang et~al.(2023{\natexlab{c}})Wang, Liu, Zhang, and
  Dou]{wang2023foundation}
Zhao Wang, Chang Liu, Shaoting Zhang, and Qi Dou.
\newblock Foundation model for endoscopy video analysis via large-scale
  self-supervised pre-train.
\newblock In \emph{International Conference on Medical Image Computing and
  Computer-Assisted Intervention}. Springer, 2023{\natexlab{c}}.

\bibitem[Yan et~al.(2024)Yan, Yu, Primiero, Vico-Alonso, Wang, Yang, Tschandl,
  Hu, Tan, Tang, et~al.]{yan2024general}
Siyuan Yan, Zhen Yu, Clare Primiero, Cristina Vico-Alonso, Zhonghua Wang, Litao
  Yang, Philipp Tschandl, Ming Hu, Gin Tan, Vincent Tang, et~al.
\newblock A general-purpose multimodal foundation model for dermatology.
\newblock \emph{arXiv preprint arXiv:2410.15038}, 2024.

\bibitem[Yang et~al.(2024)Yang, Zeng, Zhang, and Zhang]{yang2024x}
Jie Yang, Ailing Zeng, Ruimao Zhang, and Lei Zhang.
\newblock X-pose: Detecting any keypoints.
\newblock In \emph{ECCV}, 2024.

\bibitem[Ye et~al.(2024)Ye, Filippova, Lauer, Schneider, Vidal, Qiu, Mathis,
  and Mathis]{ye2024superanimal}
Shaokai Ye, Anastasiia Filippova, Jessy Lauer, Steffen Schneider, Maxime Vidal,
  Tian Qiu, Alexander Mathis, and Mackenzie~Weygandt Mathis.
\newblock Superanimal pretrained pose estimation models for behavioral
  analysis.
\newblock \emph{Nature communications}, 2024.

\bibitem[Yuan et~al.(2023)Yuan, Gundavarapu, Zhao, Zhou, Cui, Jiang, Yang, Jia,
  Weyand, Friedman, et~al.]{yuan2023videoglue}
Liangzhe Yuan, Nitesh~Bharadwaj Gundavarapu, Long Zhao, Hao Zhou, Yin Cui, Lu
  Jiang, Xuan Yang, Menglin Jia, Tobias Weyand, Luke Friedman, et~al.
\newblock Videoglue: Video general understanding evaluation of foundation
  models.
\newblock \emph{arXiv preprint arXiv:2307.03166}, 2023.

\bibitem[Zhang et~al.(2024)Zhang, Zhou, Adhikarla, Yan, Liu, Yu, Liu, Chen,
  Davison, Ren, et~al.]{zhang2024generalist}
Kai Zhang, Rong Zhou, Eashan Adhikarla, Zhiling Yan, Yixin Liu, Jun Yu,
  Zhengliang Liu, Xun Chen, Brian~D Davison, Hui Ren, et~al.
\newblock A generalist vision--language foundation model for diverse biomedical
  tasks.
\newblock \emph{Nature Medicine}, 2024.

\bibitem[Zhao et~al.(2024{\natexlab{a}})Zhao, Gundavarapu, Yuan, Zhou, Yan,
  Sun, Friedman, Qian, Weyand, Zhao, et~al.]{zhaovideoprism}
Long Zhao, Nitesh~Bharadwaj Gundavarapu, Liangzhe Yuan, Hao Zhou, Shen Yan,
  Jennifer~J Sun, Luke Friedman, Rui Qian, Tobias Weyand, Yue Zhao, et~al.
\newblock Videoprism: A foundational visual encoder for video understanding.
\newblock In \emph{ICML}, 2024{\natexlab{a}}.

\bibitem[Zhao et~al.(2024{\natexlab{b}})Zhao, Gu, Yang, Usuyama, Lee, Naumann,
  Gao, Crabtree, Abel, Moung-Wen, et~al.]{zhao2024biomedparse}
Theodore Zhao, Yu Gu, Jianwei Yang, Naoto Usuyama, Ho~Hin Lee, Tristan Naumann,
  Jianfeng Gao, Angela Crabtree, Jacob Abel, Christine Moung-Wen, et~al.
\newblock Biomedparse: a biomedical foundation model for image parsing of
  everything everywhere all at once.
\newblock \emph{arXiv preprint arXiv:2405.12971}, 2024{\natexlab{b}}.

\bibitem[Zhao et~al.(2023)Zhao, Liu, Wu, Wang, Li, Wang, Teng, Liu, Cui, Wang,
  et~al.]{zhao2023clip}
Zihao Zhao, Yuxiao Liu, Han Wu, Mei Wang, Yonghao Li, Sheng Wang, Lin Teng,
  Disheng Liu, Zhiming Cui, Qian Wang, et~al.
\newblock Clip in medical imaging: A comprehensive survey.
\newblock \emph{arXiv preprint arXiv:2312.07353}, 2023.

\end{thebibliography}
}

\clearpage
\setcounter{page}{1}
\maketitlesupplementary
\appendix

\renewcommand{\thefigure}{A.\arabic{figure}} %
\setcounter{figure}{0} 
\renewcommand{\thetable}{A.\arabic{table}}
\setcounter{table}{0} 

In this supplementary material, we provide additional information on baselines (Sec.~\ref{sec:supp_baselines}), implementation details (Sec.~\ref{sec:supp_impl_details}), further experiments (Sec.~\ref{sec:supp_experiments}), and qualitative results (Sec.~\ref{sec:supp_qual_results}). We also encourage the reader to watch our narrated video, \textcolor{blue}{5334.mp4}. Please refer to the main paper for bibliography references.

\startcontents[sections]
{
	\hypersetup{linkcolor=black}
	\printcontents[sections]{l}{1}{}
}

\section{Baselines}\label{sec:supp_baselines}

\paragraph{\textbf{Weather forecasting.}} We compare our best models to the following weather forecasting baselines:
\begin{itemize}
    \item Eulerian Persistence is commonly used in the weather forecasting literature. It produces constant forecasts, where each frame is equal to the last input frame (also called the initial condition).
    \item IFS HRES is the main baseline considered when evaluating deterministic data-driven models. It comes from operational forecasts created with ECMWF's IFS model, typically regarded as the best global, medium-range weather forecasts. Following GraphCast, for the validation set, we evaluate forecasts initialized at 06 and 18 UTC, to ensure equal look ahead (of 3h) for HRES and data-driven models. However, as HRES forecasts are only run for up to a lead time of 3.75 days, we use HRES forecasts initialized at 00 and 12 UTC (which are run for 10 days) for longer lead times. For the test set, following Weatherbench 2~\cite{rasp2024wb2}, we ignore the difference of look ahead times and use forecasts initialized at 00 and 12 UTC for all methods, including HRES. Furthermore, following common practice, we evaluate IFS forecasts against operational analyses as ground truth, rather than ERA5, to avoid putting IFS at an unfair disadvantage compared to data-driven models.
    \item GraphCast~\cite{lam2023graphcast} is a state-of-the-art deterministic model for weather forecasting. For the validation set (2018), we use a version trained from 1979 to 2017 (included); whereas for the test set (2020), we use a version trained from 1979 to 2019 (included).
    \item GenCast~\cite{price2025gencast} further improves upon GraphCast with a probabilistic diffusion-based approach. Like for GraphCast, in each setting, we use a version trained up until the year before the evaluation set. 50 members are used for the ensemble on the validation set, and 56 members on the test set. EDA initialization is used on the validation set and normal initial conditions are used when running on the test set. Due to the important inference cost involved in evaluating GenCast, we leave these small inconsistencies unaddressed, as they have been observed in practice to have negligible impact on performance.
\end{itemize}

Results for these baselines were communicated to us privately by the GenCast and Weatherbench 2 authors. Each model is \emph{evaluated} at its native \emph{training} resolution,  i.e.~0.25°. In practice, we found that when \emph{evaluating} on 1° resolution the model \emph{trained} at 0.25° resolution, the results are indistinguishable from the ones we report. When  \emph{training} and \emph{evaluating} on 1° resolution, performance for GraphCast and GenCast models are very similar to 0.25° results, except for very short lead times.
For additional details on the evaluation on IFS HRES and GraphCast baselines, we refer the reader to the Weatherbench 2 paper~\cite{rasp2024wb2}. For additional details on GenCast, we refer the reader to~\cite{price2025gencast}.

\renewcommand{\thefigure}{B.\arabic{figure}}
\setcounter{figure}{0} 
\renewcommand{\thetable}{B.\arabic{table}}
\setcounter{table}{0} 
\section{Implementation details}\label{sec:supp_impl_details}

In this section, we describe our readouts (Sec.~\ref{subsec:supp_readouts}) and evaluation metrics for each task (Sec.~\ref{subsec:supp_evaluation_metrics}) as well as our \weatherbench conditioning setup (Sec.~\ref{subsec:weatherbench_conditioning}). 

\subsection{Readout design and training details}\label{subsec:supp_readouts}

We train all readouts with a batch size of 32. For weather forecasting where we observe overfitting for time steps above 10k in most experimental setups. Therefore, unless explicitly mentioned, models for \flyvsfly, \calms, \stir and \digitaltyphoon are trained for 40k steps following~\cite{carreira2024scaling4drepresentations} while models trained on \weatherbench are trained for 10k steps.
We use the AdamW optimizer, with weight decay of $1e^{-4}$, a learning rate cosine schedule from $3e^{-4}$ to $1e^{-7}$, and a linear learning rate warmup for 1k steps. Please refer to Tab.~\ref{table:readout_modules} for details on readout architectures.

\noindent \textbf{Classification.} The readout module consists of 16 heads, with 64 and 8 feature channels per head for \calms and \flyvsfly respectively. We also apply data augmentations on the input videos which we found beneficial. In particular, we resize \flyvsfly and \calms images by a factor of 1.4 and 2 times the model input resolution respectively and randomly extract a crop of the model input resolution.
Note that weight decay is set to 0 for \flyvsfly only and that for \calms, we additionally apply random left-right flipping as well as Gaussian blurring with a kernel size of 36 during clip preprocessing. 

\noindent \textbf{Tracking.} The readout module consists of 8 heads, with 128 feature channels per head. We trained it on 16-frame windows randomly selected from the 24-frame videos in the Kubric MOVi-E dataset~\cite{Greff_2022_CVPR}, which features static and dynamic objects rendered on photorealistic backgrounds. Supervision was derived by sampling 64 point tracks per window, prioritizing foreground objects. 

\noindent \textbf{Weather forecasting.} We use a standard four-stage DPT design, with all intermediate feature dimensions of 1024. We refer the reader to~\cite{Ranftl2021} for in-depth details on the DPT architecture. We use gradient clipping of 1 which we found to prevent training instabilities. To account for the fact that different channels have values of different orders of magnitude, we follow~\cite{lam2023graphcast} by weighting the loss terms by the inverse standard deviation of the time difference residuals. As input to the readout, we provide the backbone features corresponding to the last time step. We observed that additionally using features from other prior frames to be detrimental to the results. As opposed to the original DPT implementation~\cite{Ranftl2021}, which inputs different features depths at different upsampling stages, we provide features from a single backbone level to all upsampling stages, as we found this to work better in our setting. We further find it beneficial to remove the penultimate convolution layer and increase the intermediate feature dimensions.

\noindent \textbf{Pressure forecasting.} The readout module consists of 16 heads, with 64 feature channels per head. The evaluation is performed on the first 24 frames of the typhoon sequence. While the original training~\cite{digital_typhoon_NEURIPS2023} uses only the first 24 frames of the sequence, we empirically find that randomly shifting the start of the 24-frame sequence between frames 0 to 8 improves performance. Our readout regresses the difference to the average train pressure, of 983.9 hPa, computed across the train set. 

\begin{table}[h]
\centering
\small
\begin{tabular}{c|l}
\hline
\textbf{Eval} & \textbf{Architecture} \\ \hline
FlyVsFly & 
\begin{tabular}[c]{@{}l@{}}
\texttt{CrossAttention(} \\
\texttt{\quad qkv\_size=128,} \\
\texttt{\quad num\_heads=16)} \\
\\ 
\texttt{Linear(output\_size=7)}
\end{tabular}  \\ \hline
\calms & 
\begin{tabular}[c]{@{}l@{}}
\texttt{CrossAttention(} \\
\texttt{\quad qkv\_size=1024,} \\
\texttt{\quad num\_heads=16)} \\
\\
\texttt{Linear(output\_size=4)}
\end{tabular}  \\ \hline
\stir & 
\begin{tabular}[c]{@{}l@{}}
\texttt{CrossAttention(} \\
\texttt{\quad qkv\_size=1024,} \\
\texttt{\quad num\_heads=8)} \\
\\
\texttt{Linear(output\_size=4)}
\end{tabular}\\ \hline
\weatherbench & 
\begin{tabular}[c]{@{}l@{}}
\texttt{DensePredictionTransformer(} \\
\texttt{\quad num\_blocks=4,} \\
\texttt{\quad feature\_dim=1024)} \\
\\
\texttt{Conv(} \\
\texttt{\quad kernel\_size=3 $\times$ 3,} \\
\texttt{\quad feature\_dim=512)} \\
\\
\texttt{Conv(} \\
\texttt{\quad kernel\_size=1 $\times$ 1,} \\
\texttt{\quad output\_size=48 (16 $\times$ 3))} \\
\end{tabular}  \\ \hline
\digitaltyphoon & 
\begin{tabular}[c]{@{}l@{}}
\texttt{CrossAttention(} \\
\texttt{\quad qkv\_size=1024,} \\
\texttt{\quad num\_heads=16)} \\
\\
\texttt{Linear(output\_size=12)}
\end{tabular}\\ \hline
\end{tabular}
\caption{Architectures of readout modules for different tasks. For Weatherbench 2, \texttt{num blocks} refers to the number of reassamble and fusion operations. Please refer to~\cite{Ranftl2021} for further details on DPT components.}
\label{table:readout_modules}
\end{table}

\subsection{Evaluation metrics}\label{subsec:supp_evaluation_metrics}

\paragraph{\textbf{Classification.}} We follow the multilabel average precision implementation from VideoPrism~\cite{zhaovideoprism}.
In particular, the background class is excluded by computing the average precision after removing the \emph{not interacting} results from both the ground truth and predictions. 

\paragraph{\textbf{Tracking.}} We compute the percentage of predicted track endpoints that fall within specific distance thresholds of their corresponding ground truth locations. These thresholds are set at several pixel distances (4, 8, 16, 32, and 64 pixels) to assess accuracy at varying levels of precision. An average accuracy across these thresholds, denoted as $\delta_{avg}^{x}$, is then computed to provide a single, overall measure of tracking performance. 
On the validation and test sets, the query and target points are extracted separately, with no matching available. Following~\cite{STIRChallenge}, a simple closest-target point heuristic is used to assign initial query points to their corresponding final location. The computed accuracy is therefore an estimate of the performance, as some tracked queries and target points can be erroneously matched. 

\paragraph{\textbf{Pressure forecasting.}} The paper introducing \digitaltyphoon~\cite{digital_typhoon_NEURIPS2023} reports root mean squared errors for the following time steps: 1, 2, 3, 6, 12. For comparing model performance using a single value, we compute the error averaged across these time steps. RMSE for \digitaltyphoon is therefore defined as following:

\begin{equation}
\text{RMSE} =  \frac{1}{T=5}  \sum_{t \in \{1, 2, 3, 6, 12\}}  \sqrt{\frac{1}{K} \sum_{k=1}^{K} (f_{k,t} - o_{k,t})^2}
\end{equation}

where $k$ is the sample index among $K$ samples, $f_i$ is the forecasted central pressure at time step $i$ and $o_i$ is the ground truth typhoon central pressure for this time step.

\paragraph{\textbf{Weather forecasting.}} Following common practice~\cite{rasp2024wb2}, we measure performance using the area-weighted root mean squared error (wRMSE).
Area-weighting is used because on an equiangular latitude-longitude grid, grid cells at the poles have a much smaller area compared to grid cells at the equator. 
For a grid cell with latitude index $i$, 
the latitude weight $w(i)$ is computed as:

\begin{equation}
w(i) = \frac{\sin\theta_i^u - \sin\theta_i^l}{\frac{1}{I} \sum_{i}^{I} (\sin\theta_i^u - \sin\theta_i^l)},
\end{equation}

where $\theta_i^u$ and $\theta_i^l$ indicate upper and lower latitude bounds, respectively. For each variable (corresponding to a specific pressure level), the wRMSE is defined as follows:
\begin{equation}
\text{wRMSE} = \sqrt{\frac{1}{TIJ} \sum_t^T \sum_i^I \sum_j^J w(i) (f_{t,i,j} - o_{t,i,j})^2},
\end{equation}

where $f_{t,i,j}$ and $o_{t,i,j}$ are respectively the forecasted and the ground truth values at spatio-temporal position $(t, i, j)$.

\subsection{\weatherbench conditioning}
\label{subsec:weatherbench_conditioning}

When conditioning on all other upper-level variables (\withcond) for the \weatherbench task, we use %
atmospheric variables proposed in GraphCast~\cite{lam2023graphcast}: geopotential, specific humidity, temperature, u component of wind, v component of wind and vertical velocity.
The 13 vertical levels are folded into the channel dimension for atmospheric variables
resulting in 78 channel.
In order to adapt the models to the larger number of channels, we reinitialize the first input layer to accommodate the larger number of channels, using the same random initialization as in the original implementation for VideoMAEv2-g~\cite{wang2023videomae} and 4DS-e~\cite{carreira2024scaling4drepresentations}.
In Sec.~\ref{subsec:weatherbench_withcond} we discuss the results \withcond for which we reinitialize the embedding layer to provide more than three variables as input to these two models.

\renewcommand{\thefigure}{C.\arabic{figure}}
\setcounter{figure}{0} 
\renewcommand{\thetable}{C.\arabic{table}}
\setcounter{table}{0} 
\section{Additional experiments}\label{sec:supp_experiments}

Here, we provide additional results on backbone finetuning (Sec.~\ref{subsec:supp_backbone_finetuning}), efficient backbone low-rank adaptation~\cite{hu2022lora_rebuttal} (Sec.~\ref{subsec:lora}),
data-efficiency (Sec.~\ref{subsec:data_efficiency}),
per-threshold SOTA comparison on \stir (Sec.~\ref{subsec:supp_threshold_stir_comparison}), SOTA comparison for Digital Typhoon (Sec.~\ref{subsec:supp_digital_typhoon_results}), conditioning, readout comparison and prediction strategy on \weatherbench (Sec.~\ref{subsec:weatherbench_withcond},~\ref{subsec:supp_readout_wb_comparison},~\ref{subsec:supp_pred_strategy_wb} ), backbone feature depth (Sec.~\ref{subsec:backbone_feature_depth}), backbone and readout training length (Sec.~\ref{subsec:backbone_training_length},~\ref{subsec:readout_training_length}),
single input frame performance (Sec.~\ref{subsec:temporal_dynamics}) and a noise study of our evaluations (Sec.~\ref{subsec:eval_noise}).

\subsection{Backbone finetuning} \label{subsec:supp_backbone_finetuning}

\begin{figure}[H]
\centering
\includegraphics[width=\linewidth]{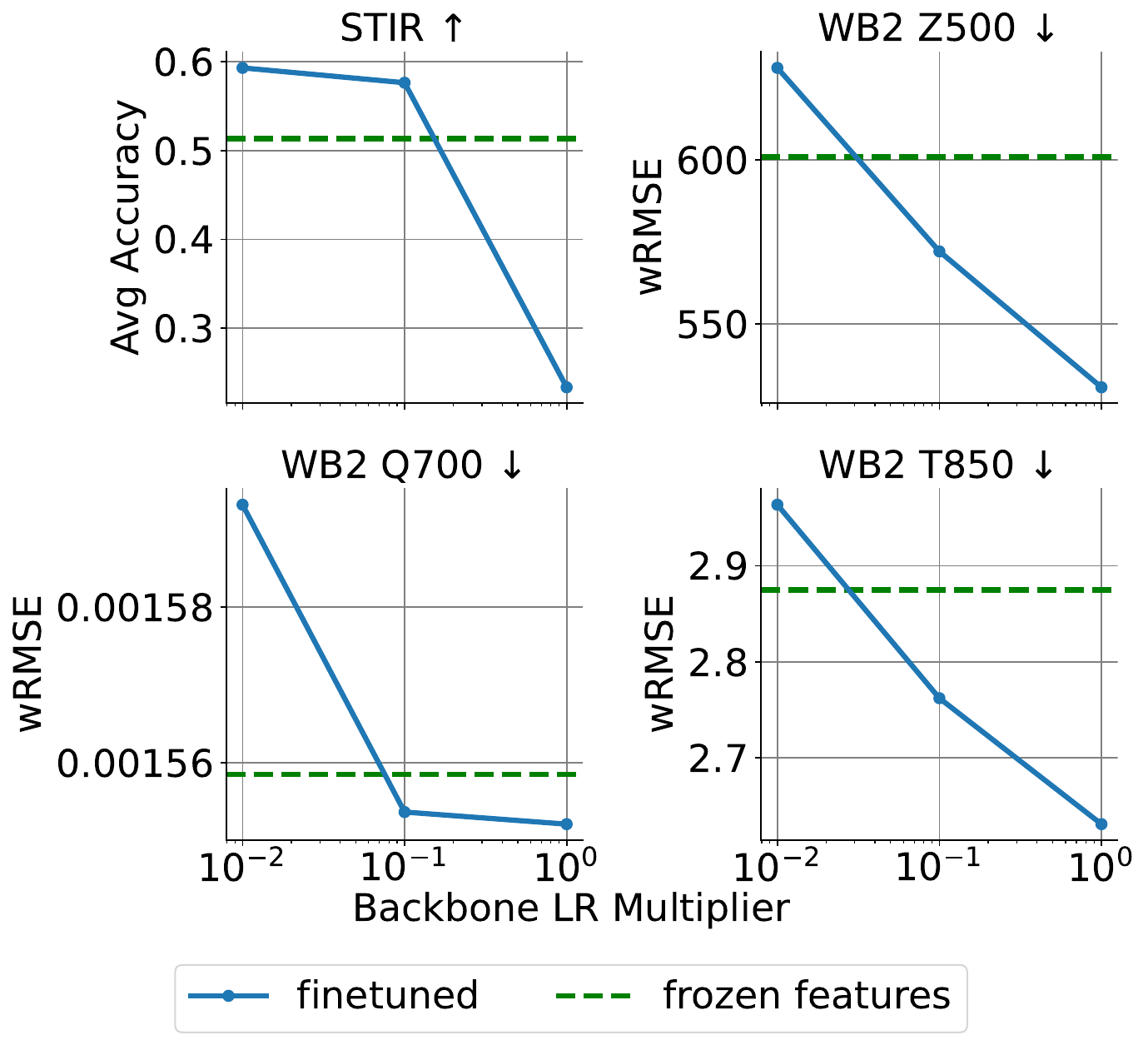}
\caption{\textbf{Backbone finetuning STIR \& WeatherBench 2.} Finetuning the backbone helps improve the final performance for both \weatherbench and \stir provided an appropriate task-dependant learning rate is applied to the backbone. A 100x smaller backbone learning achieves the best performance for STIR while applying the same learning rate to both the readout and the backbone yields the best results for \weatherbench.}

\label{fig:finetuning}
\end{figure}

We reach state-of-the-art performance by training readouts on top of frozen features for three out of five tasks (\calms, \flyvsfly and \digitaltyphoon).
We investigate whether finetuning the backbone improves the final performance on \stir and \weatherbench, the two tasks for which an important gap compared to the state of the art remains.
For this analysis, we finetune the 4DS-e backbone in addition to training the readout, using different fractions of the readout learning (1, 0.1 or 0.01) to account for the fact that the pretrained backbone might benefit from smaller updates, while larger updates could be needed for the readout which is trained from random initializations.
For all other settings, we follow the protocol described in Sec.~\ref{sec:eval} of the main paper and Sec.~\ref{sec:supp_impl_details} of supp. mat.
In both cases we observe that performance can be improved at the additional computation cost of finetuning the backbone.
In Fig.~\ref{fig:finetuning}, we see that updates of the backbone with the same learning rate as the readout yields best results on \weatherbench, which has a large domain gap with the original video input distribution.
Using a learning rate two orders of magnitude smaller for the backbone compared to the readout results in the best performance on \stir.
Variations in the backbone learning rate multiplier incur large variation in performance.
We note that the optimal learning rate for one task (0.01 for \stir, 1 for \weatherbench) yields results that are \textit{worse} than frozen features on the other task, indicating that this parameter needs to be carefully tuned for each task. Finetuning the backbone with appropriate learning rates applied to the backbone updates helps  significantly, but a performance gap compared to the state of the art remains for both tasks, which we detail in Fig.~\ref{fig:stir_model_comparison} for \stir and Fig.~\ref{fig:wb2_sota} for \weatherbench. 

\subsection{Low rank adaptation} \label{subsec:lora}

\begin{figure}[h!]
    \centering
    \includegraphics[width=0.5\textwidth]{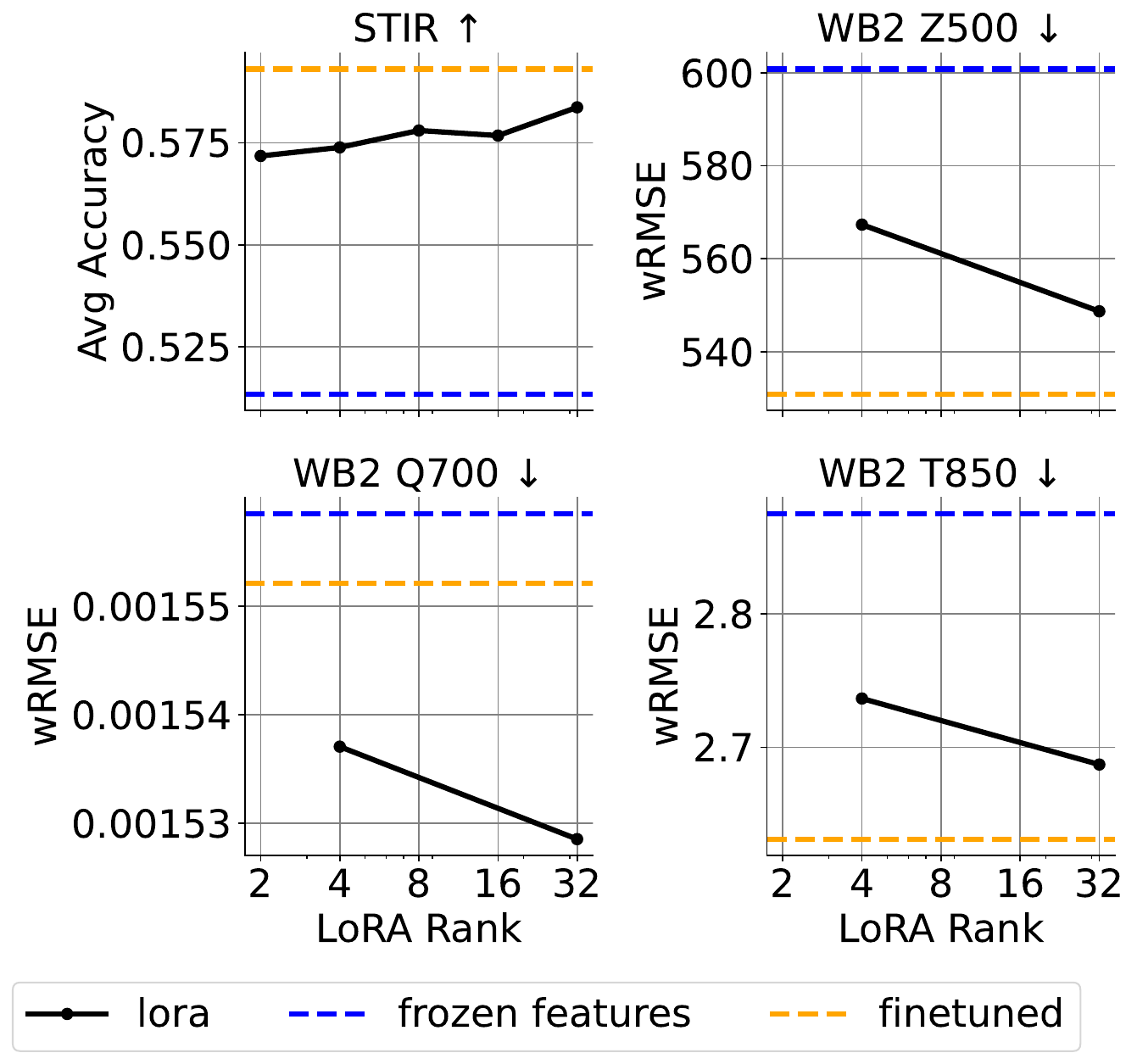} %
    \caption{\textbf{Comparison of different adaptation strategies}. With the 4DS-e backbone, LoRA adaptation with rank (r) 32 approaches the performance of full finetuning at a fraction of the parameter cost.}
    \label{fig:lora_rank}
\end{figure}

To investigate lightweight backbone adaptation techniques, in Fig.~\ref{fig:lora_rank}, we compare three training strategies on \stir and \weatherbench: training the readout with (i) a frozen \frozen 4DS-e backbone, (ii) full backbone finetuning \finetuned and (iii) finetuning with low rank adaptation (LoRA)~\cite{hu2022lora_rebuttal}.
We observe that LoRA finetuning strikes an effective balance and that setting the LoRA rank r to 32 yields some improvement over a more lightweight adapter (r=4).
When setting r to 32, results approaches the full finetuning performanche, while requiring training only 2.3\% parameters compared to the backbone full finetuning (see Tab.~\ref{tab:lora}).

\begin{table}[t!]
\centering
\resizebox{\columnwidth}{!}{%
\small
\begin{tabular}{lcccc}
\toprule
 Benchmark & Frozen \frozen & LoRA r=4 & LoRA r=32 & Finetuned \finetuned \\
 \% of params & 0 & 0.3 & 2.3 & 100 \\
\midrule
\stir $\uparrow$ & 51.3 & 57.4 & \underline{58.4} & \textbf{59.3}  \\
WB2 T850 $\downarrow$ & 2.87 & 2.74 & \underline{2.69} & \textbf{2.63} \\  %
\bottomrule
\end{tabular}
}
\caption{\textbf{Comparison of different adaptations}. With the 4DS-e backbone, LoRA adaptation with rank (r) 32 approaches the performance of full finetuning at a fraction of the parameter cost.}
\label{tab:lora}
\end{table}

\subsection{Data-efficiency}
\label{subsec:data_efficiency}

While SciVid offers a diverse suite of scientific applications, it does not focus on the development of \textit{data-efficient} adaptation methods. Indeed, datasets within \scivid contain tens of thousands of training samples, with the exception of Digital Typhoon which contains 696. 
Collecting such datasets requires considerable effort, %
which might be prohibitive for some applications.

\begin{figure}[h!]
\vspace{-3mm}
    \centering 
    \includegraphics[width=0.5\textwidth,height=2.5cm]{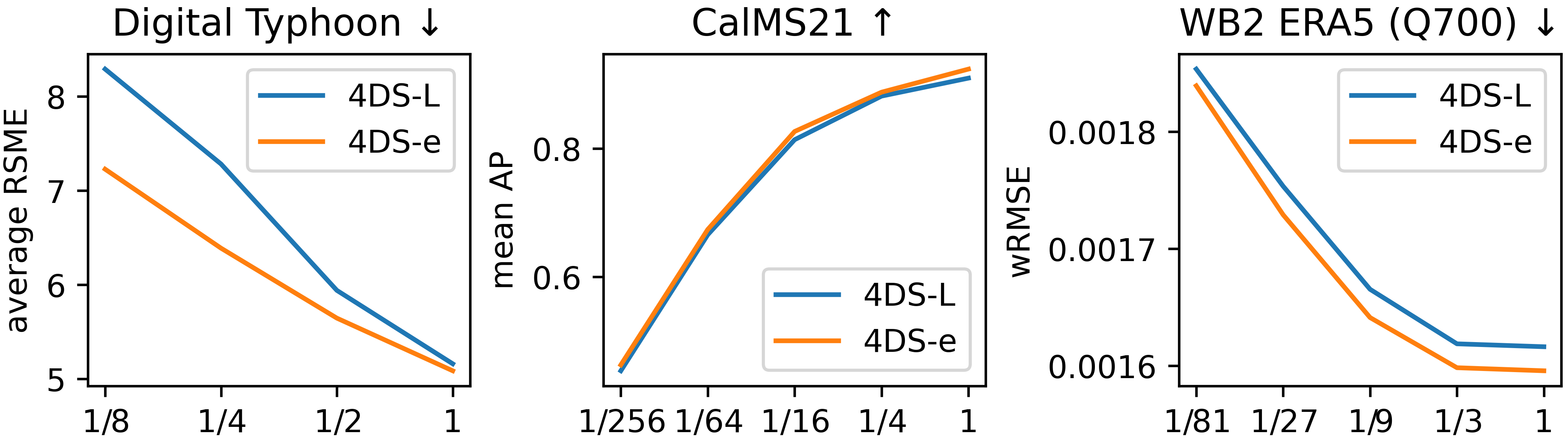} %
    \vspace{-7mm}
    \caption{\textbf{\frozen~Model performance vs.~fraction of training data.}}
    \vspace{-2mm}
    \label{fig:data-efficiency}
\end{figure}

To take a step towards addressing this limitation, we provide initial comparisons of models in data-constrained settings.
Fig.~\ref{fig:data-efficiency} illustrates the relationship between frozen~(\frozen) backbone performance and the fraction of training data used for two models, 4DS-L and 4DS-e.
For Digital Typhoon, we divide the dataset size by powers of 2, with 1/8 corresponding to ~100 samples.  %
For CalMS21, which has 27k training clips,  we divide the dataset size by powers of 4, with 1/256 corresponding to ~100 samples. %
For ERA5, we use powers of 3 to preserve forecast times spanning a full day (00/06/12/18), with 1/81 corresponding to ~700 samples.  %
We observe that 4DS-e, which performs better in the full data regime than 4DS-L, generally maintains its advantage when we reduce the amount of training data.
Finally, this analysis also highlights room for improvement in low data regimes, and provides a more challenging setting that can potentially better differentiate model capabilities.

\begin{table}[]
\centering
\begin{tabular}{llrr}
\toprule
      & fraction &  4DS-L &  4DS-e \\
\midrule
\multirow{4}{*}{\begin{tabular}{@{}c@{}}Digital\\ Typhoon \\ (RMSE ↓) \end{tabular}}  & 1             										   & 5.16	 & 5.09 \\
                & {\large \sfrac{1}{2}}  & 5.94	 & 5.65 \\
                & {\large \sfrac{1}{4}}  & 7.28	 & 6.39 \\
                & {\large \sfrac{1}{8}}  & 8.29	 & 7.23 \\
\midrule
\multirow{5}{*}{\begin{tabular}{@{}c@{}}CalMS21\\ (mAP↑) \end{tabular}}  & 1 
                                            & 91.0	 & 92.4 \\
                & {\large \sfrac{1}{4}}  	& 88.2	 & 88.8 \\
                & {\large \sfrac{1}{16}}  	& 81.4	 & 82.6 \\
                & {\large \sfrac{1}{64}}  	& 66.5	 & 67.4 \\
                & {\large \sfrac{1}{256}}  	& 45.3	 & 46.3 \\
\midrule
\multirow{5}{*}{\begin{tabular}{@{}c@{}}WB2 ERA5\\ Q700\\ (wRMSE ↓) \end{tabular}}  & 1                 					 	& 1.62e-03	 & 1.60e-03 \\
                & {\large \sfrac{1}{3}}     & 1.62e-03	 & 1.60e-03 \\
                & {\large \sfrac{1}{9}}     & 1.67e-03	 & 1.64e-03 \\
                & {\large \sfrac{1}{27}}    & 1.75e-03	 & 1.73e-03 \\
                & {\large \sfrac{1}{81}}    & 1.85e-03	 & 1.84e-03 \\
\bottomrule
\end{tabular}
\caption{\frozen~Model performance vs.~fraction of training data.}
\label{tab:data-efficiency}
\end{table}

\subsection{SOTA comparison on \stir}
\label{subsec:supp_threshold_stir_comparison}
\begin{figure}
\centering
\includegraphics[width=\linewidth]{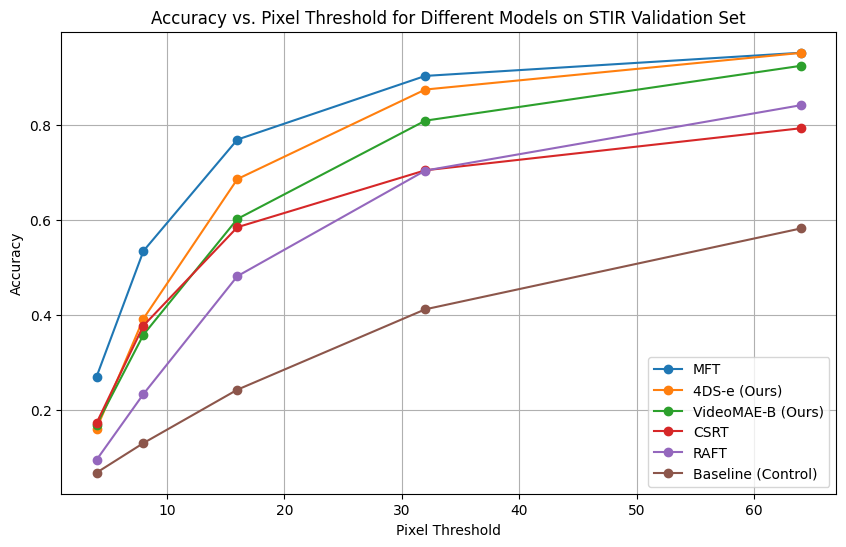}
\caption{\textbf{SOTA comparison on STIR.} We indicate for different models the percentage of predicted track endpoints that  fall  within  specific  distance  thresholds (4, 8, 16, 32, and 64 pixels) of their  corresponding  ground  truth  locations. \textit{Ours} refer to our readout with 4DS-e or VideoMAE-B backbones with finetuning~(\finetuned).}
\label{fig:threshold}
\end{figure}

Fig.~\ref{fig:threshold} illustrates tracking performance on \stir at various threshold distances (4, 8, 16, 32, and 64 pixels).  MFT~\cite{neoral2024mft} surpasses other methods at all thresholds except the largest (64 pixels), where it performs comparably to our readout on 4DS-e with finetuning.  The performance gap between methods are generally consistent across thresholds.  A notable exception is CSRT~\cite{Lukezic_IJCV2018}, which outperforms RAFT~\cite{teed2020raft} at smaller distances but exhibits lower performance at larger distances.

\subsection{SOTA comparison Digital Typhoon}
\begin{figure}[H]
\centering
\small
\includegraphics[width=\linewidth]{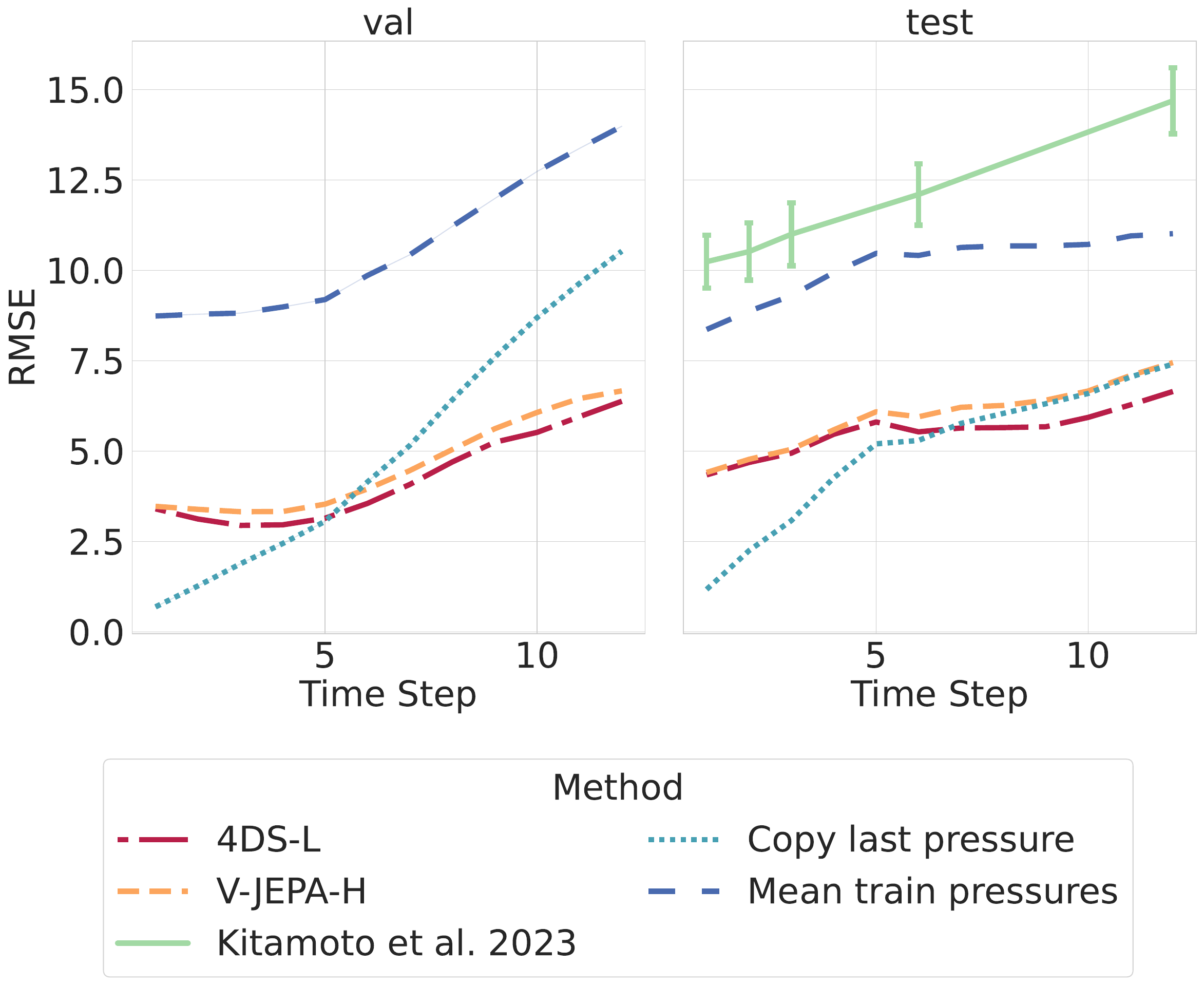}
\caption{\textbf{SOTA comparison Digital Typhoon.} We significantly outperform the method from Kitamoto et al.~\cite{digital_typhoon_NEURIPS2023} (numbers obtained from their paper). While the \emph{copy last pressure} baseline achieves a lower RMSE at smaller time steps, our readout training using frozen features (\frozen) from 4DS-L performs best overall.}
\label{fig:typhoon_sota}
\end{figure}

In Fig.~\ref{fig:typhoon_sota}, we qualitatively compare our readout training with the 4DS-L and V-JEPA-H backbones to the method from~\cite{digital_typhoon_NEURIPS2023}, the \emph{mean train pressure} baseline and the \emph{copy last pressure} oracle. Our method outperforms both~\cite{digital_typhoon_NEURIPS2023} and the \emph{mean train pressure} baseline on both the validation and test sets for all time steps. We observe that the \emph{copy last pressure} oracle, which uses the last input central pressure value, achieves a lower RMSE at smaller time steps,  benefitting from temporal continuity.  At later time steps, our readout  with  the  4DS-e  backbone  consistently outperforms all approaches.

\label{subsec:supp_digital_typhoon_results}

\subsection{\weatherbench \withcond}
\label{subsec:weatherbench_withcond}

We compare finetuning \withcond and \withoutcond (see Sec.~\ref{subsec:weatherbench_conditioning} for more details) for the two backbones which perform best on \weatherbench: 4DS-e and VideoMAEv2-g.
We observe that training \withcond yields small improvements on geopotential forecasting (Z500) while
\withoutcond works better for temperature and specific humidity forecasting for the two backbones.
These initial results highlight the need to develop adaptation approaches which can efficiently leverage the extra input information while efficiently leveraging representations learnt during pretraining.  

\begin{table}[H]
\small
\label{tab:wb_with_cond}
\begin{tabular}{lcccc}
\toprule
Model & \withcond & Z500 ↓ & T850 ↓ & Q700 ↓ \\
\midrule
4DS-e & \xmark & \underline{531} & \underline{2.63} & 1.55e-03 \\
4DS-e & \cmark & 549 & 2.71 & 1.54e-03 \\
VideoMAEv2-g & \xmark & \textbf{512} & \textbf{2.57} & \underline{1.51e-03} \\ %
VideoMAEv2-g & \cmark & 539 & 2.65 & \textbf{1.49e-03} \\
\bottomrule
\end{tabular}
\caption{\textbf{\weatherbench results with and without conditioning.} We compare finetuning (\finetuned) \withcond and \withoutcond for the two strongest backbones on the \weatherbench task on the validation set.}
\end{table}

\subsection{Readout comparison on \weatherbench}
\label{subsec:supp_readout_wb_comparison}
We experimented with a pure attention-based readout for \weatherbench, compared to the DPT-based \cite{Ranftl2021} readout used. We follow the attention readout design in \cite{carreira2024scaling4drepresentations} for depth prediction and adapt it by predicting a 3 channel output for all target frames. The results are shown in Tab.~\ref{tab:wb_attention_dpt_comparison}.  We evaluate the performance both when the backbone model is frozen and when it is finetuned together with the readout. We observe that in both scenarios, the DPT readout outperforms the attention one by a large margin.

\begin{table}[H]
\setlength{\tabcolsep}{5.5pt}
\small{\begin{tabular}{lllll}
\toprule
Model & Readout & Z500 ↓ & T850 ↓ & Q700 ↓ \\
\midrule
 VideoMAEv2-g \frozen & Attn. & 892 & 4.09 &2.09e-03 \\
 VideoMAEv2-g \frozen & DPT  &\underline{594} & \underline{2.82} & \underline{1.56e-03} \\
 VideoMAEv2-g \finetuned &Attn. & 853 & 3.98  &2.09e-03 \\
 VideoMAEv2-g \finetuned & DPT  & \textbf{587} & \textbf{2.79} & \textbf{1.54e-03} \\
\bottomrule
\end{tabular}}
\caption{\textbf{Attention vs.\ DPT-based readout on \weatherbench.} We compare the two readouts for the strongest backbone model on this task. In both the frozen (\frozen) and finetuning (\finetuned) settings, the DPT readout outperforms the attention one by a large margin.}
\label{tab:wb_attention_dpt_comparison}
\end{table}

\subsection{Prediction strategy on \weatherbench}
\label{subsec:supp_pred_strategy_wb}
We ablate the prediction strategy on the \weatherbench by either predicting the target variables directly or residuals with respect to the last input weather state. Results in Tab.~\ref{tab:wb_residual} show that when finetuning \withcond %
predicting the residual yields the best quantitative performance. Interestingly, increasing training steps from 10k to 40k marginally improves the results depending on the variable when predicting residuals but significantly helps in the direct target prediction setting. We posit that residual training allow to reach reasonable performance fast, but limits potential further gains. In contrast, direct target prediction is more challenging, but may allow to reach a superior optimum with sufficient training. We also compare the two approaches qualitatively in Fig~\ref{fig:wb_residual} and observe that residual prediction yields high-frequency artefacts, in particular on the humidity Q700. Direct prediction of the targets removes these artefacts, but tends to over-smooth the results, leading to lower overall accuracy.

\begin{table}[H]
\small
\setlength{\tabcolsep}{5pt}
\begin{tabular}{llllll}
\toprule
Model & Steps & Res. & Z500 ↓ & T850 ↓ & Q700 ↓ \\
\midrule
VideoMAEv2-g & 10k & \xmark & {571} & {2.78} & {1.52e-03} \\
VideoMAEv2-g & 10k & \cmark & \textbf{539} & \underline{2.65} & {1.49e-03} \\
VideoMAEv2-g & 40k & \xmark & {563} & {2.67} & \textbf{1.45e-03} \\
VideoMAEv2-g & 40k & \cmark & \underline{540} & \textbf{2.64} & \underline{1.48e-03} \\
\bottomrule
\end{tabular}
\caption{\textbf{Residual vs.\ direct prediction on \weatherbench.} We compare directly predicting the target variables or a residual (res.) with respect to the last input weather state. We show results for our readout trained on top of the VideoMAEv2-g backbone with finetuning (\finetuned) and \withoutcond. While at short training lengths, predicting the residual yields the best quantitative results, at longer ones, the performance gap is marginal.  }
\label{tab:wb_residual}
\end{table}

\begin{figure}[H]
\centering
\setlength{\tabcolsep}{1pt}
\small
\begin{tabular}{>{\centering\arraybackslash}m{0.25cm}l}
    & ~~~~~ Ground truth ~~~~~~~~~~~~~~~  Residual ~~~~~~~~~~~~~~~~~~~  Direct \\
 \adjustbox{valign=c,rotate=90,raise=12em}{Q700 ~~~~~~ T850 ~~~~~~ Z500} & \includegraphics[width=0.95\linewidth]{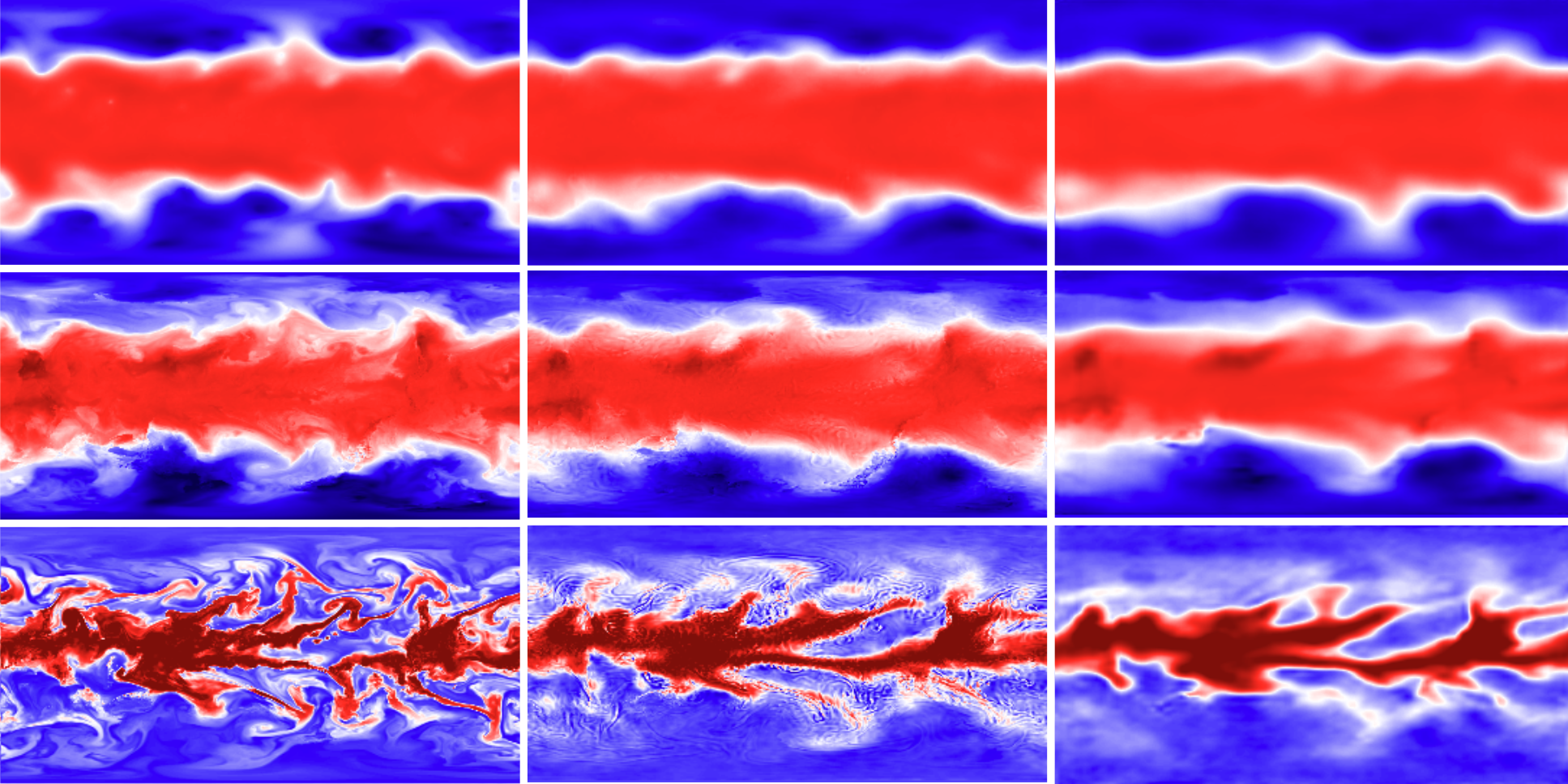} \\
\end{tabular}
\vspace{-10em}
\caption{\textbf{Residual vs.\ direct prediction on \weatherbench.} We compare directly predicting the target variables or a residual with respect to the last input weather state. We show results on the last predicted frame for our readout on top of the VideoMAEv2-g backbone with finetuning (\finetuned) and \withcond. We observe that residual prediction yields high-frequency artefacts, in particular on the humidity Q700. Direct prediction of the target variable removes these artefacts, but potentially leads to over-smoothing, highlighting different failure modes of the two approaches.}
\label{fig:wb_residual}
\end{figure}

\subsection{Backbone feature depth}
\label{subsec:backbone_feature_depth}
\begin{figure*}[!htbp]
\centering
\includegraphics[width=\textwidth]{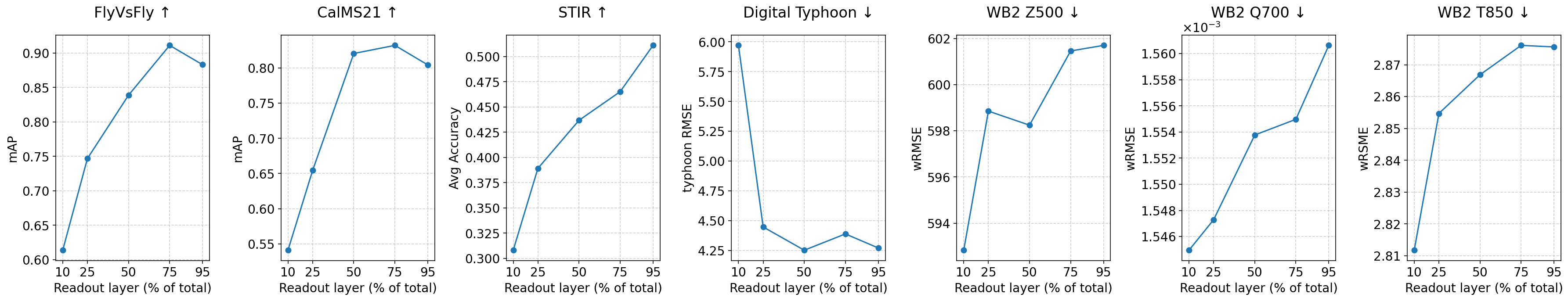}
\caption{\textbf{Feature depth.} We show the performance of the frozen (\frozen) 4DS-e model when we attach the readouts at different layers of the model (as a
percentage of the total number of self-attention blocks, 56 for 4DS-e). We observe that features from the layer at 95\% depth offers a good performance across tasks, similarly to findings in~\cite{carreira2024scaling4drepresentations}, except for \weatherbench where early features perform better.}
\label{fig:feature_depth}
\end{figure*}

We investigate where in the model are the best representations for each task by extracting features at different model depths from the 4DS-e backbone -- 10\%, 25\%, 50\%, 75\% and 95\% (for example, 75\% corresponds to layer 42 out of 56 for 4DS-e). We observe in Fig.~\ref{fig:feature_depth} that the features from the layer at 95\% depth reaches a good compromise across tasks.
One notable exception is \weatherbench where for all forecasted variables, early features perform better, and performance degrades progressively with model depth for each of the three forecasted variables.

\subsection{Backbone training length}
\label{subsec:backbone_training_length}

\begin{figure*}[!htbp]
\centering
\includegraphics[width=\linewidth]{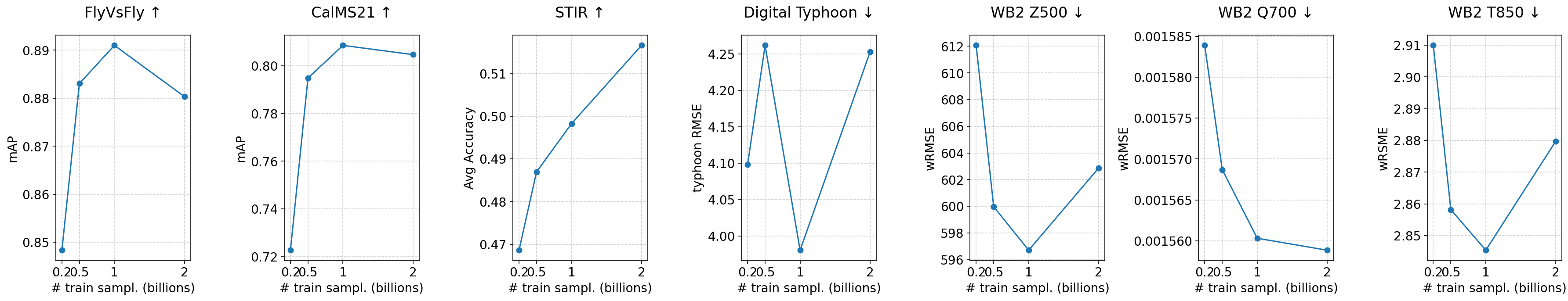}
\caption{\textbf{Backbone training length.} We show the performance of the readout training on top of the frozen 4DS-e model (\frozen), when the backbone has seen more or less examples during pretraining. We observe that longer pretraining helps overall across tasks, with peak performance around 1B seen examples for all tasks, except for \stir and \weatherbench Q700.}
\label{fig:backbone_training_length}
\end{figure*}

In Fig.~\ref{fig:backbone_training_length}, we investigate the importance of pretraining duration, measured by the number of examples the 4DS-e backbone sees during training. We observe that longer pretraining helps  cross tasks, with peak performance around 1B seen examples for most tasks.
On Digital Typhoon, the observed irregular variations are likely caused by the noise levels for this evaluation (see~\ref{subsec:eval_noise} for more details).

\begin{figure*}[!htbp]
\centering
\includegraphics[width=\textwidth]{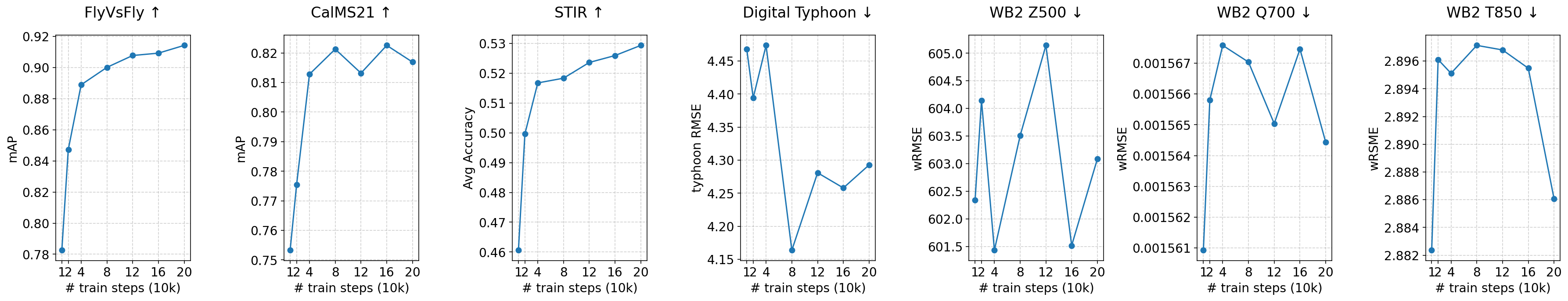}
\caption{\textbf{Readout training steps.} We show the performance of the readout trainings on top of the frozen 4DS-e model (\frozen) when varying the training duration. We observe that longer training helps continuously for STIR and FlyVSFly. For \calms and Digital Typhoon, we observe peak performance around 80K steps. For \weatherbench, the best overall results are at 10k steps.}
\label{fig:readout_training_length}
\end{figure*}

\subsection{Readout training length}
\label{subsec:readout_training_length}

We investigate whether training the readouts for longer improves performance across tasks. In Fig.~\ref{fig:readout_training_length}, we observe that this is the case for \stir and \flyvsfly. For \calms and \digitaltyphoon, we observe best performance at 80K steps. We find our selected 40k step training schedule to be a good compromise across all tasks for performance and evaluation speed.
Further reducing the training to 20k steps negatively affects performance -- for example, on \calms and \flyvsfly, reducing the number of training steps from 40k to 20k results in a 3\% absolute mAP decrease -- while increasing the training to 80k yields at best a +1\% mAP increase.
The readout training length has little impact on \weatherbench, within noise levels (see Tab.~\ref{tab:task_stats}).

\begin{table*}[!htbp]
\centering
\small
\setlength{\tabcolsep}{15pt}
\begin{tabular}{lcccc|c|c|c}
\toprule
& & &  & & \multicolumn{3}{c}{WeatherBench 2} \\
 & {CalMS21} & {FlyVsFly} & {STIR} & {Digital Typhoon} & {Z500} & {T850} & {Q700} \\
\cmidrule(lr){2-5} \cmidrule(lr){6-8}
Mean & 0.791 & 0.889 & 0.746 & 4.32 & 603 & 2.88 & 1.56e-03 \\
Std & 1.12e-02 & 2.68e-03 & 5.97e-03 & 0.113 & 1.04 & 5.35e-03 & 6.37e-07 \\
\bottomrule
\end{tabular}
\caption{We assess the stochasticity of our evaluations by computing the standard deviation for each task across 5 random seeds. We train the readout with frozen features from the 4DS-e model (\frozen) as input in the same setting as Tab.~\ref{tab:backbone_comparison} of the main paper.}
\label{tab:task_stats}
\end{table*}

\subsection{Temporal dynamics for frame-based model}
\label{subsec:temporal_dynamics}
\begin{figure}[H]
\centering
\includegraphics[trim = 0 1.1cm 0 0, clip=true,width=0.8\linewidth]{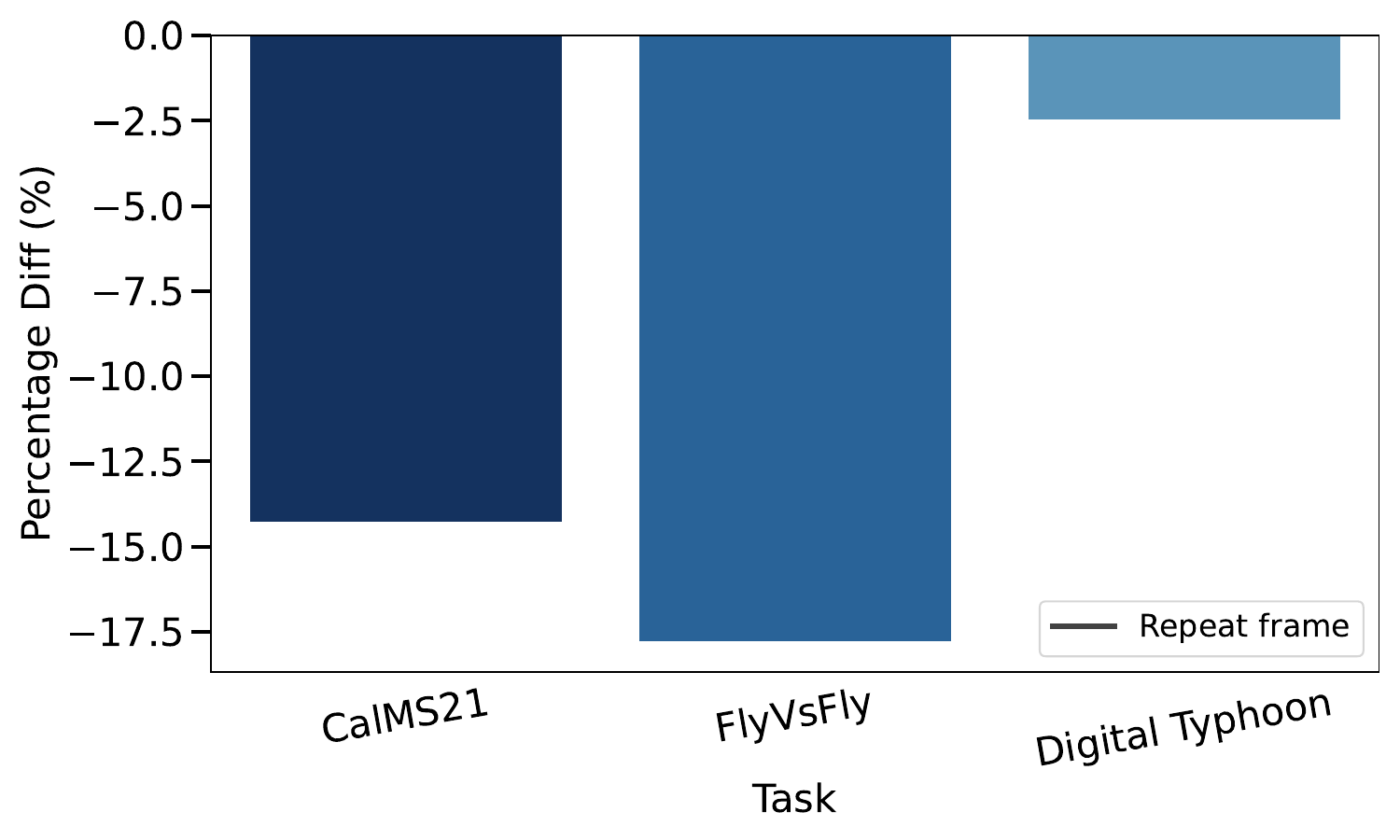}
\caption{\textbf{Performance change when relying on a single frame for the readout training.} In this figure we show the (percentage) performance when training a readout with a frozen DINOv2-g backbone (\frozen) and a single video frame repeated along the temporal dimension, compared to using all original video frames. We observe a consistent performance drop when using a single frame, highlighting the importance of temporal understanding in our benchmarks.}
\label{fig:temporal_dynamics_dino_v2}
\end{figure}

In this section, we investigate frame-based performance on a subset of \scivid tasks which are easily amenable to this setup.
For this, compare how performance changes when training a readout with a frozen DINOv2-g backbone when repeating a single frame along the temporal dimension with respect to its counterpart trained on the original videos. We repeat the middle frame for CalMS21 and FlyVsFly datasets, and use the last frame for Digital Typhoon, which should be the most informative with respect to the task.
We apply the same logic on both training and evaluation sets for consistency.
We show the results in Fig.~\ref{fig:temporal_dynamics_dino_v2}. We observe that training our readout with a frame-based model using a single frame compared to all frames consistently degrades performance, with the largest drop on FlyVsFly -- this highlights the importance on temporal understanding in our selected benchmarks.

\subsection{Evaluation noise}
\label{subsec:eval_noise}

\begin{figure}[H]
\centering
\includegraphics[trim = 0 1.2cm 0 0, clip=true,width=\linewidth]{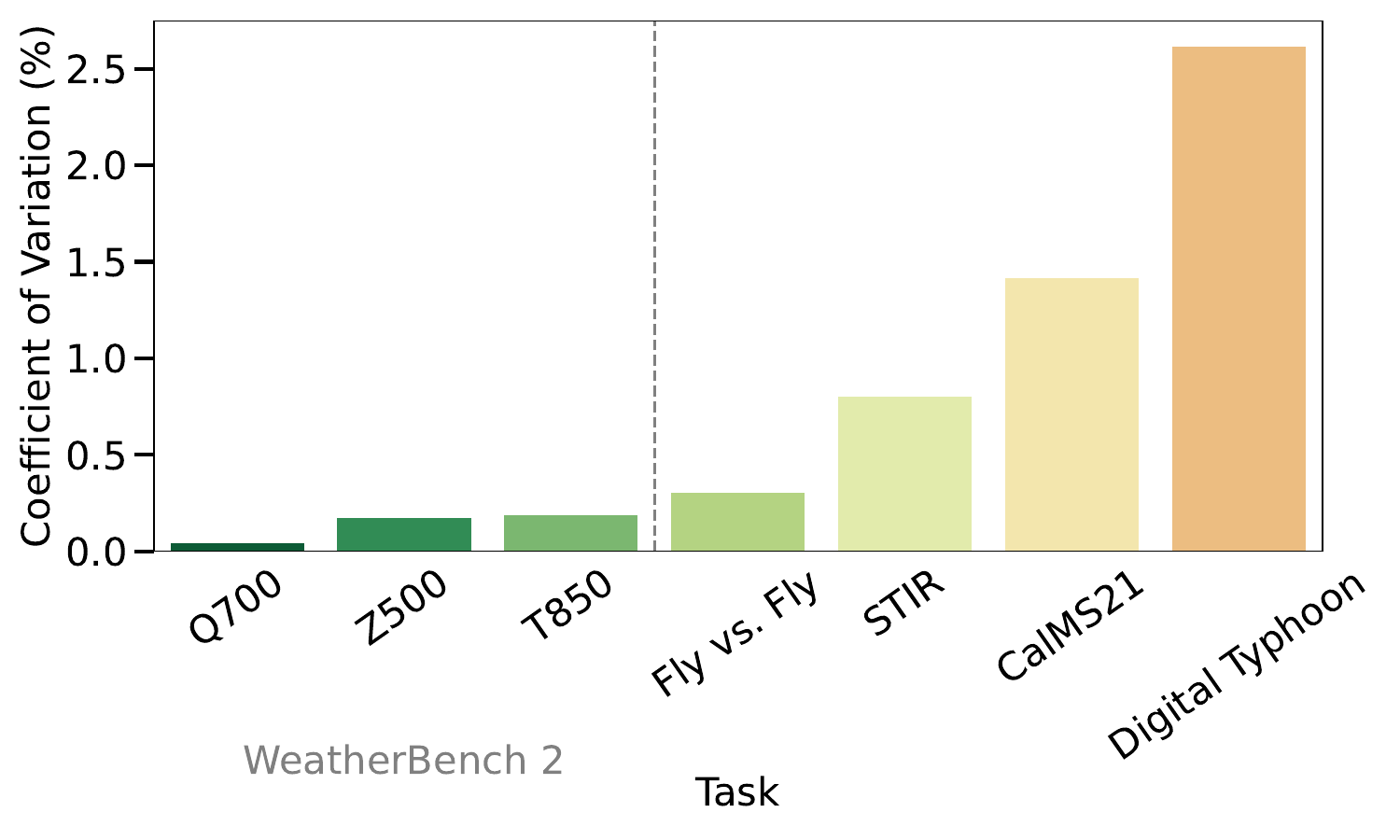}
\caption{\textbf{Evaluation noise.} We compare the noise levels for each task, with our readout on top of the frozen 4DS-e backbone (\frozen), across different value ranges by computing the coefficient of variations. We observe varying levels of relative noise across tasks, with the highest levels for Digital Typhoon~\cite{digital_typhoon_NEURIPS2023}, which has the smallest evaluation set.}
\label{fig:eval_noise}
\end{figure}

Each dataset and evaluation comes with a different level of stochasticity across experiments which are identical up to random seed initialization.
We evaluate the noise of each evaluation for our readout with the frozen 4DS-e backbone in the setup described Sec.~\ref{tab:backbone_comparison} of the main paper. In Fig.~\ref{fig:eval_noise}, we observe varying levels of relative noise across tasks, with the highest levels for Digital Typhoon~\cite{digital_typhoon_NEURIPS2023}, which has the smallest evaluation set. To complement Fig.~\ref{fig:eval_noise}, we provide in see Tab.~\ref{tab:task_stats} the corresponding mean and standard deviations computed across 5 seeds. In particular, on \digitaltyphoon we observe a standard deviation of 0.113 for a mean value of 4.32 -- this confirms the important noise levels observed in Fig.~\ref{fig:eval_noise}.

\renewcommand{\thefigure}{D.\arabic{figure}}
\setcounter{figure}{0} 
\renewcommand{\thetable}{D.\arabic{table}}
\setcounter{table}{0} 
\section{Additional qualitative results}\label{sec:supp_qual_results}

Here, we first qualitatively analyse the input distributions across tasks (Sec.~\ref{subsec:supp_input_distribution}). We then show additional qualitative results, for FlyVsFly and \calms (Sec.~\ref{subsec:supp_classif_qual}), \stir (Sec.~\ref{subsec:supp_stir_qual}) and \weatherbench (Sec.~\ref{subsec:supp_wb2_qual}).

\begin{table}[!h]
    \centering
    \setlength{\tabcolsep}{1pt}
    \hspace{8em} {\small Pixel intensities}
    \begin{tabular}{c|c}
        \small Kinetics  &  \includegraphics[trim = 0 0.8cm 0 0, clip=true,width=0.73\linewidth]{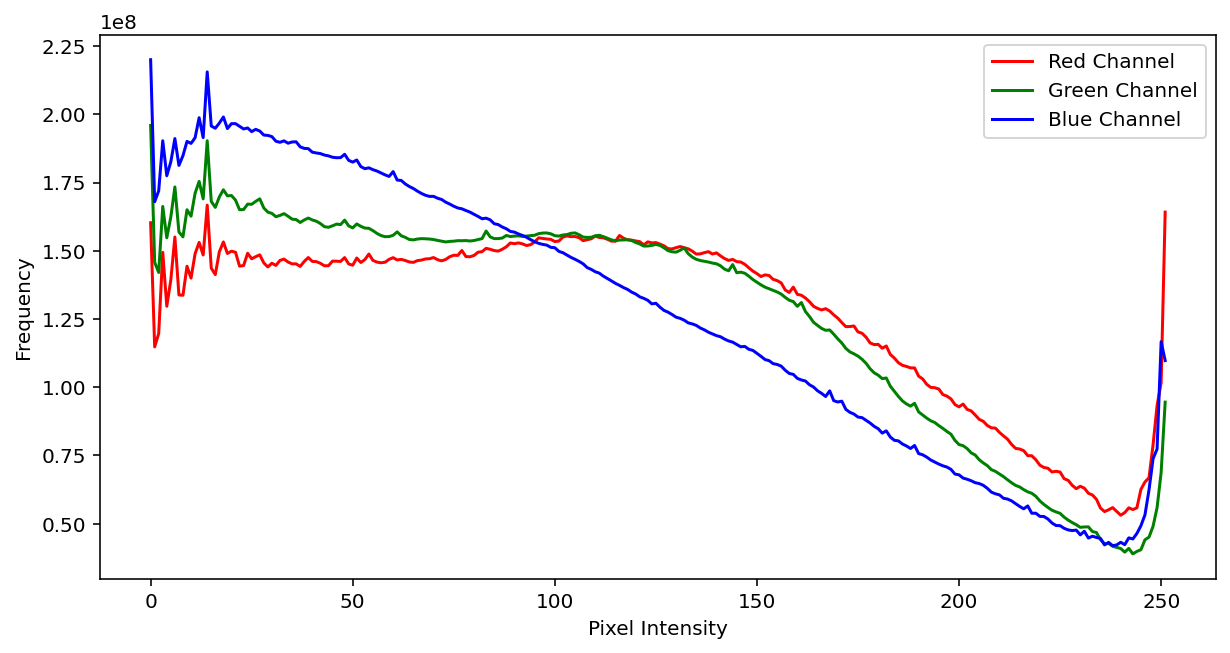} \\
        \midrule
       \small \flyvsfly & \includegraphics[trim = 0 0.8cm 0 0, clip=true,width=0.73\linewidth]{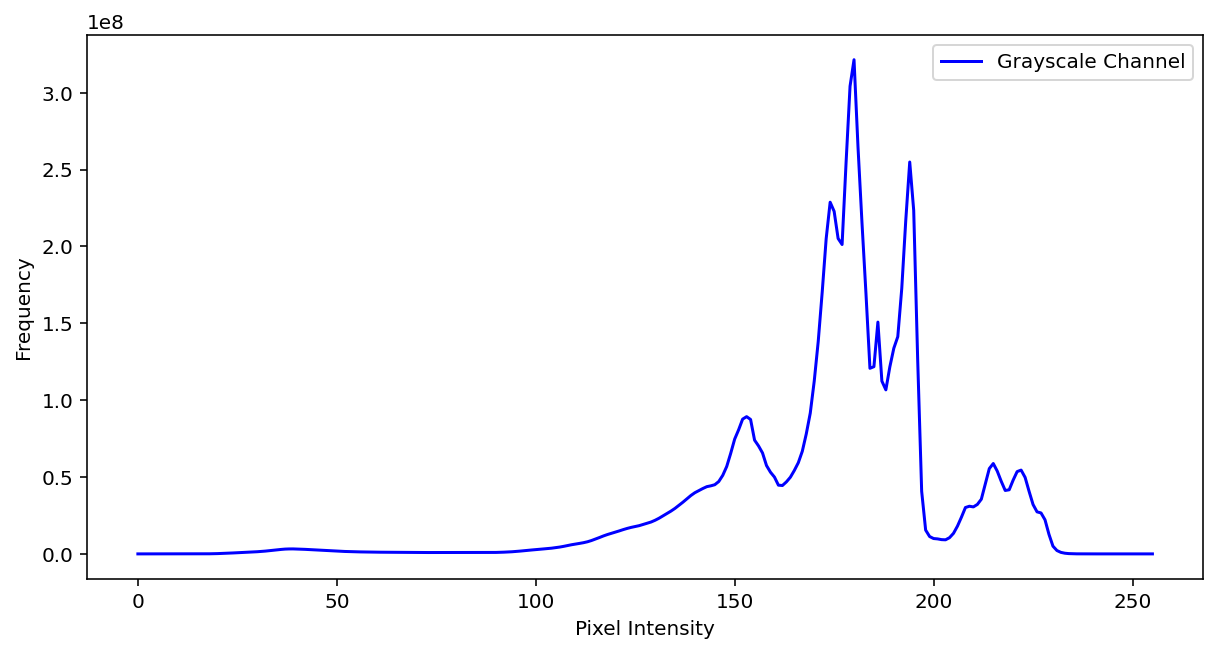} \\
        \midrule
       \small \calms &   \includegraphics[trim = 0 0.8cm 0 0, clip=true,width=0.73\linewidth]{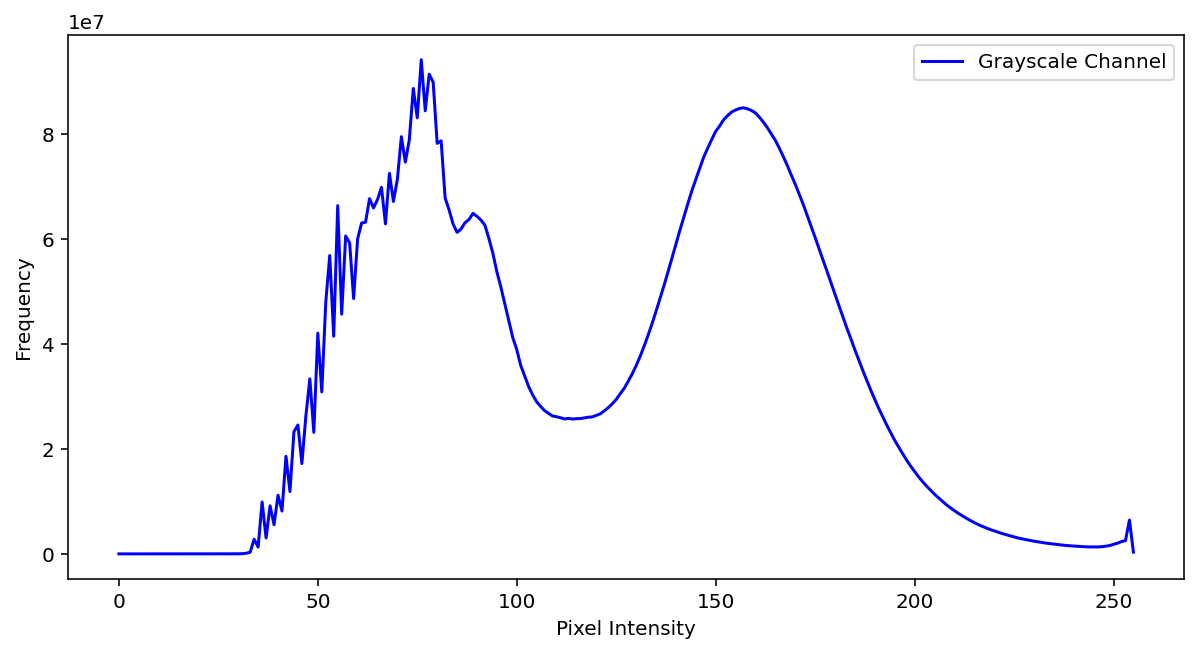} \\
        \midrule
        \small \stir &  \includegraphics[trim = 0 0.8cm 0 0, clip=true,width=0.73\linewidth]{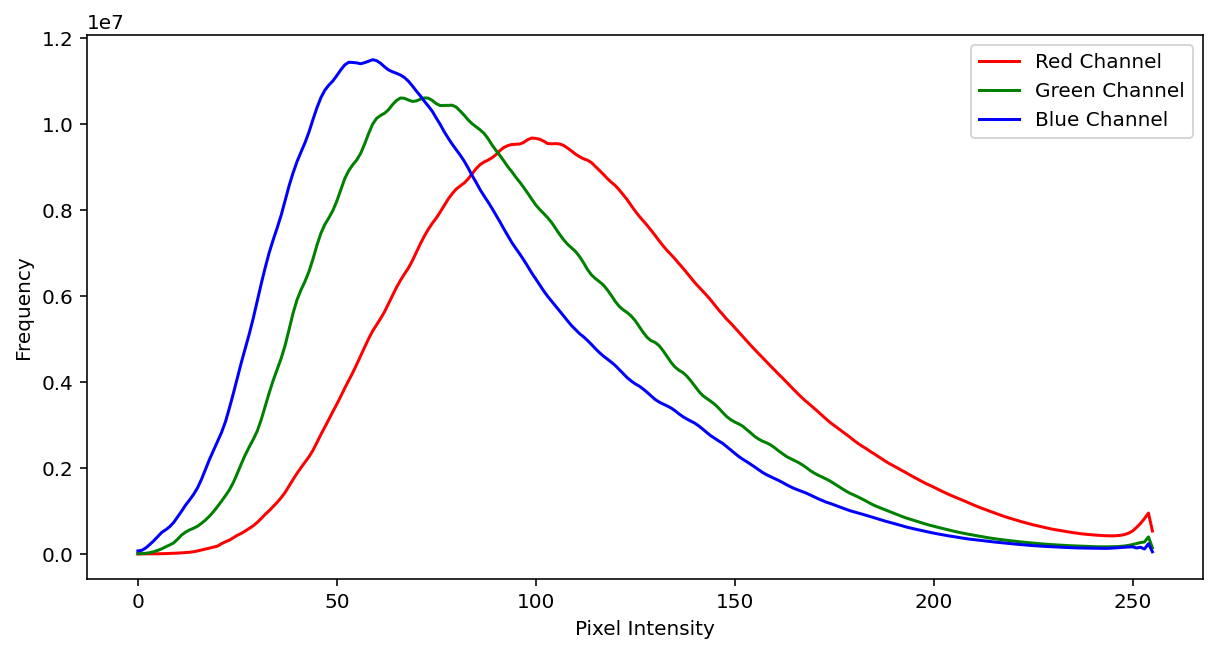} \\
         \midrule
        \small \weatherbench &  \includegraphics[trim = 0 0.8cm 0 0, clip=true,width=0.73\linewidth]{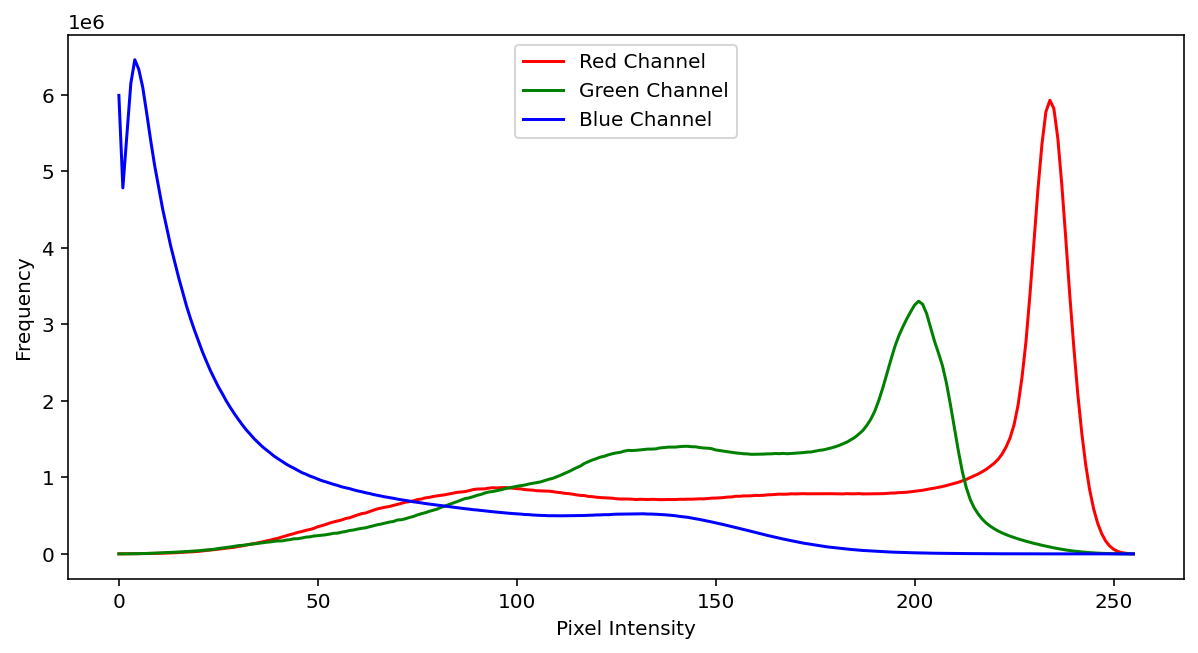} \\
         \midrule
        \small \digitaltyphoon &  \includegraphics[trim = 0 0.8cm 0 0, clip=true,width=0.73\linewidth]{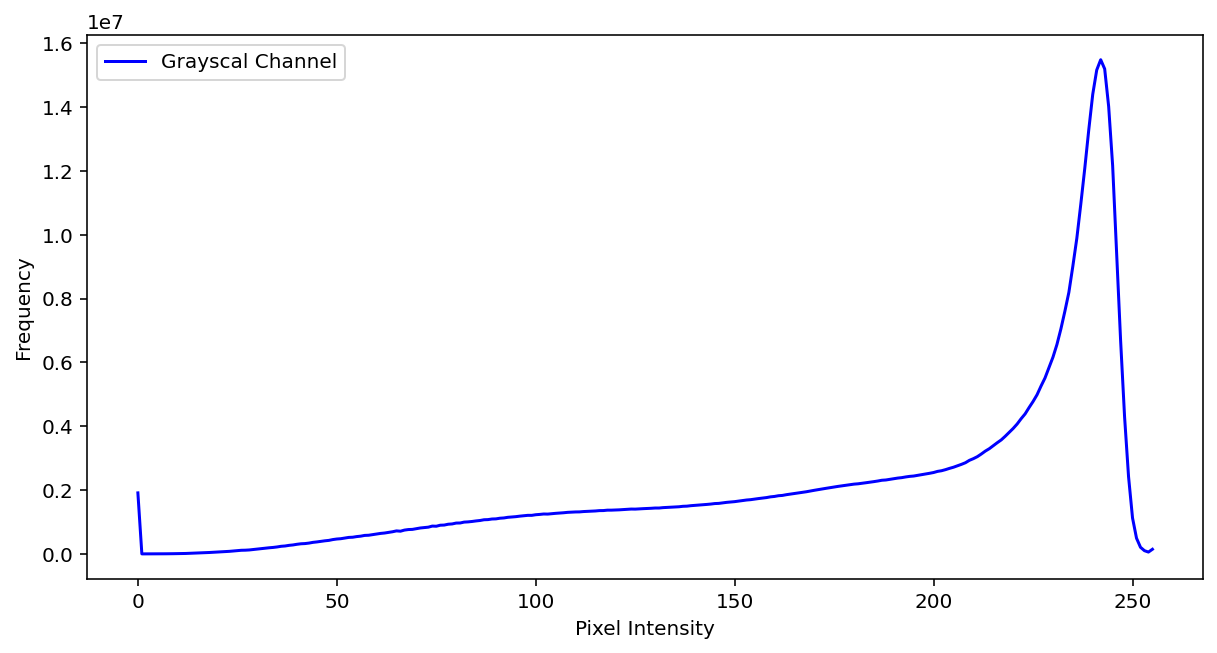} \\
    \end{tabular}
    \caption{\textbf{Pixel histograms.} The histograms of pixel intensity values of input variables (shown as Greyscale and RGB channels) in \scivid (last 5) exhibit significant deviations from the histograms of pixel values in datasets commonly used for pre-training ViFMs such as Kinetics~\cite{kay2017kineticshumanactionvideo} (top row), indicating a notable shift in the underlying data distributions.  \label{tab:histograms}}
\end{table}

\subsection{Distribution of input variables} \label{subsec:supp_input_distribution}
To highlight some of the challenges associated with our benchmark, we plot the histograms of input values in the different tasks of \scivid in Tab.~\ref{tab:histograms}, and compare them to the histograms of pixel values in natural image videos which are typically used for ViFM pretraining. While the pretraining distribution of ViFMs is often not released, it is frequently mined from YouTube. As a proxy, we use the public Kinetics dataset \cite{kay2017kineticshumanactionvideo} and visualize the histogram of the pixel values in that dataset in Tab.~\ref{tab:histograms} top. In the 5 tasks that we consider, we interpret the input values either as greyscale or RGB channels for tasks that involve frames with single-scalar and multi-scalar data input respectively. As can be observed in Tab.~\ref{tab:histograms} the distribution of pixel values in our tasks differs significantly from that observed in a typical video datasets, which consists of ``natural'' online videos, e.g., depicting people performing various actions.

This shift in data distribution highlights the unique characteristics in the tasks and data in \scivid, and underscores the  need to develop methods that can accommodate such variability. We also point out that in natural videos there is often a strong correlation between the values in the three RGB channels, which might be captured by a trained ViFM. On the other hand, as can be seen for \weatherbench in Tab.~\ref{tab:histograms}, the three variables that we model have very different correlation properties, which might make it difficult to exploit the pretrained model. In this plot, the RGB channels correspond to, respectively, Geopotential@500, Temperature@850 and Specific Humidity@700. This shift, again, highlights the need for generalizable models that can adapt to potentially significantly different new data, by efficiently taking advantage of the available supervision.

\subsection{FlyVsFly and \calms}
\label{subsec:supp_classif_qual}

\begin{figure*}[h]
  \centering
  \includegraphics[width=\textwidth]{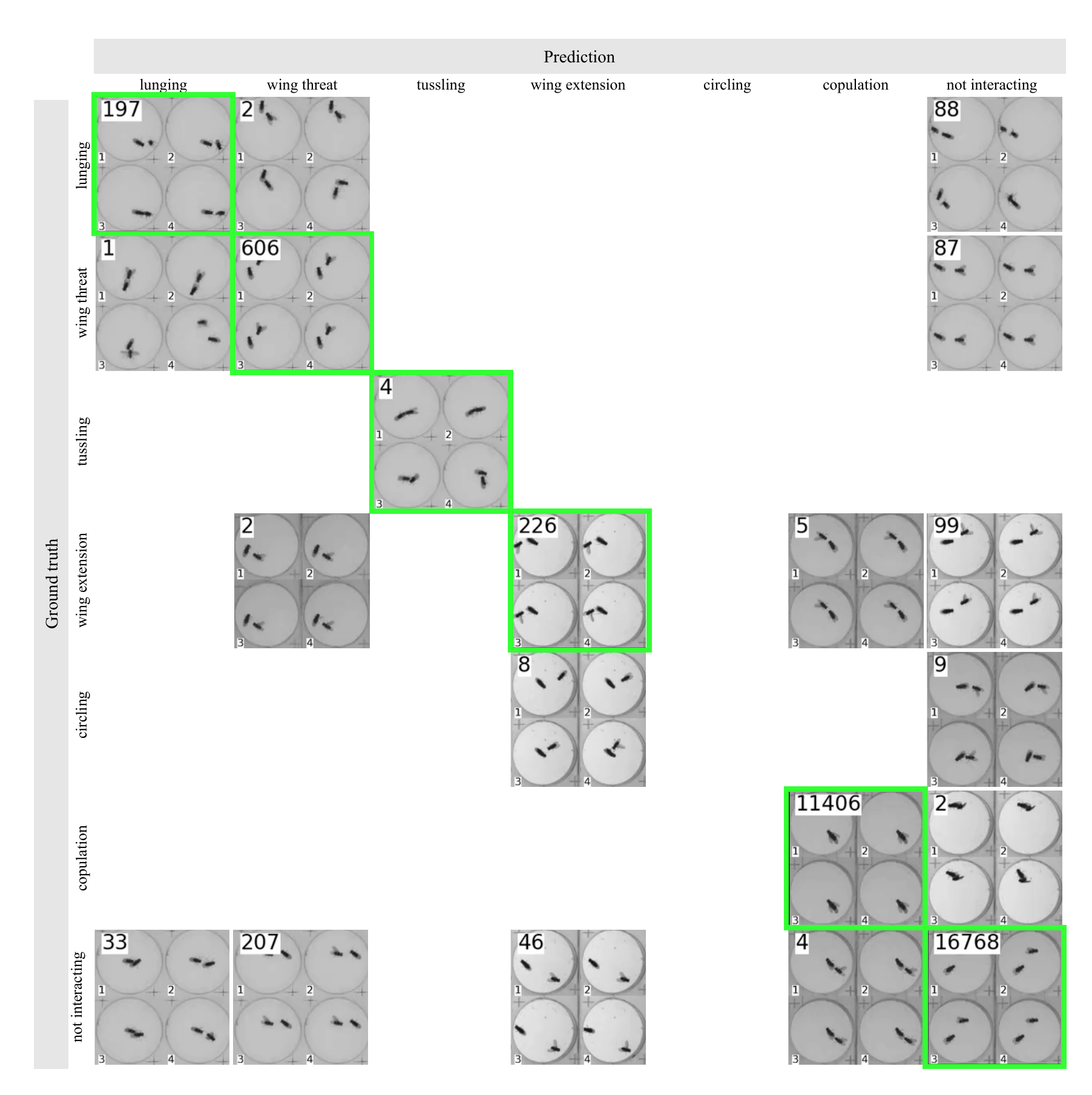}
  \caption{\textbf{Visual confusion matrix on the \flyvsfly dataset.} We show results from our readout trained on top of the frozen 4DS-e backbone (\frozen).
    This matrix is generated from $\sim$30k samples of the validation set, after filtering out multi-label video clips. Each cell represents a ground truth/prediction pair and displays four representative frames (first, two middle, and last) from a corresponding video clip. 
    The number in the top-left corner of each cell indicates the total count of videos for that specific pair. We observe instances of ambiguity, such as the video labeled as ``circling'' but predicted as ``wing extension'' where the behavior indeed transitions to include ``wing extension'' towards the end. This is due to the fact that the behavior active around the middle frame was selected during labeling, while the model may capture behavior prominent at other moments.}
  \label{fig:flyvsfly_confusion_matrix}
\end{figure*}

\begin{figure*}[h]
  \centering
  \vspace{3em}
  \includegraphics[width=\textwidth]{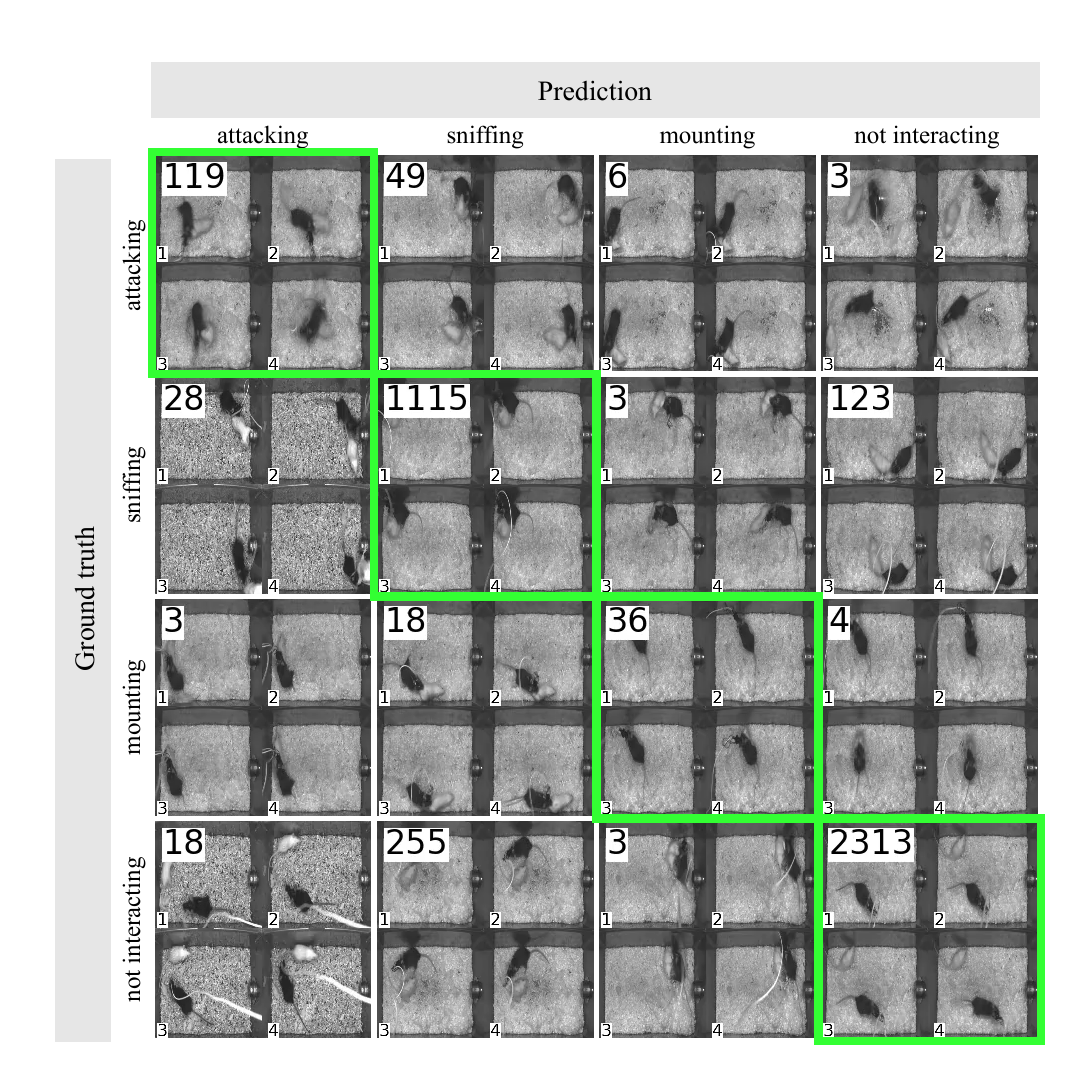}
  \caption{\textbf{Visual confusion matrix on the \calms dataset.}  We show results from our readout trained on top of the frozen 4DS-e backbone (\frozen).
    This matrix is generated from $\sim$4k samples of the validation set, after filtering out multi-label video clips. Each cell represents a ground truth/prediction pair and displays four representative frames (first, two middle, and last) from a corresponding video clip. 
    The number in the top-left corner of each cell indicates the total count of videos for that specific pair. Again, we observe instances of ambiguity, such as the video labeled as ``not interacting'' but predicted as ``mounting'', but where the animals are clearly on top of one another.}
  \label{fig:calms21_confusion_matrix}
  \vspace{3em}
\end{figure*}

We illustrate the animal behavior classification performance for our readout with the frozen 4DS-e model by displaying the  confusion matrices on \flyvsfly in Fig.~\ref{fig:flyvsfly_confusion_matrix} and \calms in Fig.~\ref{fig:calms21_confusion_matrix}. In both cases we observe instance of ambiguity and possible labelling errors in Fig.~\ref{fig:calms21_confusion_matrix}. 

\subsection{STIR}
\label{subsec:supp_stir_qual}

\begin{figure*}
\centering
\begin{tabularx}{0.8\linewidth}{YYY}
    Query points & 4DS-e & DinoV2-g \\
\end{tabularx}
\newlength{\imagewidth}
\settowidth{\imagewidth}{\includegraphics{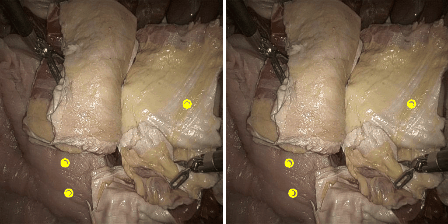}}
\includegraphics[trim = 0 0 0.5\imagewidth{} 0, clip=true, width=0.26\linewidth]{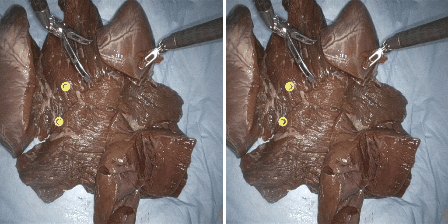} \hspace{-0.4em} \includegraphics[width=0.52\linewidth]{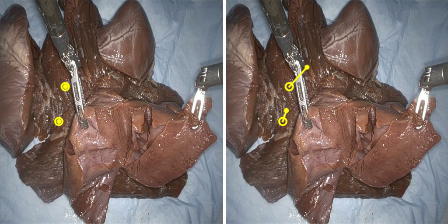} \\
\includegraphics[trim = 0 0 0.5\imagewidth{} 0, clip=true, width=0.26\linewidth]{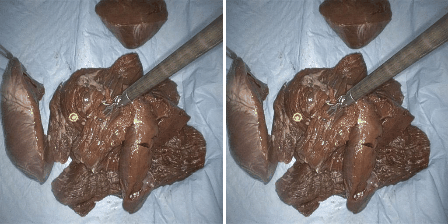} \hspace{-0.4em} \includegraphics[width=0.52\linewidth]{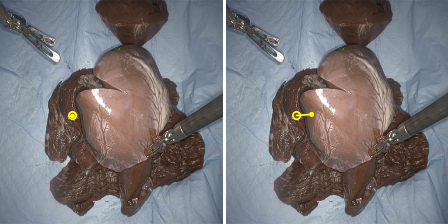} \\
\includegraphics[trim = 0 0 0.5\imagewidth{} 0, clip=true, width=0.26\linewidth]{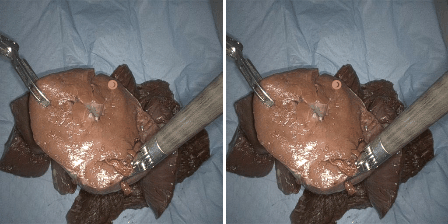} \hspace{-0.4em} \includegraphics[width=0.52\linewidth]{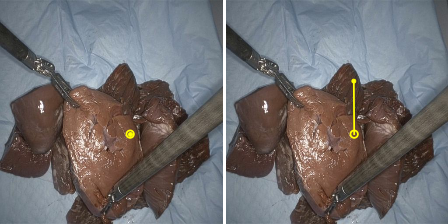} \\
\includegraphics[trim = 0 0 0.5\imagewidth{} 0, clip=true, width=0.26\linewidth]{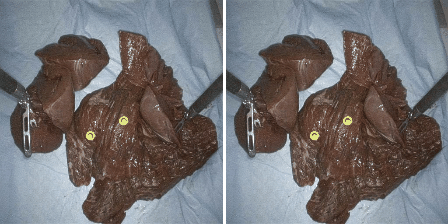} \hspace{-0.4em} \includegraphics[width=0.52\linewidth]{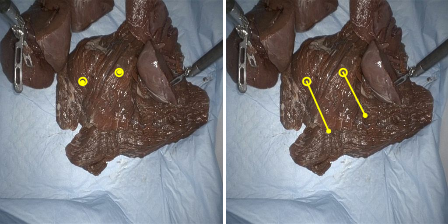} \\
\caption{\textbf{Point tracking error comparison on the STIR dataset.} The first column shows the initial query points, while the subsequent columns display the predicted locations in the final frame for our readout with frozen 4DS-e and DinoV2-g backbones (\frozen). Ground truth points are represented by large circles, predicted locations by smaller circles, and the error between them by connecting lines. Notably, DinoV2-g exhibits significant tracking errors, even falling short of the control baseline. In contrast, 4DS-e demonstrates superior performance in diverse and challenging scenarios: points close to moving objects (row 1), occlusion events (row 2), points on moving objects (row 3), and camera motion (row 4).}
\label{fig:stir_model_comparison}
\end{figure*}

In Fig.~\ref{fig:stir_model_comparison} we compare the performance of our readout using the 4DS-e and DinoV2-g backbones with finetuning.
We observe significant tracking errors when using the image-based DinoV2-g features while 4DS-e demonstrates stronger performance and successful tracking across challenging scenarios, including camera motion and occlusion events.

\subsection{\weatherbench}
\label{subsec:supp_wb2_qual}

We provide a qualitative comparison between the residual (i.e., predicting the differences between the target and the last seen frame) and direct target prediction in Fig.~\ref{fig:wb_residual}. We plot the results of the best-performing backbone in the finetuning setting, and visualize the predictions on the last frame for a particular test sample. We observe that by predicting the residual (Fig.  \ref{fig:wb_residual}  middle column), the model can forecast more detailed results but produces visible high-frequency artefacts. On the other hand, direct prediction (Fig.~\ref{fig:wb_residual} right column) does not exhibit these artefact but seems to over-smooth the results. These two complementary failure modes suggest that capturing high-frequency details using the readouts we experimented with without producing artefacts is non-trivial in this task.

\begin{figure*}[!htbp]
\centering
\setlength{\tabcolsep}{1pt}
\settowidth{\imagewidth}{\includegraphics{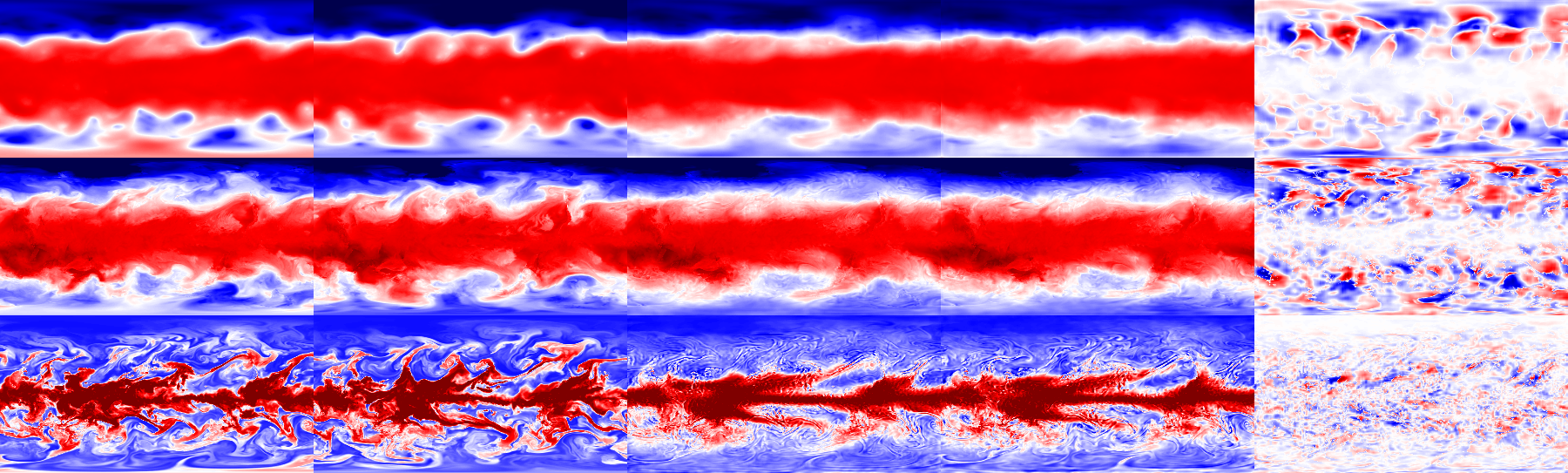}}

\begin{tabularx}{\linewidth}{>{\centering\arraybackslash}m{0.25cm}YYYY}
    & Ground truth & 4DS-e & DinoV2-g & Error diff.\ \\
 \adjustbox{valign=c,rotate=90,raise=17em}{Q700 ~~~~~~~~~~~~ T850 ~~~~~~~~~~~~ Z500} & \includegraphics[trim = 0.2\imagewidth{} 0 0.6\imagewidth{} 0, clip=true, width=\linewidth]{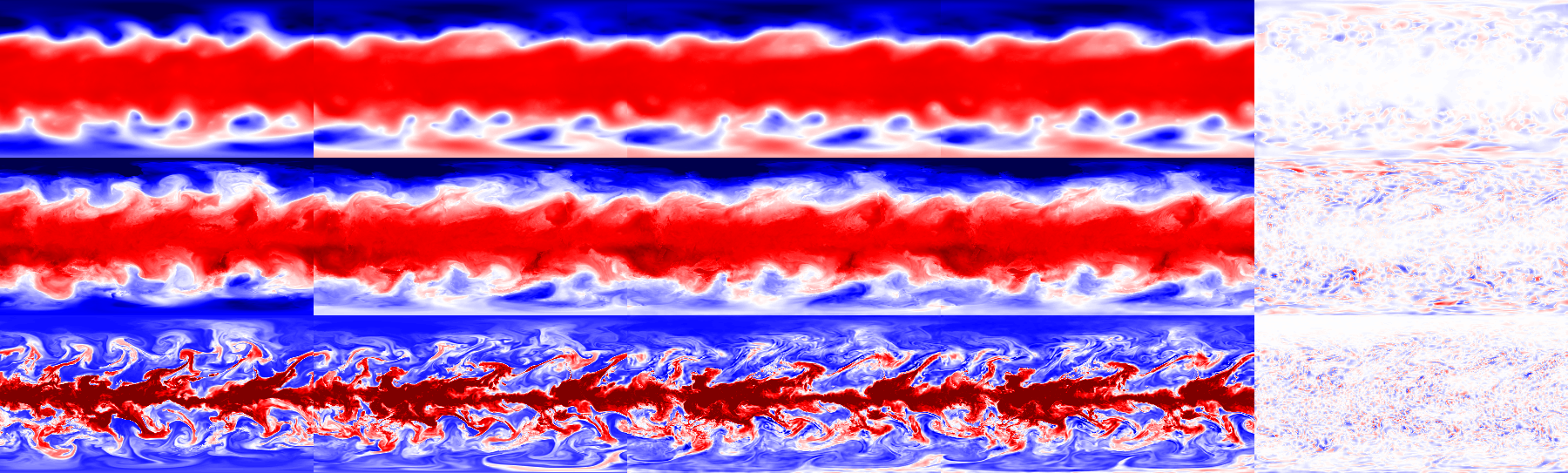} & \includegraphics[trim = 0.4\imagewidth{} 0 0.4\imagewidth{} 0, clip=true, width=\linewidth]{figures/images/weather/150607198_150606667_test_eval_weather_first_1.png} & \includegraphics[trim = 0.6\imagewidth{} 0 0.2\imagewidth{} 0, clip=true, width=\linewidth]{figures/images/weather/150607198_150606667_test_eval_weather_first_1.png} & \includegraphics[trim = 0.8\imagewidth{} 0 0 0, clip=true, width=\linewidth]{figures/images/weather/150607198_150606667_test_eval_weather_first_1.png} \\[-14.5em]
 & \multicolumn{4}{c}{(a) 1st future step.}
\end{tabularx}

\begin{tabularx}{\linewidth}{>{\centering\arraybackslash}m{0.25cm}YYYY}
    & Ground truth & 4DS-e & DinoV2-g & Error diff.\ \\
 \adjustbox{valign=c,rotate=90,raise=17em}{Q700 ~~~~~~~~~~~~ T850 ~~~~~~~~~~~~ Z500} & \includegraphics[trim = 0.2\imagewidth{} 0 0.6\imagewidth{} 0, clip=true, width=\linewidth]{figures/images/weather/150607198_150606667_test_eval_weather_last_1.png} & \includegraphics[trim = 0.4\imagewidth{} 0 0.4\imagewidth{} 0, clip=true, width=\linewidth]{figures/images/weather/150607198_150606667_test_eval_weather_last_1.png} & \includegraphics[trim = 0.6\imagewidth{} 0 0.2\imagewidth{} 0, clip=true, width=\linewidth]{figures/images/weather/150607198_150606667_test_eval_weather_last_1.png} & \includegraphics[trim = 0.8\imagewidth{} 0 0 0, clip=true, width=\linewidth]{figures/images/weather/150607198_150606667_test_eval_weather_last_1.png} \\[-14.5em]
 & \multicolumn{4}{c}{(b) 16th future step.}
\end{tabularx}

\caption{\textbf{Qualitative comparison on \weatherbench.} We present ground truth and predicted values for Geopotential at 500hPa (Z500), Temperature at 850hPa (T850), and Specific Humidity at 700hPa (Q700) at (a) the 1st and (b) the 16th future step (12-hour intervals between each step). We show predictions for our readout on top of frozen 4DS-e and DinoV2-g backbones (\frozen), both \withoutcond. Ground truth and predictions are normalized using the training set's mean and twice the standard deviation for each variable. These are visualized using the ``seismic'' colormap. Error differences are also shown, normalized to highlight relative performance: blue indicates 4DS-e outperforms DinoV2-g, red indicates DinoV2-g outperforms 4DS-e, and white indicates comparable performance.\label{fig:weatherbench_4ds_vs_dinov2}}
\end{figure*}

We also show a visual comparison between our readout on top of a frozen ViFM (4DS-e) and an image-based backbone (DinoV2-g), both \withoutcond, in Fig.~\ref{fig:weatherbench_4ds_vs_dinov2}. We show the prediction of the first and the last (16th) future frame for a particular test sample. We note that both backbones lead to strong artefacts (discussed above) that are especially visible at the last predicted frame. Moreover, quantitatively, in this particular example, 4DS-e outperforms DinoV2-g by approximately 20\%. However, it is difficult to note specific differences between the two predictions, highlighting that qualitative interpretation is not easy for an untrained eye.

\end{document}